\pdfoutput=1

\documentclass{article}

\usepackage{bosonai_techreport}

\usepackage[utf8]{inputenc}
\usepackage[T1]{fontenc}
\usepackage{microtype}
\usepackage{graphicx}
\usepackage{subcaption}
\usepackage{booktabs}
\usepackage{tabularx}
\usepackage{hyperref}
\usepackage{url}
\usepackage{xcolor}
\usepackage{xspace}
\usepackage{wrapfig}   %
\usepackage{needspace} %

\usepackage{amsmath}
\usepackage{amssymb}
\usepackage{mathtools}
\usepackage{alex}

\usepackage[capitalize,noabbrev]{cleveref}

\usepackage[textsize=tiny]{todonotes}

\usepackage{algorithm}
\usepackage{algorithmic}
\usepackage{mdwlist}
\usepackage{longtable}

\newcommand{\R}{\mathbb{R}}

\newcommand{\Aset}{\mathcal{A}}

\newcommand{\Smat}{B}  %

\newcommand{\tr}{\mathop{\mathrm{tr}}}

\newcommand{\MMLU}{\textsc{MMLU}\xspace}
\newcommand{\MTEB}{\textsc{MTEB}\xspace}
\newcommand{\Merged}{\textsc{Merged}\xspace}
\newcommand{\BenchPress}{BenchPress\xspace}
\newcommand{\OLLtwo}{OLL~v2\xspace}            %
\newcommand{\OLLone}{OLL~v1 items\xspace}      %
\newcommand{\Epoch}{Epoch\xspace}              %
\newcommand{\HELMLite}{\textsc{HELM} Lite\xspace}          %
\newcommand{\HELMMMLU}{\textsc{HELM} \textsc{MMLU}\xspace} %
\newcommand{\HELMClassic}{\textsc{HELM} Classic\xspace}    %
\newcommand{\MTEBnew}{\textsc{MTEB}-2026\xspace}           %
\newcommand{\MergedVone}{\textsc{Merged}-v1\xspace}        %

\newcommand{\Abar}{\bar{\Aset}}                 %
\newcommand{\Rh}{\hat{R}}                       %
\newcommand{\rplus}{r^{+}}                      %
\newcommand{\FFL}{F_{\mathrm{FL}}}              %
\newcommand{\FEV}{F_{\mathrm{EV}}}              %
\newcommand{\FH}{F_{\mathrm{H}}}                %
\newcommand{\FMI}{F_{\mathrm{MI}}}              %
\newcommand{\Rsqrest}[1]{R^2_{#1|\mathrm{rest}}}%
\newcommand{\OPT}{\mathrm{OPT}}

\input{cr/numbers_cmds}
\input{cr/numbers_paper}

\newcommand{\techreport}{}

\newcommand{\codeurl}{https://github.com/smolix/benchmark-selection}

\title{Submodular Benchmark Selection}

\author{Alex Smola}
\affiliation{Boson AI \\ 4677 Old Ironsides Dr \\ Santa Clara CA 95054}
\email{smola@boson.ai}

\renewcommand{\bosonlogo}{boson}

\begin{document}

\maketitle

\begin{abstract}
Evaluating a new language model means running it on a large, growing collection of benchmarks, at considerable cost.
Much of this cost is redundant: arranged as a matrix of models by benchmarks, scores are strongly correlated and concentrate in a few dimensions, so a few well-chosen benchmarks largely determine the rest.
This allows us to run a new model on a few benchmarks and predict the rest with a simple Gaussian model that copes with missing scores. The key question is \emph{which few to run}.
An answer comes from sensor placement, which chooses the observations that carry the most information, measured by entropy or mutual information.
We argue that the objective should instead reward a selection for how well it predicts every benchmark.
This leads to two coverage objectives: explained variance, the predictions' accuracy if the estimated correlations were exact, and facility location, which credits each benchmark with its best proxy among the selected ones.
On eleven score matrices built from public evaluation results, the coverage objectives predict the scores of held-out models best wherever the correlations can be estimated.
The information objectives fall behind because they favor benchmarks that the selection predicts poorly (entropy) or that the others predict well (mutual information), not benchmarks that predict the others. By contrast, facility location predicts practically as well as explained variance, holds up on the addition of newer models and under real costs, has classical guarantees (even under a budget), and gives readable selections.%
\ifdefined\codeurl\\[2pt]\url{\codeurl}\fi

\end{abstract}

\section{Introduction}
\label{sec:intro}

Language models are judged by their scores on many benchmarks.
Leaderboards such as \textsc{HELM}~\citep{Liang23} and the Open LLM Leaderboard~\citep{OpenLLMLeaderboard1,OpenLLMLeaderboard2} run every model on dozens of tasks, the suites keep growing, and the cost recurs for every new model, checkpoint and ablation.

\suppressfloats[t]
\begin{figure}[t]
\centering
\begin{minipage}[c]{0.42\textwidth}
\includegraphics[width=\linewidth]{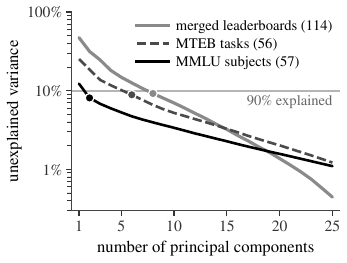}
\end{minipage}\hfill
\begin{minipage}[c]{0.54\textwidth}
\caption{Benchmark scores are nearly low-rank, so a few benchmarks should reveal the rest.
Share of variance that the leading principal components of the benchmarks' correlation matrix leave unexplained, against their number.}
\label{fig:spectrum}
\end{minipage}
\vspace{-5mm}
\end{figure}

Much of this cost is redundant. A poor-quality model is likely to underperform on every benchmark.  
Arrange the scores in a matrix with one row per model and one column per benchmark.
The columns are strongly correlated (a model that does well in college chemistry tends to do well in high-school chemistry), and the matrix is nearly low-rank.
\Cref{fig:spectrum} shows this for three score matrices that we assembled from public results:
the accuracies of \nv{DataMmluM} language models on the \nv{DataMmluN} subjects of the knowledge test \MMLU~\citep{Hendrycks21mmlu}, which we collected from the first version of the Open LLM Leaderboard~\citep{OpenLLMLeaderboard1};
the scores of \nv{DataMtebM} text-embedding models on \nv{DataMtebN} tasks of the Massive Text Embedding Benchmark (\MTEB)~\citep{Muennighoff23};
and the scores of \nv{DataMergedM} models on \nv{DataMergedN} benchmarks that we pooled from nine leaderboards by matching model names.
For these matrices, \nv{EigKMmluVarninety}, \nv{EigKMtebVarninety} and \nv{EigKMergedVarninety} principal components, respectively, explain 90\% of the variance.
A few well-chosen benchmarks should suffice to determine a model's scores. 

This suggests a cheaper evaluation: run a new model on a few selected benchmarks and predict its other scores with a deliberately simple model, the conditional mean of a multivariate Gaussian, which copes with missing scores and needs only the correlations between benchmarks, estimated on the models evaluated before (\cref{sec:predict}).

A classical answer to benchmark selection comes from sensor placement, which puts a few sensors where their readings carry the most information about a spatial field~\citep{GueKraSin05,krause2008near}, measured by the entropy of the readings or their mutual information (MI) with the readings elsewhere.
Yet, information is not prediction error.
For prediction, every benchmark needs a good \emph{proxy} among those run, a selected benchmark that predicts it well, and running one of two near-duplicate chemistry subjects predicts both.
Entropy instead favors a benchmark that the selected ones predict poorly, such as one that resembles no other and so predicts no other; MI favors a benchmark that the others predict well, such as a second near-duplicate or an average of several benchmarks, whether or not it stands in for them.

We therefore derive the selection objective from the metric by which predictions are judged: $R^2$, the share of the variance of the unselected scores that the predictions explain.
If the estimated correlations were exact, a selection's $R^2$ would be fixed by the variance that the selected benchmarks explain, summed over all benchmarks.
We call this sum \emph{explained variance} (EV); it extends the classical criterion of subset selection for regression~\citep{DasKem11} to many targets, and maximizing it maximizes that $R^2$.
Crediting each benchmark only with its best proxy among the selected ones gives \emph{facility location} (FL), which \citet{Yao26coresets} used to select benchmarks; with squared correlations as similarities, it is a lower bound on explained variance.
Both are \emph{coverage} objectives: they reward a selection for how well it predicts every benchmark.
In short: coverage, not information.

On \nv{ScoreNMatrices} score matrices built from public evaluation results, compared on shared splits of the models into training and held-out models, the coverage objectives predict best.
Facility location is never significantly worse than entropy, MI or three simple baselines and is significantly better than each in about half to two-thirds of the comparisons; explained variance is significantly ahead of it in a few, by at most \nv{ScoreFaclocVsTraceMaxDeficit} in $R^2$.
Facility location stays at or near the top on models released later and under real costs.
Selection pays most when only a few benchmarks fit the budget.
We recommend facility location: it predicts practically as well as explained variance; greedy selection, which adds one benchmark at a time, provably reaches at least $1-1/e$ of its optimum (a cost-aware variant half of that under a budget); it uses only pairwise correlations, except to break exact ties; and each selected benchmark is the best proxy for the benchmarks it covers, so selections are easy to check.

Our contributions are
(i)~a derivation of selection objectives from the prediction metric, which shows why the information objectives do not follow from it (\cref{sec:predict,sec:objectives});
(ii)~a paired comparison of seven selectors on \nv{ScoreNMatrices} score matrices, including models released later and per-benchmark costs (\cref{sec:protocol,sec:results,sec:robust});
(iii)~an explanation of the negative result for the information objectives (\cref{sec:mechanism});
and (iv)~a practical recommendation, with code and data.

\section{Predicting scores from a few benchmarks}
\label{sec:predict}

We first fix how the unselected scores are predicted and judged, because the selection objectives of \cref{sec:objectives} follow from both.

\paragraph{Setup, predictor and metric.}
The score matrix $\Smat\in\R^{M\times N}$ holds the scores $\Smat_{mj}$ of $M$ models on $N$ benchmarks. Many scores are typically missing, since leaderboards rarely run every model on every benchmark.
We split the models into \emph{training models}, which stand for the models evaluated so far, and \emph{validation models}, held out to stand for models not yet evaluated.
On the training models we standardize each benchmark, $z_{mj}=(\Smat_{mj}-\hat\mu_j)/\hat\sigma_j$, and estimate the correlation matrix~$\Rh$ of the benchmarks.
A new model runs only a selected set $\Aset\subseteq[N]=\{1,\dots,N\}$ of $k$ benchmarks, and its scores on the unselected benchmarks $\Abar=[N]\setminus\Aset$ are to be predicted.
We treat a model's standardized scores as a draw from a Gaussian distribution with mean zero and covariance~$\Rh$, and predict the unselected scores by their conditional mean given the selected ones, $\hat z_{\Abar}=\Rh_{\Abar\Aset}\Rh_{\Aset\Aset}^{-1}z_{\Aset}$: a linear regression on the selected benchmarks, with coefficients taken from~$\Rh$.
For a validation model~$m$, let $\Aset_m\subseteq\Aset$ be the selected benchmarks it has scores on and $T_m\subseteq\Abar$ its observed unselected benchmarks, the \emph{targets}.
We predict the targets with a small ridge $\lambda=10^{-2}$, clip each prediction to its benchmark's valid range, and pool the squared errors over all validation models:
\begin{equation}
  \hat z_{m,T_m} = \Rh_{T_m\Aset_m}\bigl(\Rh_{\Aset_m\Aset_m}+\lambda I\bigr)^{-1} z_{m,\Aset_m},
  \qquad
  R^2(\Aset) = 1-\frac{\sum_m\sum_{j\in T_m}(z_{mj}-\hat z_{mj})^2}{\sum_m\sum_{j\in T_m} z_{mj}^2}.
  \label{eq:r2}
\end{equation}
An $R^2$ of one means perfect predictions; zero means no better than the training mean ($z=0$).

\paragraph{Why a linear Gaussian model.}
We chose this simple model for four reasons.
First, it needs only means and correlations, which can be estimated from incomplete score tables, pairwise (each correlation from the models that report both scores) or by expectation maximization (EM)~\citep{DempsterLairdRubin77,LittleRubin19}, and it predicts from whichever selected scores a model has.
Second, the same formula with the true correlations~$R$ and no ridge, $R_{\Abar\Aset}R_{\Aset\Aset}^{-1}z_{\Aset}$, is the linear predictor with the smallest expected squared error for every distribution with mean zero and covariance~$R$, Gaussian or not.
Squared error is also how we judge the predictions, so we need no normality, which tests reject for most benchmarks (\cref{app:protocol-normality}).
Third, the predictor, the $R^2$ it can be expected to reach (\cref{sec:objectives}) and every selection objective are functions of the same correlation matrix, so the whole pipeline is coherent and in closed form.
Fourth, the predictor is simple and the same for every selector, so differences in $R^2$ reflect the selections, not a flexible predictor that makes up for a poor choice.
Flexible predictors also did not help: low-rank matrix completion~\citep{Mazumder10softimpute} was not significantly better on facility location's selections, the pretrained imputer TabImpute~\citep{Feitelberg26tabimpute} was far worse, and logit-transformed scores did not help the coverage objectives (\cref{app:checks-imputers,app:checks-logit}).

\section{From the metric to selection objectives}
\label{sec:objectives}

We want to find the $k$ benchmarks that maximize the $R^2$ on validation models.
$R^2$ is unknown when the benchmarks are chosen, since the validation scores are held out, but we can compute the $R^2$ that a selection would reach if $\Rh$ were the true correlation matrix, its \emph{plug-in} $R^2$, and turn it into objectives.

\paragraph{Greedy selection.}
Searching all sets of $k$ benchmarks is infeasible beyond small~$k$, so we select greedily: starting from the empty set, we add one benchmark at a time, each time the candidate~$j$ with the largest gain $\Delta(j|\Aset)=F(\Aset\cup\{j\})-F(\Aset)$ in an objective~$F$, until $k$ are selected.
Greedy selection is provably near-optimal when $F$ is normalized ($F(\emptyset)=0$), \emph{monotone}, so that a pick never lowers it, and \emph{submodular}, so that gains diminish as the selection grows: $\Delta(j|\Aset)\ge\Delta(j|\Aset')$ for $\Aset\subseteq\Aset'\not\ni j$.
The greedy set then reaches at least $1-1/e\approx0.63$ of the best value over all sets of size~$k$~\citep{nemhauser1978analysis}, a factor that no efficient algorithm can improve in general~\citep{Feige98}.
Submodularity formalizes redundancy: once one of two near-duplicate chemistry subjects is selected, the other adds little.

\paragraph{Explained variance.}
If a model's standardized scores have covariance~$R$ (a correlation matrix; $\Rh$ in practice) and it has run the benchmarks in~$\Aset$, the errors of the predictor of \cref{sec:predict} (without ridge) on the unselected benchmarks have the \emph{residual covariance} $S(\Aset)=R_{\Abar\Abar}-R_{\Abar\Aset}R_{\Aset\Aset}^{-1}R_{\Aset\Abar}$.
For each benchmark~$j$, $\sigma^2_{j|\Aset}=1-R_{j\Aset}R_{\Aset\Aset}^{-1}R_{\Aset j}$ is the variance that a linear regression on the benchmarks in~$\Aset$ leaves unexplained (a diagonal entry of~$S$ for $j\notin\Aset$), and $R^2(j|\Aset)=1-\sigma^2_{j|\Aset}$ is the variance it explains ($1$ for $j\in\Aset$, $0$ for $\Aset=\emptyset$, and the squared correlation $R_{ij}^2$ for $\Aset=\{i\}$).
A good selection makes $S$ small, and the four objectives differ in how they measure its size. Note that squared error measures it by its trace. As such, \emph{explained variance}, $\FEV(\Aset)=\sum_{j=1}^N R^2(j|\Aset)=N-\tr S(\Aset)$, is the metric in disguise.

\begin{proposition}[Explained variance determines $R^2$]
\label{prop:identity}
Let $R$ be a positive definite correlation matrix (unit diagonal), and suppose that a new model's standardized scores $z\in\R^N$ have mean~$0$ and covariance~$R$, are fully observed, and are predicted by $\hat z_{\Abar}=R_{\Abar\Aset}R_{\Aset\Aset}^{-1}z_{\Aset}$ (no ridge, no clipping).
Then, for $0<k<N$,
\begin{equation}
  R^2_{\mathrm{pop}}(\Aset) := 1-\frac{\E\|z_{\Abar}-\hat z_{\Abar}\|^2}{\E\|z_{\Abar}\|^2}
  = \frac{1}{N-k}\sum_{j\in\Abar}R^2(j|\Aset) = \frac{\FEV(\Aset)-k}{N-k}.
  \label{eq:identity}
\end{equation}
\end{proposition}

The identity is well known: \citet{DasKem11} maximize one of its terms, $R^2$ of a single target, and we include a short proof in \cref{app:proofs} for completeness.
Only means and covariances enter it.
Evaluated at~$\Rh$, $R^2_{\mathrm{pop}}$ is the plug-in $R^2$, so greedy explained variance maximizes the plug-in $R^2$ one benchmark at a time.
It should do best where $\Rh$ is accurate; where it is not, the plug-in value can overstate the \emph{realized} $R^2$ measured on validation models (\cref{sec:results}).

\paragraph{A pairwise approximation: facility location.}
Explained variance has two drawbacks.
It is not submodular, because a benchmark that explains little alone can explain more next to another (\cref{app:proofs}), so greedy selection has only a data-dependent guarantee; and its gains need the inverse $\Rh_{\Aset\Aset}^{-1}$, which amplifies the error of correlations estimated from few or partly observed models.
Both drawbacks disappear if each benchmark is credited only with what its best single proxy in~$\Aset$ explains.
With squared correlations as similarities, this is \emph{facility location}~\citep{CorFisNem77}, named after the problem of opening a few facilities so that every customer is close to one; here the benchmarks are the customers and the selected benchmarks the facilities. It also turns out to be a (mathematically more amenable) lower bound on explained variance.

\begin{proposition}[A pairwise lower bound on explained variance]
\label{prop:lower}
Facility location with squared similarities, $\FFL^{\mathrm{sq}}(\Aset):=\sum_{j=1}^N\max_{i\in\Aset}R_{ij}^2$, satisfies $\FFL^{\mathrm{sq}}\le\FEV$, with equality if $|\Aset|\le1$.
\end{proposition}

A joint regression explains at least what its best single member explains; the difference is what proxies add in combination.
Facility location is monotone and submodular, since a new proxy adds only where it beats the best one already selected, and its gains need only pairwise correlations.
\citet{Yao26coresets} use it to select benchmarks by their score correlations, with the clipped similarity $\rplus_{ij}=\max(R_{ij},0)$, and we adopt their objective $\FFL(\Aset)=\sum_{j=1}^N c_j(\Aset)$, where the \emph{cover} $c_j(\Aset)=\max_{i\in\Aset}\rplus_{ij}$ (zero for $\Aset=\emptyset$) is the clipped correlation of~$j$ with its best proxy.
We keep their similarity, fixed before any experiment, to test their method as published; while it gives no bound on explained variance (unlike squared similarities), in practice it is essentially indistinguishable from squared similarities. See \cref{app:checks-similarity} for a comparison of the two scores. 

\paragraph{Information-theoretic objectives.}
Sensor placement~\citep{GueKraSin05,krause2008near}, where sensors play the role of benchmarks and a kernel gives the covariance that we estimate from past models, measures $S$ by its determinant instead.
For Gaussian scores and fixed~$k$, maximizing the \emph{entropy} of the selected scores, a measure of how uncertain they are jointly, amounts to maximizing $\FH(\Aset)=\log\det R_{\Aset\Aset}=\log\det R-\log\det S$.
The \emph{mutual information} $\FMI(\Aset)=\tfrac12\log\bigl(\det R_{\Aset\Aset}\det R_{\Abar\Abar}/\det R\bigr)=\tfrac12\log\det R_{\Abar\Abar}-\tfrac12\log\det S$ measures how much the selected scores tell about the unselected ones.
Both are submodular.
Explained variance measures how much residual variance a selection removes; entropy and MI measure only the factor by which it shrinks, so halving a tiny residual variance counts as much as halving a large one, although only the large one matters for squared error.

To make this more concrete, consider three near-duplicate chemistry subjects, two benchmarks that correlate weakly with them, a slightly noisy average of these two, and an \emph{outlier} that correlates with no other benchmark.
The coverage objectives start with a chemistry subject, which stands in for its two twins, and then add the average, which stands in for both of its parts; entropy makes the same first pick (by the tie rule below) but then adds the outlier, which the selection predicts worst yet which stands in for no benchmark but itself.
MI starts with the average, which the other benchmarks predict best jointly, and its third pick is a second chemistry subject, redundant with the first but well predicted by the third.
In the first step MI ranks~$j$ by $\Rsqrest{j}=R^2(j\,|\,[N]\setminus\{j\})$, how well the others jointly predict~$j$, quite different from the coverage objectives (\cref{tab:objectives}).

\begin{table}[t]
\caption{Selection objectives: coverage of every benchmark (facility location, FL; explained variance, EV) or information in the selected scores (entropy; mutual information, MI).
Only the coverage objectives score a benchmark by what it predicts.
$c_i$: cover; $\sigma^2_{j|\Aset}$: unexplained variance; $\Rsqrest{j}$: variance of~$j$ that all others explain; $\gamma$: submodularity ratio (\cref{prop:guarantees}).}
\label{tab:objectives}
\centering
\small
\setlength{\tabcolsep}{5pt}
\begin{tabular}{@{}llll@{}}
\toprule
 & First-step gain of $j$ & Greedy then adds the benchmark that \ldots & Greedy guarantee\\
\midrule
FL & $1+\sum_{i\neq j}\rplus_{ij}$ & most raises the covers $c_i(\Aset)$ & $1-1/e$, knapsack $\tfrac12(1-1/e)$\\
EV & $1+\sum_{i\neq j}R_{ij}^2$ & explains most remaining variance jointly & $1-e^{-\gamma}$ \\
\midrule
Entropy & $0$ for every $j$ (tie) & $\Aset$ predicts worst (largest $\sigma^2_{j|\Aset}$) & only after an arbitrary shift\\
MI & $-\tfrac12\log\bigl(1-\Rsqrest{j}\bigr)$ & $\Aset$ predicts poorly, the rest predict well & $e^{-1}$ for random greedy\\
\bottomrule
\end{tabular}
\end{table}

\begin{proposition}[Greedy guarantees]
\label{prop:guarantees}
Let $\Aset_G$ be the greedy set of size~$k$, and $\OPT_k$ the best value of the same objective over sets of at most $k$ benchmarks.
\begin{enumerate*}
  \item As $\rplus\ge0$, $\FFL$ is normalized, monotone and submodular, so $\FFL(\Aset_G)\ge(1-1/e)\,\OPT_k$ \citep{nemhauser1978analysis}.
  \item Under costs $\kappa_j>0$ and a budget~$C$, \emph{knapsack selection} returns the better of cost-effective greedy (add the affordable benchmark with the largest gain per unit cost) and the best affordable single benchmark; for~$\FFL$ it reaches $\tfrac12(1-1/e)$ of the best affordable set \citep{KraGue05b}.
  \item $\FEV$ is normalized and monotone but not submodular; $\FEV(\Aset_G)\ge(1-e^{-\gamma})\,\OPT_k$ with a submodularity ratio $\gamma\in[0,1]$, which measures how far $\FEV$ is from submodular ($\gamma=1$ if it is) and depends on $R$ and~$k$ \citep{DasKem11,Bian17nonsubmod}.
  \item $\FMI\ge0$ is submodular and symmetric, $\FMI(\Aset)=\FMI(\Abar)$, hence not monotone unless it vanishes: deterministic greedy has no known constant-factor guarantee, only a bound up to an additive loss \citep{krause2008near}, while Random Greedy, a randomized variant, achieves $\E\,\FMI(\Aset)\ge\OPT_k/e$ \citep{BucFelNaoSch14}.
  \item $\FH$ is submodular, but its gains $\log\sigma^2_{j|\Aset}$ are at most~$0$, so a bound needs an arbitrary shift $\FH(\Aset)+\delta|\Aset|$ and weakens as $\delta$ grows.
\end{enumerate*}
\end{proposition}
Of the deterministic methods, only facility location thus has a guarantee that does not depend on the data, with or without a budget; \cref{app:proofs} checks that these known results apply here and that greedy comes much closer to the optimum on our data.
Entropy's first step is an exact tie ($\sigma^2_{j|\emptyset}=1$ for every~$j$); all selectors break near-ties by the explained-variance gain (per unit cost under a budget), then by index (\cref{app:protocol-rules}), which keeps the guarantees up to a negligible term.
Note that all guarantees concern the objective on the estimate $\Rh$, not the realized $R^2$.

\paragraph{Baselines.}
Three simpler heuristics serve as baselines.
\emph{Hub} takes the $k$ benchmarks with the largest mean absolute correlation $h_j=(N-1)^{-1}\sum_{i\neq j}|\Rh_{ij}|$, ranked once rather than greedily, so it ignores redundancy and would take both near-duplicates; its first pick is usually that of facility location.
\emph{PCA} takes the $k$ benchmarks that best span the top $k$ principal components of~$\Rh$, by column-pivoted QR~\citep{BusGol65}, and \emph{random} takes $k$ benchmarks uniformly at random.

\section{Experimental protocol}
\label{sec:protocol}

Our experiments answer three questions: which objective best predicts the scores of validation models (\cref{sec:results}), why the information objectives fall behind (\cref{sec:mechanism}), whether the selected benchmarks still work for models released after the selection, and whether selection works when benchmarks differ in cost (\cref{sec:robust}).
Good selectors often differ by less than the split-to-split variation of one selector's $R^2$, so every comparison is paired.

\paragraph{Score matrices.}
We call the three matrices of \cref{fig:spectrum} \MMLU{} (fully observed), \MTEB{} (\nv[1]{DataMtebObsPct}\% observed) and \Merged{} (\nv[1]{DataMergedObsPct}\% observed).
With \BenchPress~\citep{Papailiopoulos26benchpress,Zeng26benchpress}, which compiles the scores of \nv{DataBenchpressM} models on \nv{DataBenchpressN} benchmarks from papers and model cards (\nv[1]{DataBenchpressObsPct}\% observed), they are our four main matrices, from dense to sparse.
We call \BenchPress{} \emph{estimation-limited}: too few models share scores to estimate its correlations well.
Seven extended matrices test generality: the sub-tasks of the Open LLM Leaderboard v2 (\OLLtwo) \citep{OpenLLMLeaderboard2}, the scores of \nv{DataOlvOneItemsM} models from its first version on \nv{DataOlvOneItemsN} benchmarks, with public per-question results (\OLLone) \citep{Polo24tiny}, Epoch AI's benchmark data (\Epoch) \citep{EpochBenchmarks26}, three \textsc{HELM} suites (\HELMLite, \HELMMMLU, \HELMClassic) \citep{Liang23}, and a 10/2026 version of \MTEB{} that adds the models submitted after our 3/2026 snapshot (\MTEBnew) \citep{MTEBResults26}.
\cref{app:data} provides details. 

\paragraph{Matched splits.}
Five random partitions of the models into ten folds give 50 splits per matrix; in each, one fold supplies the validation models and the other nine the training models.
All methods share each split's training models, correlation estimate~$\Rh$, candidate benchmarks and random numbers, so their $R^2$ values pair split by split.
$\Rh$ is the pairwise correlation matrix if the training matrix is at least 90\% observed and has at least as many models as benchmarks, and an EM estimate otherwise.
EM assumes that scores are missing at random~\citep{Rubin76} (\cref{app:protocol-em}).
A benchmark can be selected if at least three training models observe it (it is \emph{eligible}) and serves as a target only if at least two do; we report $k\le\min(15,N-3)$.
On sparse matrices a validation model is predicted from the selected scores it has and skipped if it has none, so \eqref{eq:r2} measures prediction from the scores a leaderboard already holds, not deployment, where a new model would run every selected benchmark.

\paragraph{Paired statistics.}
We compare methods $a$ and~$b$ through the per-split differences $d_s=R^2_{a,s}-R^2_{b,s}$ on the $n\le50$ splits where both are defined.
Overlapping training sets correlate the~$d_s$, so we use the 95\% interval of \citet{NadBen03} for repeated cross-validation \citep{BouFra04}, $\bar d \pm t_{0.975,\,n-1}\,\hat\sigma_d\sqrt{1/n + n_{\mathrm{val}}/n_{\mathrm{tr}}}$ with $\hat\sigma_d$ the standard deviation of the~$d_s$, which at $n=50$ and $n_{\mathrm{val}}/n_{\mathrm{tr}}\approx1/9$ is $2.6$ times as wide as the naive interval.
A difference is \emph{significant} if its interval excludes~$0$; we test each matrix and $k$ separately, without correction for multiple comparisons.

\section{Coverage objectives predict best}
\label{sec:results}

\begin{figure}[t]
\centering
\includegraphics[width=\linewidth]{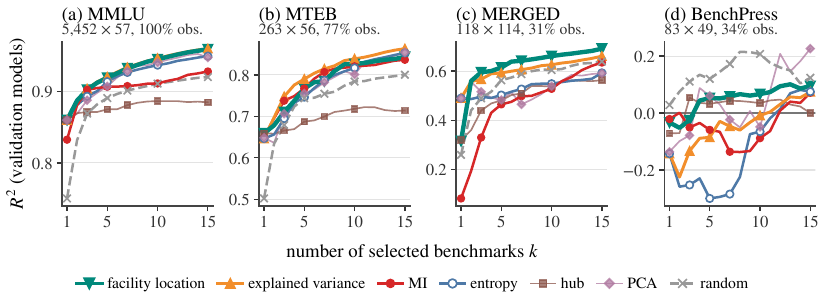}
\caption{The coverage objectives predict best wherever the correlations can be estimated (a--c); on the estimation-limited \BenchPress{} (d), no method significantly beats random.
Mean $R^2$ on the validation models over the 50 matched splits (fewer where no validation model observes a pick; $^{\dagger}$ in \cref{tab:main}); significance in \cref{tab:main}, intervals at $k=5$ in \cref{tab:app-results}.}
\label{fig:curves}
\end{figure}

The two coverage objectives predict the unselected benchmarks best, and they are practically equal (\cref{tab:main,fig:curves}).
Over all \nv{ScoreNMatrices} matrices (bottom of \cref{tab:main}), facility location is never significantly worse than MI, entropy, hub, PCA or random, and it is significantly better than each of them in \nv{ScoreFaclocVsMiSigBetter} to \nv{ScoreFaclocVsHubSigBetter} of \nv{ScoreNCells} settings of matrix and~$k$ ($k\in\{3,5,10\}$, $k\le N-3$).
We leave out $k=1$, where several selectors make the same pick: facility location usually that of hub, and entropy by construction that of explained variance.
The matrices overlap (\MTEBnew{} extends \MTEB, and three share the 57 \MMLU{} subjects), so these are not independent confirmations.
Facility location is never significantly better than explained variance and is significantly worse in \nv{ScoreFaclocVsTraceSigWorse} settings, all on the dense \MMLU, \OLLtwo{} and \OLLone{} matrices, by at most \nv{ScoreFaclocVsTraceMaxDeficit}.
The seven extended matrices follow the pattern of the main ones, and over every~$k$ the picture is the same (\cref{tab:app-results} in \cref{app:results}).
Selection pays most for small~$k$: on \MTEB, facility location leads random by \nv{PairDiffBaseMtebFaclocVsRandomKone} at $k=1$ but by only \nv{PairDiffBaseMtebFaclocVsRandomKfive}, not significantly, at $k=5$.

\paragraph{On the dense \MMLU, selections can be read by hand.}
Every selector except hub predicts the unselected subjects well, and explained variance leads facility location by a consistent but negligible margin (at $k=5$ in \nv{PairWinsBaseMmluTraceVsFaclocKfive} of \nv{PairNBaseMmluTraceVsFaclocKfive} splits, by \nv{PairDiffBaseMmluTraceVsFaclocKfive}).
Facility location picks the same five subjects on all \nv{SelMmluFaclocKfiveSameSplits} splits: \nvtext{SelMmluFaclocKfivePickOne} and \nvtext{SelMmluFaclocKfivePickThree}, the best proxies of \nv{CoverMmluFaclocKfivePickOne} and \nv{CoverMmluFaclocKfivePickThree} of the other subjects; \nvtext{SelMmluFaclocKfivePickTwo}, the best proxy of \nv{CoverMmluFaclocKfivePickTwo}; and \nvtext{SelMmluFaclocKfivePickFour} and \nvtext{SelMmluFaclocKfivePickFive}, each the best proxy of no other subject, picked because the earlier picks cover them worst.
Explained variance picks \nv{SelMmluTraceKfiveSharedWithFacloc} of these subjects as well.
Random selection is close behind because the scores are nearly low-rank (\cref{fig:spectrum}): most random sets of five capture the few dominant directions.
Hub falls behind as $k$ grows because it keeps adding subjects that its earlier picks already cover.

\paragraph{On sparser matrices, the two coverage objectives stay tied.}
On \MTEB{} explained variance has the higher mean for every $k\ge2$, and on \Merged{} facility location does, but neither lead is significant at any~$k$; facility location ties MI on \MTEB{} and is significantly ahead of it on \Merged{} at $k=5$ and~$10$ (\cref{tab:main}).
Where $\Rh$ is noisy, the two pick different benchmarks of similar quality: on \Merged{} they share only \nvpct{ObjFlVsTraceSharedMergedKfive} of their picks at $k=5$, against \nvpct{ObjFlVsTraceSharedMmluKfive} on \MMLU.

\paragraph{Where correlations cannot be estimated, no selector helps.}
On \BenchPress{} no selector is significantly better than random at any~$k$; random even has the highest mean at $k=5$ and~$10$, though not significantly.
The limit is the correlation estimate, not the selector: at $k=5$ the plug-in $R^2$ of explained variance (\cref{prop:identity}) is \nv{ObjPopRsqBenchpressTraceKfive} here but its realized $R^2$ only \nv{RsqBaseBenchpressTraceKfive}, while on \MMLU{} the two agree (\nv{ObjPopRsqMmluTraceKfive} and \nv{RsqBaseMmluTraceKfive}), so $\Rh$ cannot tell good proxies from poor ones (\cref{app:mechanism}).
With a tenth of the \Merged{} models for training instead of nine tenths, every selector we ran falls to the level of random there too (\cref{app:results}).

\paragraph{The information objectives fall behind.}
MI and entropy are never significantly better than facility location and are significantly worse in about half of the settings, also on the fully observed \MMLU.
An earlier version recommended MI because it beat entropy at small~$k$; that comparison rested on entropy's first pick, an exact tie broken by rounding noise, and does not survive (\cref{app:changes}).

\begin{table}[t]
\caption{Facility location (FL) is never significantly worse than the information objectives or the baselines, and practically equal to explained variance (EV).
Top: mean $R^2$ over the 50 matched splits on the main matrices (others in \cref{app:results}); bottom: settings over all \nv{ScoreNMatrices} matrices ($k\in\{3,5,10\}$, $k\le N-3$) in which FL is significantly better, not different, or worse.
$^{\triangledown}$/$^{\blacktriangle}$: significantly below/above FL (paired 95\% interval, uncorrected); bold: best in the row; $^{\dagger}$: mean over the splits where a validation model observes the pick (at least \nv{PairNBaseBenchpressFaclocVsRandomKone} of 50).}
\smallskip
\label{tab:main}
\centering
\small
\setlength{\tabcolsep}{4.5pt}
\begin{tabular}{@{}l r r r r r r r r@{}}
\toprule
Matrix & $k$ & FL & EV & MI & Entropy & Hub & PCA & Random\\
\midrule
\csname @@input\endcsname cr/tables/tab_main_body.tex
\bottomrule
\end{tabular}
\end{table}

\section{Why coverage beats information}
\label{sec:mechanism}

\begin{figure}[t]
\centering
\includegraphics[width=\linewidth]{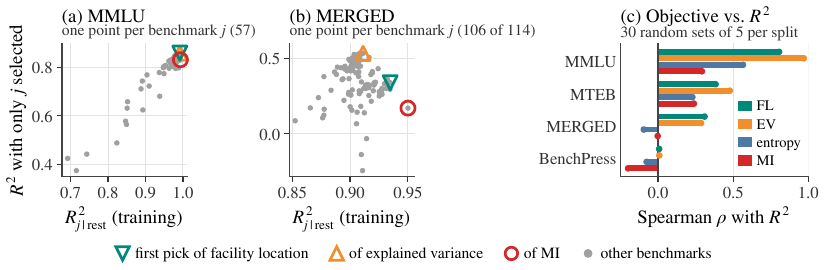}
\caption{The information objectives choose poor proxies.
(a,~b)~One point per benchmark: $\Rsqrest{j}$ under~$\Rh$, which drives MI's first pick, against the realized $R^2$ when $j$ alone is selected (means over the 50 splits); a good proxy sits high, and markers show each method's most frequent first pick (on \Merged, MI's lies far right but low; (b) omits \nv{FigMechMergedNOmitted} columns that no validation model has run when they can be selected).
(c)~Spearman correlation of each objective with realized $R^2$ over 30 random sets of five per split: the coverage objectives track the metric best, except on \BenchPress, where none does.}
\label{fig:mechanism}
\end{figure}

The coverage objectives win because they reward what the metric rewards: good predictions of the left-out benchmarks; the information objectives behave as in the toy example of \cref{sec:objectives} (\cref{app:mechanism}).
To see how well each objective guides selection, we draw random sets of five benchmarks and correlate each objective's value with the $R^2$ that the set actually reaches (\cref{fig:mechanism}c).
On \MMLU{} both coverage objectives rank the sets as the metric does; on \Merged, where every objective is a weak guide, they track the realized $R^2$, while the information objectives do not.

\paragraph{Entropy rewards novelty, not coverage.}
After its tie-broken first pick, entropy adds the benchmark with the largest conditional variance~$\sigma^2_{j|\Aset}$, the one that the selected benchmarks predict worst.
Such a benchmark is unlike the selected ones, but nothing makes it a good proxy for the rest.
Its sets are diverse (mean $|\Rh_{ij}|$ within its five picks \nv[2]{ObjRedundMmluEntropyKfive} on \MMLU, against \nv[2]{ObjRedundMmluFaclocKfive} for facility location), but diversity within the selected set is not coverage of the rest.

\paragraph{MI rewards the predicted, not the predictor.}
MI's first pick maximizes $-\tfrac12\log(1-\Rsqrest{j})$: it is the benchmark that all the others predict best jointly.
Like the average in the toy suite, such a benchmark can be predicted well by the others whether or not it stands in for them.
On \Merged, MI's usual first pick (\nv{SelMergedMiPtenFirstFolds} of 50 splits) is the overall score of Arena-Hard-Auto~\citep{ArenaHard24}, a chat benchmark judged by a language model.
It has $\Rsqrest{j}=\nv[2]{ObjModalFirstMergedMiRestRsq}$ under~$\Rh$, yet on its own it predicts the other benchmarks with $R^2=\nv{ObjModalFirstMergedMiSingleRsq}$, against \nv{ObjModalFirstMergedTraceSingleRsq} for \MMLU's astronomy subject, the usual first pick of explained variance (\cref{fig:mechanism}b).
The score is also fragile: it equals $\tfrac12\log P_{jj}$ for the precision matrix $P=\Rh^{-1}$, and $P_{jj}=\sum_i v_{ij}^2/\lambda_i$, summed over the eigenvalues $\lambda_i$ and eigenvectors $v_i$ of~$\Rh$, is dominated by the smallest eigenvalues, which few, partly observed models estimate worst.
On the sparser matrices, MI's first pick draws nearly all of $P_{jj}$ from the smallest tenth of the eigenvalues, and that of explained variance much less (\cref{app:mechanism}).

\paragraph{Neither the optimizer nor estimation noise alone explains the gap.}
If greedy were merely a poor optimizer of MI, Random Greedy, which unlike deterministic greedy has a constant-factor guarantee for MI (\cref{prop:guarantees}), should help; yet it reaches lower MI values and predicts no better (\cref{app:proofs}).
If noisy correlations alone were to blame, the information objectives should catch up on the fully observed \MMLU; yet they trail facility location significantly at $k=5$ and~$10$ (\cref{tab:main}).

\paragraph{Where facility location falls short.}
Facility location has two weak spots. Its first-step gain, one plus the sum of positive correlations, lets many weak correlations outweigh a few strong ones: on \Merged{} its first pick reaches $R^2=\nv{RsqBaseMergedFaclocKone}$, against \nv{RsqBaseMergedTraceKone} for explained variance, a large gap that is nonetheless not significant (\cref{tab:main}); with squared similarities, facility location makes the first pick of explained variance (\cref{app:checks-similarity}).
And once most benchmarks are well covered, two candidates can tie exactly, for example two benchmarks that are each other's best proxy (\cref{app:mechanism}); the explained-variance gain, which uses $\Rh_{\Aset\Aset}^{-1}$, then breaks the tie, among the first ten picks in \nv{TieFaclocSplitsMergedKten} of 50 splits on \Merged{} and \nv{TieFaclocSplitsMtebKten} on \MTEB, so late picks partly follow explained variance.

\section{New models and real budgets}
\label{sec:robust}

A selection is meant for models that appear after it is made, so we test whether it stays useful as models improve over time.

\paragraph{New models.}
\label{sec:robust-new}
We split three matrices by date: \emph{old models}, which appeared before a cutoff, estimate the correlations and choose the benchmarks, and \emph{new models}, which appeared after it, are predicted.
\HELMLite{} goes from release v1.0 to later releases, \MTEB{} from our snapshot to later models (\MTEBnew), and \OLLtwo{} from submissions before 2025 to later ones, without its mathematics sub-tasks, whose scorer changed.
Random splits of the same sizes give a control without a time shift; each method is scored on the new models that observe one of its picks, and differences are tested by bootstrapping over them (\cref{app:protocol-stats}).

\begin{figure}[t]
\centering
\includegraphics[width=\linewidth]{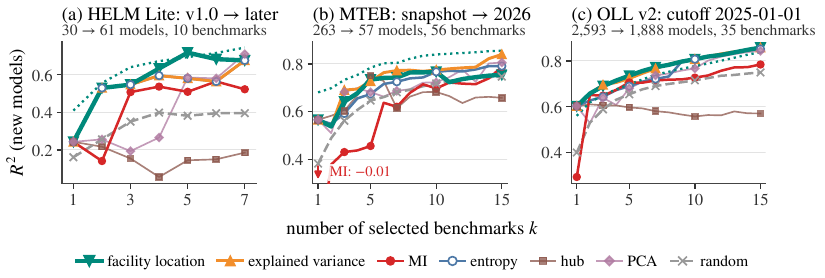}
\caption{Facility location stays at or near the top on models released after the selection, and MI falls behind.
$R^2$ on new models (one split per panel); dotted: facility location on random splits of the same sizes; more in \cref{app:newmodels-prosp}.}
\label{fig:prospective}
\end{figure}

With five benchmarks, facility location has the highest $R^2$ on the new \HELMLite{} and \OLLtwo{} models and is close to the best on \MTEBnew, while MI falls behind on all three (\cref{fig:prospective}).
On \MTEBnew{} hub has the highest $R^2$, but only \nv[0]{NModelsProspMtebHubKfive} of the \nv{NTestProspMteb} new models observe one of its picks, against \nv[0]{NModelsProspMtebFaclocKfive} for facility location and explained variance.
New models can also score higher or lower than old ones across the board, and \eqref{eq:r2}, which measures errors against the old models' mean, partly credits a selection for tracking such a shift; against the new models' own mean, hub's lead disappears (\cref{app:newmodels-prosp}).
Neither coverage objective dominates: with three benchmarks, explained variance is significantly ahead on \OLLtwo{} and \MTEBnew, and with five at the two latest cutoffs of a fourth time split, \Epoch, where new models score far above old ones (\cref{app:newmodels-prosp}).

\paragraph{Budgets.}
\label{sec:robust-cost}
Benchmarks differ widely in cost (an \MMLU{} subject has between \nv{CostMinMmlu} and \nv{CostMaxMmlu} questions), so a fixed number of benchmarks is the wrong budget.
Knapsack selection (\cref{prop:guarantees}b) prices each benchmark by its cost, which matters on \MMLU: at 1\% of the suite's questions, knapsack facility location reaches $R^2=\nv{CostRsqMmluKnapFaclocBone}$, against \nv{CostRsqMmluCardFaclocBone} for cost-blind facility location (its greedy order, cut at the budget) and \nv{CostRsqMmluRandomBone} for random (\cref{fig:budgets} in \cref{app:costs}).
Knapsack explained variance is significantly ahead of knapsack facility location at \nv{CostFixedSigPosMmluKnapTraceVsKnapFaclocFzero} of 7 \MMLU{} budgets, by at most \nv{CostFixedPairDiffMmluKnapTraceVsKnapFaclocFzeroBfive}, and taking the cheapest subjects first is a strong baseline there (\cref{app:costs-budgets}); priced by cost, facility location thus keeps nearly all of what explained variance achieves, with a constant-factor guarantee that we cannot give for explained variance.
For the same number of questions per model, whole benchmarks chosen by knapsack facility location predict all scores significantly better than questions drawn uniformly, from \nv[0]{ItemBudgetFixedCrossQSigStayOlvOneItemsBenchFaclocVsItemsCenterCzero} questions per model on (\OLLone{}; \cref{app:costs-items}).

\paragraph{Further checks.}
\label{sec:robust-other}
Facility location and explained variance remain the two best methods on \MMLU, \MTEB{} and \Merged{} when each family of related models is held out in turn, though a held-out family is usually predicted less well (\cref{app:newmodels-family}); a selection made without a group of benchmarks, such as one type of \MTEB{} task, predicts that group no better than random, since coverage extends only to the benchmarks one asks to cover (\cref{app:checks-benchfam}).
On the sparse matrices, subsampled training models and other correlation estimators leave facility location's $R^2$ unchanged, and a benchmark inflated by simulated contamination is never among the first five picks of a coverage objective (\cref{app:checks-stability,app:checks-contam,app:checks-estimator}).

\section{Related work}
\label{sec:related}

\paragraph{Fewer items or fewer benchmarks.}
Most work on cheaper evaluation keeps a few items (questions) per benchmark, chosen by model confidences \citep[Anchor Points;][]{Vivek24anchor}, by item response theory \citep[IRT;][]{Polo24tiny,Kipnis25metabench,Spagliardi26irtsafety}, adaptively \citep{Hofmann25fluid,Truong25amortized}, or for reliability \citep{Perlitz24efficient}; random items plus regression outperform most variants of the first two, which degrade when new models outperform old ones \citep{ZhangDornerHardt25}.
Across benchmarks, scores are low-dimensional \citep{Burnell23structure,RuanMaddisonHashimoto24,Polo25sloth}, so low-rank completion predicts missing scores from a few probe benchmarks, which \citet{Zeng26benchpress}, in the paper behind \BenchPress, choose greedily by the completion error of their imputer; other selectors cluster benchmark embeddings \citep{Ye23smallbench}, enumerate predictive subsets \citep{Surkov24vygotsky}, or search for representative subsets \citep{Subramani25simba}.
We instead derive the objective from the metric of a fixed predictor; for our linear predictor, the rule of \citeauthor{Zeng26benchpress} becomes greedy explained variance.

\paragraph{Facility location and experimental design.}
Closest to us, \citet{Yao26coresets} select prompts by facility location on text embeddings and, in our benchmark-level setting, found facility location on clipped score correlations to match or beat their reimplementation of our earlier entropy and MI for most~$k$ on \MMLU{} and \MTEB; we adopt their objective, relate it to the metric, compare it with explained variance on many more matrices, and explain why the information objectives fall behind.
Measured on the residual covariance~$S$, entropy and explained variance are the D- and A-optimality criteria of Bayesian experimental design \citep{Pukelsheim06}; greedy column subset selection maximizes a sample version of explained variance \citep{Farahat11greedycss}, and sensor placement also uses variance reduction \citep{krause2008near,KraMcmGueGup08}.

\section{Discussion and limitations}
\label{sec:discussion}

Benchmark scores are redundant enough that a few benchmarks predict the rest, and the objective that chooses them should follow from how the predictions are judged: coverage, not information.
Explained variance is the metric on the estimated correlations; in cross-validation it never trails facility location significantly and leads it significantly only in a few dense settings, by a few thousandths of $R^2$, and on later models neither dominates.
In exchange, facility location offers a data-independent guarantee, also under a budget, and readable selections, so we recommend it, with knapsack selection when costs differ.
Selection pays most when few benchmarks fit the budget; with many, a random set is a reasonable fallback on dense matrices, and where the correlations cannot be estimated, the choice of selector matters little.
For benchmark designers, a new benchmark adds information when the existing ones jointly predict it poorly, as \citet{Harfi26proactbench} report for a new evaluation dimension.

\paragraph{Limitations.}
Our predictor is linear, though it needs no normality (\cref{app:protocol-normality}), and the correlation estimate assumes that scores are missing at random, which is unlikely, since which models run which benchmarks depends on age, provider and popularity.
The guarantees bound the objective on the estimated correlations, not the accuracy of predictions; the evidence on models released later rests on one split per setting, unseen model families cost accuracy, and how fast a selection goes stale remains open.
We did not model human-evaluation cost or test item selectors based on item response theory~\citep{Polo24tiny}; contamination was simulated.

\paragraph{Broader impact.}
Pruning benchmarks could lead to overfitting on a narrowly selected set, rather than improving system performance. Seeding the selection with mandatory benchmarks, which keeps the facility-location gain monotone and submodular, and periodic full evaluations reduce these risks.

\needspace{5\baselineskip}
\begin{ack}
This work received no third-party funding. The author declares no competing interests.

\end{ack}

\bibliographystyle{plainnat}
\bibliography{bibfile,relwork,cr/refs_cr,cr/cr_additions}

\clearpage
\appendix
\crefalias{section}{appendix}
\crefalias{subsection}{subappendix}

\section{Proofs and guarantees}
\label{app:proofs}

This appendix supports \cref{sec:objectives}. \Cref{prop:identity,prop:lower} are simple facts about linear regression, and we include short proofs for completeness. \Cref{prop:guarantees} collects known guarantees; for each part we name its source, check its conditions for our objectives and tie rule, and report how close greedy comes to the optimum on our data. Throughout, $R$ is a positive definite $N \times N$ matrix with unit diagonal and $k = |\Aset|$, with $0 < k < N$ in \cref{prop:identity}. For a nonempty set $\Aset$, $R^2(j \mid \Aset) = R_{j\Aset} R_{\Aset\Aset}^{-1} R_{\Aset j}$ is the share of the variance of benchmark~$j$ that a linear regression on the benchmarks in $\Aset$ explains. It equals $1$ for $j \in \Aset$, and $R^2(j \mid \emptyset) = 0$.

\paragraph{Proof of \cref{prop:identity}.}
The identity is standard (cf.\ \citealp{DasKem11}, who maximize one of its summands, the $R^2$ of a single target). Let $z \in \R^N$ be the standardized scores of a new model, with mean $0$ and covariance matrix~$R$ (unit variances), and let $W = R_{\bar\Aset\Aset} R_{\Aset\Aset}^{-1}$. The prediction error $e = z_{\bar\Aset} - W z_\Aset$ has covariance
\[
  \E[e e^\top] = R_{\bar\Aset\bar\Aset} - W R_{\Aset\bar\Aset} - R_{\bar\Aset\Aset} W^\top + W R_{\Aset\Aset} W^\top
  = R_{\bar\Aset\bar\Aset} - R_{\bar\Aset\Aset} R_{\Aset\Aset}^{-1} R_{\Aset\bar\Aset}.
\]
Its diagonal entry for $j \in \bar\Aset$ is $1 - R^2(j \mid \Aset)$, so $\E\|e\|^2 = \sum_{j \in \bar\Aset} (1 - R^2(j \mid \Aset))$. The denominator is $\E\|z_{\bar\Aset}\|^2 = \tr R_{\bar\Aset\bar\Aset} = N - k$. Hence
\[
  R^2_{\mathrm{pop}}(\Aset) = 1 - \frac{\E\|e\|^2}{\E\|z_{\bar\Aset}\|^2}
  = \frac{1}{N-k} \sum_{j \in \bar\Aset} R^2(j \mid \Aset)
  = \frac{F_{\mathrm{EV}}(\Aset) - k}{N - k},
\]
where the last step uses $R^2(j \mid \Aset) = 1$ for the $k$ selected benchmarks. The same computation shows that explained variance is $N$ minus the total residual variance, $F_{\mathrm{EV}}(\Aset) = N - \tr(R_{\bar\Aset\bar\Aset} - R_{\bar\Aset\Aset} R_{\Aset\Aset}^{-1} R_{\Aset\bar\Aset})$. Only the mean and the covariance of $z$ entered the argument. \hfill$\square$

$R^2_{\mathrm{pop}}$ is the population version of the metric \eqref{eq:r2}: with every cell observed, a predictor built from~$R$ and no ridge or clipping, the numerator and the denominator of \eqref{eq:r2}, divided by the number of validation models, converge to $\E\|e\|^2$ and $\E\|z_{\bar\Aset}\|^2$. At fixed~$k$, $R^2_{\mathrm{pop}}$ is an increasing affine function of $F_{\mathrm{EV}}$, so each greedy step of explained variance on $\hat R$ maximizes the plug-in value of the metric; \cref{app:mechanism} compares that value with the realized~$R^2$.

\paragraph{Proof of \cref{prop:lower}.}
The quantity $R^2(j \mid \Aset)$ is the largest share of the variance of $z_j$ that any linear combination of $z_\Aset$ explains: $R^2(j \mid \Aset) = 1 - \min_w \E(z_j - w^\top z_\Aset)^2$. Restricting $w$ to a single benchmark $i \in \Aset$ gives $\min_a \E(z_j - a\, z_i)^2 = 1 - R_{ij}^2$. Hence $R^2(j \mid \Aset) \ge \max_{i \in \Aset} R_{ij}^2$ for every~$j$, and summing over $j$ gives $F^{\mathrm{sq}}_{\mathrm{FL}}(\Aset) \le F_{\mathrm{EV}}(\Aset)$. For $\Aset = \{i\}$ the regression has a single regressor, so $R^2(j \mid \{i\}) = R_{ij}^2$ and the two sides agree; for $\Aset = \emptyset$ both vanish. \hfill$\square$

In words, explained variance credits each benchmark with what the selected set explains jointly, and facility location with squared similarities credits it with what its best single proxy explains. The difference is what proxies add in combination. It vanishes whenever one proxy suffices.

\paragraph{\Cref{prop:guarantees}(a): facility location.}
Part~(a) is classical: \citet{CorFisNem77} analyzed greedy selection for facility location, and \citet{nemhauser1978analysis} proved the factor $1 - 1/e$ for every normalized monotone submodular function. We repeat the short argument, which also covers the other similarities of \cref{app:checks-similarity} and the restriction to eligible candidates. Fix a target benchmark~$j$ and let $c_j(\Aset) = \max_{i \in \Aset} s_{ij}$ be its cover, with similarities $s_{ij} \ge 0$ and $c_j(\emptyset) = 0$. Adding benchmark~$i$ raises the cover by $\max(s_{ij} - c_j(\Aset), 0)$. This gain is nonnegative, and it can only shrink as $\Aset$ grows, because $c_j(\Aset)$ can only grow. Each $c_j$ is therefore normalized, monotone and submodular, and so is their sum $F_{\mathrm{FL}} = \sum_j c_j$. This holds for $s = r^+$ and equally for the similarities $R_{ij}^2$ and $|R_{ij}|$ of \cref{app:checks-similarity}, so greedy satisfies $F_{\mathrm{FL}}(\Aset_G) \ge (1 - 1/e) \max_{|\Aset| \le k} F_{\mathrm{FL}}(\Aset)$. No polynomial-time algorithm improves this factor for facility location with arbitrary nonnegative similarities, which contains maximum coverage, unless P\,=\,NP \citep{Feige98}. Eligibility restricts the candidates, not the targets. $F_{\mathrm{FL}}$ restricted to subsets of the eligible benchmarks is still monotone submodular, so the bound holds against the best eligible set.

Both coverage objectives count each selected benchmark once for itself ($c_j(\Aset) = 1$ and $R^2(j \mid \Aset) = 1$ for $j \in \Aset$), so $F \ge k$, and a factor on~$F$ promises less for the unselected benchmarks, $F(\Aset) - k$. The brute-force comparison under (c) is the practical check.

\paragraph{Tie-breaking keeps the guarantees.}
The greedy bound needs only that each step takes a largest gain, so any way of breaking exact ties keeps it, and a step that is optimal up to a small additive error loses only that error. Our rule picks among candidates within $\tau_t = 10^{-9} \max(g_t^*, 1)$ of the largest gain~$g_t^*$, so the induction of \citet{nemhauser1978analysis} gives $F(\Aset_k) \ge (1 - 1/e)\,\mathrm{OPT} - \sum_t \tau_t$.
Facility-location gains are at most~$N$, so the loss is at most $kN \cdot 10^{-9} < 2 \times 10^{-6}$ on our matrices. The same argument covers the ratio ties of knapsack selection and the tied gains in Random Greedy.

\paragraph{\Cref{prop:guarantees}(b): knapsack selection.}
Part~(b) is the budgeted result of \citet{KraGue05b}; we describe the algorithm as we run it. Let $\kappa_j > 0$ be benchmark costs and $C$ the budget. Cost-effective greedy starts from the empty set and repeatedly adds the eligible benchmark that still fits the remaining budget and has the largest gain per unit cost. It stops when no benchmark fits or no gain is positive. The best singleton is the affordable eligible benchmark with the largest $F(\{j\})$. Knapsack selection returns the singleton if it beats the greedy set~$G$ by more than $10^{-9} \max(1, F(G))$, and $G$ otherwise. For a normalized monotone submodular $F$, the better of the two reaches at least $\tfrac12 (1 - 1/e) \approx 0.316$ times the value of the best affordable set; our tie rule lowers this bound by at most the $10^{-9}$ tolerances. The bound covers facility location only. Explained variance is not submodular, MI is not monotone and entropy needs a shift (below), so knapsack selection with these objectives is a heuristic. Cost-blind selection (\cref{app:protocol-designs}) has no guarantee, because an expensive early pick can use up most of the budget.

\paragraph{\Cref{prop:guarantees}(c): explained variance.}
Part~(c) combines a small counterexample with the data-dependent bound of \citet{DasKem11}. $F_{\mathrm{EV}}$ is normalized and monotone, because a regression on more benchmarks never explains less. It is not submodular: a benchmark that explains little alone can explain more next to another one. Consider five benchmarks built from independent standard normal variables $u, v, \epsilon_3, \epsilon_4, \epsilon_5$:
\[
  z_1 = u, \qquad z_2 = \rho u + \sqrt{1 - \rho^2}\, v, \qquad z_i = \beta v + \sqrt{1 - \beta^2}\, \epsilon_i \quad (i = 3, 4, 5),
\]
with $0 < \rho, \beta < 1$. Alone, benchmark~1 explains only benchmark~2, so its gain is $\Delta(1 \mid \emptyset) = 1 + \rho^2$. Next to benchmark~2 it lets the regression subtract $u$ from $z_2$ and recover~$v$, which predicts the three targets. Its gain becomes $\Delta(1 \mid \{2\}) = 1 - \rho^2 + 3\rho^2\beta^2$. The gain grows by $\rho^2 (3\beta^2 - 2)$, which is positive for $\beta^2 > 2/3$.

Our data show the same effect. Along the greedy path of explained variance, \nvpct{ObjTheoryMmluTraceIncreaseFrac} (\MMLU) to \nvpct{ObjTheoryBenchpressTraceIncreaseFrac} (\BenchPress) of the candidates' gains grow after a pick, $\Delta(j \mid \Aset_t) > \Delta(j \mid \Aset_{t-1})$; each such case violates diminishing returns.

Without submodularity the guarantee depends on the data: greedy reaches $F_{\mathrm{EV}}(\Aset_G) \ge (1 - e^{-\gamma}) \max_{|\Aset| \le k} F_{\mathrm{EV}}(\Aset)$, where the submodularity ratio
\[
  \gamma = \min_{L \subseteq \Aset_G,\; S \cap L = \emptyset,\; |S| \le k} \;
  \frac{\sum_{x \in S} \bigl[ F(L \cup \{x\}) - F(L) \bigr]}{F(L \cup S) - F(L)}
\]
measures how far $F$ is from submodular ($\gamma \le 1$ always, with equality for submodular~$F$). \citet{DasKem11} state the bound for the $R^2$ of one target, but the proof uses only monotonicity and~$\gamma$, so it carries over to the sum $F_{\mathrm{EV}}$ (see also \citealp{Bian17nonsubmod}). We computed $\gamma$ exactly at $k = 2$. Averaged over the 50 splits, it ranges from \nv[2]{ObjTheoryBenchpressTraceGamma} (\BenchPress) to \nv[2]{ObjTheoryMmluTraceGamma} (\MMLU), so at $k = 2$ the guarantee averages \nv[2]{ObjTheoryBenchpressTraceGammaBound} to \nv[2]{ObjTheoryMmluTraceGammaBound} of the optimum (as low as \nv[2]{ObjTheoryBenchpressTraceGammaBoundMin} on one \BenchPress{} split). In practice greedy comes much closer. Against the brute-force optimum over all sets of size $k = 3$, and averaged over splits, greedy explained variance reaches at least \nvpct[1]{ObjBruteMtebTraceKthreeFRatio} of the optimal value on each of the four main matrices (lowest on \MTEB), and facility location at least \nvpct[1]{ObjBruteMergedFaclocKthreeFRatio} (lowest on \Merged; \cref{tab:app-mech}).

\paragraph{\Cref{prop:guarantees}(d): mutual information.}
Part~(d) restates known properties of MI, which we then measure on our data. The objective $F_{\mathrm{MI}}(\Aset) = I(z_\Aset; z_{\bar\Aset}) \ge 0$ is submodular \citep{krause2008near} and symmetric, $F_{\mathrm{MI}}(\Aset) = F_{\mathrm{MI}}(\bar\Aset)$. In particular $F_{\mathrm{MI}}(\emptyset) = F_{\mathrm{MI}}([N]) = 0$, so unless it vanishes it must decrease somewhere. Two benchmarks with correlation $0.95$ already show this:
$F_{\mathrm{MI}}(\{1\}) = -\tfrac12 \log(1 - 0.95^2) \approx 1.164 > F_{\mathrm{MI}}(\{1, 2\}) = 0$. For non-monotone submodular functions, deterministic greedy has no constant-factor guarantee in general, and none is known for MI. \citet{krause2008near} bound it up to an additive loss: if no gain on a set of at most $2k$ benchmarks falls below $-\epsilon$, their bound implies that greedy reaches at least $(1 - 1/e)\,\mathrm{OPT}_k - k\epsilon$. On \MMLU{} and \MTEB{} this loss is not small: the most negative gain of any candidate along greedy MI's path (on sets of at most $2k$ benchmarks) puts $k\epsilon$ at no less than \nvpct{ObjTheoryMmluMiKepsRel} and \nvpct{ObjTheoryMtebMiKepsRel} of greedy's MI value at $k = 5$ (on \Merged{} and \BenchPress{} no such gain was negative).

Along its own path, greedy MI stays in the increasing region.
Its gain was positive at all \nv{RevMonoDatasetsMmluNSteps} greedy steps (50 splits, 15 picks) on each of the four main matrices, and never below \nv[2]{RevMonoDatasetsMmluMinGain} nats (\MMLU), whereas off the path some candidate had a negative gain in \nvpct{ObjTheoryMmluMiNonmonotone} of the \MMLU{} splits and \nvpct{ObjTheoryMtebMiNonmonotone} of the \MTEB{} splits. MI is monotone along the greedy path, not as a function.

Random Greedy restores a guarantee for every nonnegative submodular function: its set of at most $k$ benchmarks satisfies $\E F(\Aset) \ge \mathrm{OPT}_k/e$ \citep{BucFelNaoSch14}. We implement their Algorithm~1, which pads the candidates with $2k$ dummy benchmarks of zero gain and at each step draws uniformly among the $k$ candidates with the largest gains.
The guarantee holds for the $k$ the algorithm ran with, not for prefixes of its output, so we ran $k \in \{5, 10, 15\}$ separately with five seeds per split. At $k = 5$ Random Greedy reaches a lower MI value than deterministic greedy on all four main matrices, by \nv[2]{RevMonoDatasetsMergedPairedFiveMiMeanDiff} to \nv[2]{RevMonoDatasetsMtebPairedFiveMiMeanDiff} nats on average, with every paired interval above zero. It does not predict better. Deterministic greedy is ahead on \MMLU{} (\nv{RevMonoDatasetsMmluDeterministicFiveRTwoMean} against \nv{RevMonoDatasetsMmluRandomGreedyFiveRTwoMean}; difference \nvd{RevMonoDatasetsMmluPairedFiveRTwoMeanDiff} [\nvd{RevMonoDatasetsMmluPairedFiveRTwoCiLow}, \nvd{RevMonoDatasetsMmluPairedFiveRTwoCiHigh}]), and on \MTEB, \Merged{} and \BenchPress{} the paired intervals include zero. The guarantee does not buy better predictions.

\paragraph{\Cref{prop:guarantees}(e): entropy.}
Part~(e) is a simple remark on a known objective. $F_{\mathrm{H}}(\Aset) = \log\det R_{\Aset\Aset}$ is submodular, because its gain $\log \sigma^2_{j \mid \Aset}$ can only shrink as $\Aset$ grows: conditioning on more benchmarks never increases a variance \citep{GueKraSin05}. On a correlation matrix every conditional variance is at most~1, so every gain is at most~0. $F_{\mathrm{H}}$ never rises with a pick, and its first step is an exact tie. Over sets of at most $k$ benchmarks its optimum is therefore the empty set, and the bound of \citet{nemhauser1978analysis}, which needs a normalized monotone function, applies only to the shifted objective $F_{\mathrm{H}}(\Aset) + \delta|\Aset|$ for any $\delta \ge -\log \sigma^2_{\min}$, where $\sigma^2_{\min}$ is the smallest conditional variance of a benchmark given a set of fewer than $2k$ others. The shift does not change greedy's picks at fixed~$k$, and the bound becomes $\log\det R_{\Aset_G\Aset_G} \ge (1 - 1/e)\,\mathrm{OPT}_{\mathrm{H}} - \delta k/e$ with $\mathrm{OPT}_{\mathrm{H}} = \max_{|\Aset| = k} \log\det R_{\Aset\Aset}$, the best set of exactly $k$ benchmarks. Since $\delta$ grows without limit as a conditional variance approaches zero, and $\log\det R_{\Aset\Aset}$ is not the metric, this bound says nothing about prediction.

\section{Protocol details}
\label{app:protocol}

\subsection{Splits and random numbers}
\label{app:protocol-splits}

Repetition~$r$ of the cross-validation permutes the models with seed $42$ ($r = 0$) or $(42, r)$ and cuts the permutation into 10 folds whose sizes differ by at most one. Split $s = 10r + f$ validates on fold~$f$ and trains on $\min\{\mathrm{round}((1-p)M),\ \text{pool size}\}$ models drawn from the other nine folds, where the holdout fraction~$p$ is $0.1$ in the main results.
At $p = 0.1$ this is the whole pool, except where rounding leaves one pool model out (\Merged, \Epoch, \HELMMMLU{} and \HELMClassic, in 10, 15, 5 and 15 of the 50 splits). At $p = 0.9$ (\cref{app:results}) a random tenth of the models trains and the validation folds stay the same. Repetition~0 reproduces the folds of the submitted version. Every random stream is seeded by the split and by a tag of its own (\texttt{SEED\_TAGS} in \texttt{cv\_split.py}), never by the order in which methods run, so the random baseline draws the same permutation on a split in every experiment.

\subsection{Estimating the correlation matrix}
\label{app:protocol-em}

Each benchmark~$j$ is standardized with the mean and standard deviation of its $n_j$ observed training scores ($n_j - 1$ in the denominator).
The pairwise estimate sums the products of the standardized scores of two benchmarks over the $n_{ij}$ training models that observe both, divides by $\max(n_{ij} - 1, 1)$, clamps the eigenvalues at $10^{-6}$ and rescales the result to unit diagonal.

The EM estimate \citep{DempsterLairdRubin77} treats the standardized scores $z_m$ of the $n_{\mathrm{tr}}$ training models as draws from a Gaussian with mean~$\mu$ and covariance~$\Sigma$. For model~$m$ with observed benchmarks $O_m$ and missing benchmarks $U_m$, the E-step and the M-step are
\[
  \hat z_{m U_m} = \mu_{U_m} + \Sigma_{U_m O_m} \Sigma_{O_m O_m}^{-1} (z_{m O_m} - \mu_{O_m}), \quad
  C_m = \Sigma_{U_m U_m} - \Sigma_{U_m O_m} \Sigma_{O_m O_m}^{-1} \Sigma_{O_m U_m},
\]
\[
  \mu \leftarrow \frac{1}{n_{\mathrm{tr}}} \sum_m \tilde z_m, \qquad
  \Sigma \leftarrow \frac{1}{n_{\mathrm{tr}}} \sum_m \bigl[ (\tilde z_m - \mu)(\tilde z_m - \mu)^\top + C_m \bigr],
\]
where $\tilde z_m$ is row~$m$ completed by $\hat z_{m U_m}$ and $C_m$ fills the $U_m \times U_m$ block.
EM starts from the pairwise-complete covariance. After the start and after every M-step, eigenvalues are floored at~$\varepsilon$, and an E-step adds a ridge of $10^{-6}$ to $\Sigma_{O_m O_m}$ if it is numerically singular. EM is regularized ($\varepsilon = 10^{-3}$, at most 1{,}000 iterations) if $n_{\mathrm{tr}} < N$ or fewer than half of the training cells are observed, and unregularized otherwise ($\varepsilon = 10^{-6}$, at most 500 iterations). It stops when $\|\Sigma_{t+1} - \Sigma_t\|_F / \|\Sigma_t\|_F < 5 \times 10^{-4}$. If $n_{\mathrm{tr}} < N$, the start and the result are shrunk as $\Sigma \leftarrow (1-\alpha)\Sigma + \alpha\, (\tr\Sigma / N)\, I$ with the fixed intensity $\alpha = (N - n_{\mathrm{tr}})/N$; this is the form of the estimator of \citet{LedoitWolf04}, not its optimal intensity. The final $\Sigma$ is rescaled to a correlation matrix.
At $p = 0.1$, $\hat R$ is pairwise on \MMLU, \OLLtwo, \OLLone, \HELMLite{} and \HELMMMLU; unregularized EM on \MTEB, \MTEBnew{} and \HELMClassic; and regularized EM on \Merged{} ($\alpha = 0.07$), \BenchPress{} and \Epoch{} ($\alpha = 0$). \Cref{app:checks-estimator} shows how the results move when these settings change.

\subsection{Selection and scoring rules}
\label{app:protocol-rules}

\paragraph{Eligibility.}
An ineligible benchmark stays in every objective: as a target of facility location and explained variance, in the complement of MI, in the hub scores and in the PCA loadings. It also stays an imputation target if at least two training models observe it.
At $p = 0.1$ the eligibility rule binds on two matrices only. On \Merged{} it removes the eight MT-Bench columns, each observed for three models in total, in the 14 splits that leave one of these models out of training. On \BenchPress{} it always removes AIME~2026 and GDP-Val~AA, and it removes SciCode and MathVision in 14 and 3 splits.

\paragraph{Valid score ranges.}
Predictions are clipped to ranges that come from each metric's scale, never from observed scores: $[0, 1]$ for accuracies and other fractions (all of \MMLU, the Open LLM, \Epoch{} and HELM matrices, and most of \MTEB{} and \Merged), $[-1, 1]$ for the Spearman correlations of the \MTEB{} similarity and summarization tasks, $[0, 100]$ for percentages (win rates and pass rates in \Merged, most of \BenchPress), $[1, 10]$ for MT-Bench, $[-80, 100]$ for WildBench scores, and no bound for Elo-style ratings in \BenchPress. The clip acts in standardized units, so a prediction inside its range keeps its value; standardized inputs and true scores are also capped at $\pm 10$.

\paragraph{Near-ties.}
A score counts as tied with the best score~$s^*$ if it lies within $10^{-9} \max(|s^*|, s_0)$ of it, where $s_0$ is the scale below which rounding is absolute: $s_0 = 1$ for the gains of facility location, explained variance and MI, for entropy (whose test is on the conditional variance $\sigma^2_{j|\Aset}$ rather than on its gain $\log\sigma^2_{j|\Aset}$) and for hub scores; the largest initial loading norm for PCA (below); and $s_0 = 0$ (a purely relative test) for gains per unit cost. Ties go to the larger explained-variance gain, then to the lower index. Explained variance, whose score is that gain, goes straight to the index; hub uses the gain at the empty set, so that its ranking stays one-shot; knapsack selection uses the gain per unit cost. The rule matters in four places: entropy's first step (an exact tie, which hands entropy the first pick of explained variance), MI's picks among the near-identical MT-Bench columns of \Merged, facility location's late picks (\cref{app:mechanism}) and equal-cost benchmarks in cheapest-first selection.

\paragraph{PCA subset selection.}
For each~$k$, the Businger--Golub pivoted QR \citep{BusGol65} runs on the rows of the top-$k$ eigenvectors that belong to eligible benchmarks; each step pivots the benchmark whose loading vector has the largest norm after projecting out the earlier pivots. If $k$ cuts through a repeated eigenvalue, the whole eigenspace is used, because its basis is arbitrary, and directions without loading on any eligible benchmark go last. PCA is the only selector whose sets are not nested in~$k$.

\subsection{Statistics}
\label{app:protocol-stats}

In the Nadeau--Bengio interval of \cref{sec:protocol}, $n_{\mathrm{val}}/n_{\mathrm{tr}}$ is the ratio of the mean validation size to the mean training size over the splits used \citep{NadBen03,BouFra04}.
At $p = 0.9$ the ratio is about~1, so the interval is about $7.1$ times as wide as the naive one. Standard deviations over splits describe spread and are not tests.

A split's $R^2$ is undefined when no validation model observes a selected benchmark. This happens only on sparse matrices at small~$k$; for example, random is defined on \nv{PairNBaseBenchpressFaclocVsRandomKone} of the 50 \BenchPress{} splits at $k = 1$. Means skip undefined splits, and paired differences use the splits on which both methods are defined, so they can differ slightly from the difference of two means.

A prospective setting, which splits the models by release time (\cref{sec:robust-new}), has a single split.
Its interval comes from 2{,}000 resamples of the new models with replacement, the same resamples for every method: we recompute each method's pooled $R^2$ on every resample and report the 95\% percentile interval of the difference \citep{EfrTib93}. Random enters as its mean over 50 draws within each resample. The interval covers the sampling of new models, not the variability of the selection.

\subsection{Designs of the experiments on new models and costs}
\label{app:protocol-designs}

\paragraph{Prospective splits.}
A prospective split divides the models by release or submission date. Old models do everything that training models do in cross-validation: they fix the standardization, $\hat R$, the eligible set and every selection. New models are only imputed.
New-model rows that duplicate an old-model row are dropped (identical observed values or, on \OLLtwo, identical model weights). On \OLLtwo{} the seven MATH columns are dropped, because their scorer version changed during the leaderboard's life and is confounded with submission date; the ``+\,MATH'' variant keeps them. On \Epoch{} the benchmarks released on or after the cutoff are removed from both sides, so no score on a benchmark that did not exist at the cutoff enters the fit; the ``by model'' variant keeps them and therefore looks ahead. A new model enters a method's $R^2$ only if it observes one of the method's picks, so each method covers its own subset of the new models. The random-split control draws 50 random splits of the same models with the same training and test sizes; the prospective penalty is the prospective $R^2$ minus the control mean.

\paragraph{Model-family holdout.}
Families come from \texttt{model\_families.py}: base-architecture keywords in model names for \MMLU, \Merged, \BenchPress{} and the two Open LLM matrices; the Hugging Face organization for \MTEB{} and \MTEBnew; and the developer recorded in the metadata for \Epoch{} and the HELM matrices. Fine-tunes go with their base model's family where the name or the metadata reveals it, and models without a family always stay in training. Every family with at least 40 (\MMLU), 7 (\MTEB, \Merged), 6 (\BenchPress) or $\max(6, \lceil M/150 \rceil)$ models (other matrices) is held out in turn, and all other models train. Deterministic methods run once per family; random is averaged over five draws.

\paragraph{Budgets.}
Costs are the numbers of test questions of the \MMLU{} subjects and the numbers of evaluation samples of the \MTEB{} tasks (\cref{app:data}). A budget is a fraction of the suite's total cost, the same on every split. Knapsack selection is described in \cref{app:proofs}. Cost-blind selection fills the budget in the cardinality greedy order and skips benchmarks that no longer fit. Cheapest-first adds benchmarks by increasing cost. Random fills the budget from the split's permutation. \Cref{app:costs} adds a fixed overhead per benchmark.

\paragraph{Whole benchmarks versus random items.}
The \OLLone{} matrix comes with all 28{,}659 item scores of its \nv{DataOlvOneItemsM} models on \nv{DataOlvOneItemsN} benchmarks. Each new model may answer $Q$ questions. Whole-benchmark methods run knapsack selection with the number of items as cost and impute the rest. Random items take the first $Q$ items of one uniform permutation of all items per split, shared by all validation models and nested across~$Q$. Their estimator reads a benchmark's sample mean, centered by the training models' mean on the same items, as a noisy observation $y_t$ of its standardized score, with a noise variance estimated from the training models' item residuals (with finite-population correction). It imputes all standardized scores by the Gaussian posterior mean $\hat z = \hat R_{\cdot o} (\hat R_{oo} + D + \lambda I)^{-1} y_o$, where $o$ indexes the sampled benchmarks and $D$ holds their noise variances. $R^2$ is pooled over all \nv{DataOlvOneItemsN} benchmarks, and benchmarks run in full count exactly. \Cref{app:costs} adds an overhead per benchmark touched.

\subsection{Normality and determinism}
\label{app:protocol-normality}

The scores are not Gaussian. Shapiro--Wilk tests \citep{ShapiroWilk65} with a Benjamini--Hochberg correction at 5\% \citep{BenjaminiHochberg95} reject normality for \nv{NormMmluNRejectBh} of \nv{NormMmluN} \MMLU{} subjects and \nv{NormMtebNRejectBh} of \nv{NormMtebN} \MTEB{} tasks (median $W$ \nv[2]{NormMmluWMedian} and \nv[2]{NormMtebWMedian}), and for \nv{NormMergedNRejectBh} of \nv{NormMergedN} \Merged{} benchmarks, where fewer observations per column give the test less power. Mardia's tests of multivariate skewness and kurtosis \citep{Mardia70} also reject on \MMLU{} and on the \nv{NormMtebNComplete} complete rows of \MTEB{}; \Merged{} has no complete row. Benchmark scores are bounded and often discrete. The Gaussian model is a working model: its conditional-mean formula is the best linear predictor for any zero-mean distribution with covariance~$R$, and \cref{prop:identity} uses only second moments.

Every job of the canonical run is single-threaded with a fixed hash seed in a pinned environment (Python~3.12, NumPy~2.4.6, SciPy~1.18.0), and two runs of the CPU pipeline produce byte-identical logs.

\providecommand{\dlrr}{\raggedright\hspace{0pt}}
\providecommand{\dln}[1]{\textsuperscript{\,#1}}

\section{Data, licenses and download defects}
\label{app:data}

This appendix documents the score matrices of \cref{sec:protocol}: where each comes from, under which terms we use and redistribute it, how \Merged{} is assembled, and two defects of our March 2026 downloads that were found after submission.
Every matrix is built from public leaderboard results, and every build script, pinned revision and checksum is part of the release.

\subsection{Matrices and their sources}
\label{app:data-sources}

\Cref{tab:app-data} lists the eleven matrices of the paper, the pre-fix \MergedVone{} that serves only as a sensitivity check, the eight further collections that \Merged{} draws on, and the cost tables of the budget experiments (\cref{sec:robust-cost} and \cref{app:costs}).
\MMLU, \MTEB, \Merged{} and \BenchPress{} are the March/April 2026 snapshot used since the submission.
The seven extended matrices were built in October 2026 from pinned inputs: each build script checks the SHA-256 checksum of every raw input and refuses inputs that differ.
Epoch and \textsc{HELM} replace their files in place, so a rebuild from upstream succeeds only while the pinned versions remain available.

\begin{table}[p]
\centering
\caption{Score matrices and their sources.
\emph{Revision}: the upstream commit, release or checksum that our file corresponds to; ``equals rev.~$x$'': fetched without a pin, with all values equal to those at revision~$x$.
Letters refer to the license notes of \cref{app:data-licences}; $\ast$ marks the nine collections pooled in \Merged.}
\label{tab:app-data}
\scriptsize
\setlength{\tabcolsep}{3pt}
\renewcommand{\arraystretch}{1.08}
\begin{tabular}{@{}p{2.05cm}p{5.6cm}p{1.15cm}p{2.3cm}p{1.45cm}@{}}
\toprule
\dlrr Matrix\newline $M\times N$; observed & \dlrr Source; revision & \dlrr Retrieved & \dlrr Upstream license (as declared) & \dlrr We release \tabularnewline
\midrule
\multicolumn{5}{@{}l}{Main matrices} \tabularnewline
\dlrr \MMLU{}$^\ast$\newline $\nv{DataMmluM}\times\nv{DataMmluN}$; \nv[1]{DataMmluObsPct}\%
 & \dlrr Open LLM Leaderboard v1~\citep{OpenLLMLeaderboard1,Hendrycks21mmlu}, HF \nolinkurl{open-llm-leaderboard-old/results}; 5-shot accuracy per subject; unchanged since 2024-07-18 (rev.~23474373)
 & \dlrr 2026-03-13 & \dlrr none declared\dln{a} & \dlrr matrix \tabularnewline
\dlrr \MTEB\newline $\nv{DataMtebM}\times\nv{DataMtebN}$; \nv[1]{DataMtebObsPct}\%
 & \dlrr \citep{Muennighoff23,MTEBResults26}, GitHub \nolinkurl{embeddings-benchmark/results} at 37dfa3b5; \textsc{MTEB}(eng, classic)
 & \dlrr 2026-03-13 & \dlrr none then; CC0-1.0 since 2026-08-05 & \dlrr matrix \tabularnewline
\dlrr \Merged\newline $\nv{DataMergedM}\times\nv{DataMergedN}$; \nv[1]{DataMergedObsPct}\%
 & \dlrr the nine collections marked $\ast$, joined by canonical model names (\cref{app:data-merged}); rebuilt 2026-10-18 (\cref{app:data-fixes})
 & \dlrr --- & \dlrr as its sources & \dlrr matrix; name mapping \tabularnewline
\dlrr \BenchPress\newline $\nv{DataBenchpressM}\times\nv{DataBenchpressN}$; \nv[1]{DataBenchpressObsPct}\%
 & \dlrr \citep{Papailiopoulos26benchpress}, GitHub \nolinkurl{anadim/llm-benchmark-matrix} at e93b33f1; scores from papers, model cards and leaderboards, each with a source URL
 & \dlrr 2026-04-10 & \dlrr README says MIT; no license file\dln{e} & \dlrr pinned download script; checksum \tabularnewline
\midrule
\multicolumn{5}{@{}l}{Extended matrices (pinned inputs with SHA-256 manifests)} \tabularnewline
\dlrr \OLLtwo\newline $\nv{DataOlvTwoSubM}\times\nv{DataOlvTwoSubN}$; \nv[1]{DataOlvTwoSubObsPct}\%
 & \dlrr \citep{OpenLLMLeaderboard2}, HF \nolinkurl{open-llm-leaderboard/results} at aa81ecc3 and \nolinkurl{/contents} at 9c09a7ca; the 42 sub-task metrics behind the six v2 aggregates
 & \dlrr 2026-10-17 & \dlrr none declared\dln{a} & \dlrr matrix \tabularnewline
\dlrr \OLLone\newline $\nv{DataOlvOneItemsM}\times\nv{DataOlvOneItemsN}$; \nv[1]{DataOlvOneItemsObsPct}\%
 & \dlrr tinyBenchmarks~\citep{Polo24tiny} \nolinkurl{lb.pickle} at e9a8b103: \nv{ItemBudgetNItemsOlvOneItems} items of 57 \MMLU{} subjects and five further tasks, checked against the Open LLM Leaderboard v1
 & \dlrr 2026-10-17 & \dlrr MIT\dln{f} & \dlrr matrix; item correctness \tabularnewline
\dlrr \Epoch\newline $\nv{DataEpochM}\times\nv{DataEpochN}$; \nv[1]{DataEpochObsPct}\%
 & \dlrr \citep{EpochBenchmarks26}, \nolinkurl{epoch.ai/data/benchmark_data.zip}, pinned by SHA-256 (e1ce578c\ldots); input table of the Epoch Capabilities Index
 & \dlrr 2026-10-17 & \dlrr CC~BY~4.0; external sources keep theirs\dln{g} & \dlrr 15 Epoch columns; build script \tabularnewline
\dlrr \HELMLite\newline $\nv{DataHelmLiteM}\times\nv{DataHelmLiteN}$; \nv[1]{DataHelmLiteObsPct}\%
 & \dlrr \citep{Liang23}, bucket \texttt{crfm-helm-public}, release v1.13.0; objects pinned by storage generation and SHA-256; release dates from \nolinkurl{stanford-crfm/helm} at 63754d05
 & \dlrr 2026-10-18 & \dlrr none for results; code Apache-2.0\dln{b} & \dlrr matrix; token costs \tabularnewline
\dlrr \HELMMMLU\newline $\nv{DataHelmMmluM}\times\nv{DataHelmMmluN}$; \nv[1]{DataHelmMmluObsPct}\%
 & \dlrr as \HELMLite, release v1.13.0
 & \dlrr 2026-10-18 & \dlrr as above\dln{b} & \dlrr matrix; token costs \tabularnewline
\dlrr \HELMClassic\newline $\nv{DataHelmClassicM}\times\nv{DataHelmClassicN}$; \nv[1]{DataHelmClassicObsPct}\%
 & \dlrr as \HELMLite, release v0.4.0, few-shot runs only
 & \dlrr 2026-10-18 & \dlrr as above\dln{b} & \dlrr matrix; token costs \tabularnewline
\dlrr \MTEBnew\newline $\nv{DataMtebTwothousandtwentysixM}\times\nv{DataMtebTwothousandtwentysixN}$; \nv[1]{DataMtebTwothousandtwentysixObsPct}\%
 & \dlrr \citep{MTEBResults26}, HF \nolinkurl{mteb/results} at 9d3f92b4; task and model metadata (release dates) of the \texttt{mteb} package 2.18.7
 & \dlrr 2026-10-18 & \dlrr CC0-1.0; metadata Apache-2.0 & \dlrr matrix \tabularnewline
\midrule
\multicolumn{5}{@{}l}{Other collections pooled in \Merged, and the pre-fix \MergedVone} \tabularnewline
\dlrr Open LLM v2$^\ast$\newline 4{,}507$\times$6
 & \dlrr \citep{OpenLLMLeaderboard2}, the six aggregates; per model the result file that the leaderboard shows; rebuilt 2026-10-18 (\cref{app:data-fixes})
 & \dlrr 2026-10-17 & \dlrr none declared\dln{a} & \dlrr source table \tabularnewline
\dlrr \HELMLite{} v1.0$^\ast$\newline 30$\times$11
 & \dlrr \citep{LiangHELMLite23}, bucket \texttt{crfm-helm-public}, release v1.0.0, \texttt{core\_scenarios}
 & \dlrr 2026-03-13 & \dlrr as \HELMLite\dln{b} & \dlrr source table \tabularnewline
\dlrr AlpacaEval~2$^\ast$\newline 223$\times$10
 & \dlrr \citep{Dubois24}, GitHub \nolinkurl{tatsu-lab/alpaca_eval}, leaderboard CSV; equals rev.~2898342e
 & \dlrr 2026-03-13 & \dlrr code Apache-2.0; data badge CC~BY-NC~4.0\dln{c} & \dlrr source table \tabularnewline
\dlrr Arena-Hard-Auto$^\ast$\newline 60$\times$4
 & \dlrr \citep{ArenaHard24}, GitHub \nolinkurl{lmarena/arena-hard-auto}, leaderboard CSV of 2024-07-31; equals rev.~b00effbe
 & \dlrr 2026-03-13 & \dlrr Apache-2.0 & \dlrr source table \tabularnewline
\dlrr LiveBench$^\ast$\newline 195$\times$3
 & \dlrr \citep{White24livebench}, HF \nolinkurl{livebench/model_judgment}, mean judgment for the 3 of 6 categories it covers; rev.~9704e5da
 & \dlrr 2026-03-13 & \dlrr none declared\dln{a} & \dlrr source table \tabularnewline
\dlrr WildBench$^\ast$\newline 63$\times$11
 & \dlrr \citep{Lin24wildbench}, HF Space \nolinkurl{allenai/WildBench}, WB-Score per task category; equals rev.~4b94d1d8
 & \dlrr 2026-03-13 & \dlrr none declared\dln{a} & \dlrr source table \tabularnewline
\dlrr MT-Bench$^\ast$\newline 5$\times$8
 & \dlrr \citep{Zheng23,SUSTechMTBenchJudge23}, HF \nolinkurl{SUSTech/mt_bench_judge}, GPT-4 scores averaged over both turns; equals rev.~c489203c
 & \dlrr 2026-03-13 & \dlrr none declared; original ``other''\dln{d} & \dlrr source table \tabularnewline
\dlrr BigCodeBench$^\ast$\newline 155$\times$14
 & \dlrr \citep{Zhuo24bigcode}, HF \nolinkurl{bigcode/bigcodebench-domain}, solve rate for 7 domains $\times$ \{Complete, Instruct\}; equals rev.~6176f481
 & \dlrr 2026-03-13 & \dlrr none declared\dln{a} & \dlrr source table \tabularnewline
\dlrr \MergedVone\newline $\nv{DataMergedVOneM}\times\nv{DataMergedVOneN}$; \nv[1]{DataMergedVOneObsPct}\%
 & \dlrr as \Merged, with the Open LLM v2 table of the submission
 & \dlrr --- & \dlrr as its sources & \dlrr matrix \tabularnewline
\midrule
\dlrr Cost tables
 & \dlrr test-split sizes of HF \nolinkurl{cais/mmlu} at c30699e8 (metadata only); \MTEB{} \texttt{num\_samples} of the \texttt{mteb} package 2.18.7
 & \dlrr 2026-07-27; 2026-10-17 & \dlrr MIT; Apache-2.0 & \dlrr counts \tabularnewline
\bottomrule
\end{tabular}
\end{table}

\paragraph{Costs.}
\MMLU{} costs are test questions per subject, from \nv{CostMinMmlu} to \nv{CostMaxMmlu}.
They sum to 14{,}042.
\MTEB{} costs are the \texttt{num\_samples} of the evaluation splits that the matrix scores, from \nv{CostMinMteb} to \nv{CostMaxMteb} per task, nearly five orders of magnitude.
What one sample is depends on the task type: corpus documents plus queries for retrieval and reranking, texts for clustering and classification, sentence pairs for semantic similarity and pair classification, and source texts for summarization.
The \textsc{HELM} token costs of \cref{app:costs} are \textsc{HELM}'s own counts of prompt and output tokens per instance, times instances and requests per instance.

\paragraph{Further caveats.}
MT-Bench covers 5 models, so its 8 columns are observed for only 3 models of \Merged.
\BenchPress{} collects scores from heterogeneous public reports with different evaluation settings.
In \OLLtwo, the MATH sub-tasks were scored by three versions of the scorer, so the split by submission date in \cref{sec:robust-new} leaves MATH out.
In \OLLone, the TruthfulQA items are continuous mc2 scores rather than 0/1.
\HELMClassic{} differs from the \textsc{HELM} website in 19 cells, where the website pools zero-shot ablation runs into the few-shot results.
\Epoch{} holds Epoch's normalized scores (the best variant per model group, rescaled between the random baseline and the ceiling), so one row may merge several versions of a model.

\subsection{Licenses}
\label{app:data-licences}

We redistribute numeric scores only: model-by-benchmark matrices with observation masks, model and benchmark names, and metadata such as revisions, checksums, dates and evaluation costs.
We redistribute no benchmark questions, reference answers, prompts or model-generated text.
The one item-level file stores the per-item correctness of \OLLone{} models, indexed by position, without item text.
The code is released under Apache-2.0.
Each data file keeps the terms of its upstream source; we do not relicense upstream data. Our own additions (model-name mapping, metadata we authored) are released under CC~BY~4.0.
Attributions, our changes and the license texts or links to them are in \texttt{DATA\_LICENSES.md} and \texttt{LICENSE-DATA} of the release; the TabImpute checkpoint we use is Apache-2.0.
The letters below refer to \cref{tab:app-data}.

\begin{itemize*}
\item[(a)] The dataset card declares no license, so use is governed by the Hugging Face Terms of Service, whose grant to other users covers use through the Hub.
  We redistribute only derived numeric scores, with attribution, and will remove files on request. Takedown requests go to \texttt{alex@smola.org}.
\item[(b)] \textsc{HELM}'s code is Apache-2.0, and \textsc{HELM} states no terms for its published results.
  We redistribute derived numeric scores with attribution to \textsc{HELM}~\citep{Liang23,LiangHELMLite23}.
\item[(c)] The leaderboard CSV sits in an Apache-2.0 source tree, but the README's data badge states CC~BY-NC~4.0.
  We treat the AlpacaEval scores, including the three AlpacaEval columns of \Merged, as CC~BY-NC~4.0: attribution, non-commercial use, changes indicated.
\item[(d)] The 800 rows of the SUSTech dataset are copies of LMSYS's GPT-4 single-answer judgments for MT-Bench~\citep{Zheng23}; the LMSYS Hugging Face Space declares ``license: other'' and states no terms.
  Note~(a) applies.
\item[(e)] The README states MIT, but the repository has no license file and names no copyright holder.
  We therefore ship no \BenchPress{} data, only a download script pinned to commit e93b33f1 that checks the file's SHA-256 checksum and rebuilds the matrix.
\item[(f)] The MIT license of tinyBenchmarks (\copyright~2024 Felipe Maia Polo) covers \nolinkurl{lb.pickle}, and its notice ships with \OLLone.
  The item results in it were produced by the Open LLM Leaderboard, whose per-model detail datasets declare no license.
\item[(g)] Epoch's data are free to use and redistribute with credit, but 46 of the 61 columns come from external projects that keep their own licenses.
  The release ships only the 15 columns taken from Epoch's own files and a script that rebuilds the full matrix from Epoch's download; it credits the creators of the other columns.
\item[(h)] BIG-bench Lite~\citep{Srivastava23} (Apache-2.0) was downloaded but is used in no experiment: its models appear in no other collection. We include it in the release for completeness.
\end{itemize*}

\paragraph{Personal information.}
Rows are identified by public model names.
Hugging Face repository identifiers include the public user or organization name of the uploader; the matrices contain no other information about people.

\subsection{Download defects}
\label{app:data-fixes}

Two defects of our March 2026 downloads were found after submission.
We fixed the first, which affected cells of \Merged, and disclose the second, which only drops models from \MMLU.

\paragraph{Open LLM v2 and \Merged{} (fixed).}
Our original downloader kept each model's last result file in path order and, where a group entry was blank, stored the first sub-task's score as the group score.
Rebuilding the Open LLM v2 table with the leaderboard's own choice of result file changes 200 cells in 47 of its 4{,}507 rows: 180 values and 20 cells that were missing.
Two of these models are in \Merged, where the rebuild fills 8 missing cells, changes no observed value and raises the observed cells from 4{,}185 to 4{,}193.
All results in this paper use the fixed matrix; the submission's matrix, \MergedVone, serves as a sensitivity check (\cref{tab:app-results}). The release includes the pre-fix files \texttt{open\_llm\_v1} and \texttt{merged\_v1}, which reproduce the submission's numbers.
The fix does not change the picture: facility location reaches $R^2=\nv{RsqBaseMergedVOneFaclocKfive}$ at $k=5$ on \MergedVone{} against \nv{RsqBaseMergedFaclocKfive} on \Merged, its lead over explained variance stays non-significant (FL~$-$~EV: \nvneg{PairDiffBaseMergedVOneTraceVsFaclocKfive} \nvcineg{PairDiffBaseMergedVOneTraceVsFaclocKfive}), and its lead over MI stays significant (FL~$-$~MI: \nvneg{PairDiffBaseMergedVOneMiVsFaclocKfive} \nvcineg{PairDiffBaseMergedVOneMiVsFaclocKfive}).
Because \MergedVone{} is \Merged{} without these 8 cells, the scoreboard of \cref{tab:main} counts only \Merged.

\paragraph{\MMLU{} (disclosed, kept as frozen).}
The \MMLU{} downloader reads only the last results file of each model directory, so re-evaluated models show their last run, and 1{,}640 models whose \MMLU{} results sit only in earlier files are missing.
Writing to a case-insensitive file system also merged 18 model directories whose names differ only in case: 16 models were lost, and one row was filled from the run stored under the other casing (the two runs differ by up to 0.119).
We keep the matrix as it was frozen for the submission and the rebuttal; of its 5{,}454 rows we drop the two that observe fewer than half of the subjects, which leaves \nv{DataMmluM}.
The defects decide which models and which run of a model enter the matrix; apart from that one row, every kept value is a score that the leaderboard published for that model.
The downloader documents both effects and can recover the earlier files.

\subsection{Construction of \Merged}
\label{app:data-merged}

\Merged{} pools the nine collections marked in \cref{tab:app-data}, which share many models.
\MTEB{} stays out: its embedding models appear in no other collection.
The same model often appears under different names (for example \texttt{Meta-Llama-3-70B-Instruct} and \texttt{llama-3-70b-instruct}).
We matched names across collections with the help of an LLM (Claude Sonnet~4.6\ifdefined\techreport\else; checklist Q16\fi) and release the resulting mapping, which lists for every model of \Merged{} its original name in each collection where it appears.
A model enters \Merged{} if it appears in at least two collections, which gives \nv{DataMergedM} models.
Each column keeps its collection as a prefix (for example \texttt{mmlu/astronomy} or \texttt{helm/GSM8K}), so copies of one benchmark in two collections stay separate columns.
We drop the ten auxiliary columns of AlpacaEval and Arena-Hard-Auto (standard errors, counts, interval bounds, response lengths and token counts), which are not benchmark scores.
\HELMLite's ``Mean win rate'' stays: it is \textsc{HELM}'s ranking statistic, which depends on the set of models, rather than a benchmark score, and we flag it here.
The result has \nv{DataMergedN} columns, of which \nv[1]{DataMergedObsPct}\% of cells are observed.

\section{All eleven matrices and training-set size}
\label{app:results}

\Cref{tab:app-results} gives the numbers behind two statements of \cref{sec:results}: panel~(a) shows that the seven extended matrices follow the pattern of the four main ones, and panel~(b) that with few training models the selector stops mattering on sparse matrices.

\begin{table}[t]
\centering
\small
\caption{Mean $R^2$ over the 50 matched splits at $k = 5$. (a)~All eleven matrices with the holdout fraction $p = 0.1$ of the main results. FL\,$-$\,EV and FL\,$-$\,Random are paired differences with Nadeau--Bengio 95\% intervals (\cref{app:protocol-stats}); an interval that excludes zero is significant. \MergedVone{} (last row) is \Merged{} before the data fix of \cref{app:data-fixes}; it is a sensitivity check and not part of the scoreboard of \cref{tab:main}. (b)~Training on about a tenth of the models ($p = 0.9$) with the same validation folds; MI, hub and PCA were not run at this holdout fraction. Values for every $k$ from 1 to 15 are in the released numbers file; \cref{fig:curves} plots them for the four main matrices.}
\label{tab:app-results}
{\footnotesize\setlength{\tabcolsep}{2.2pt}
\textit{(a) All matrices, $k = 5$, holdout $p = 0.1$}\\[2pt]
\begin{tabular}{@{}lrrrrrrrll@{}}
\toprule
Matrix & FL & EV & MI & Entropy & Hub & PCA & Random & FL$\,-\,$EV \scriptsize [95\% CI] & FL$\,-\,$Random \scriptsize [95\% CI] \\
\midrule
\MMLU & \nv{RsqBaseMmluFaclocKfive} & \textbf{\nv{RsqBaseMmluTraceKfive}} & \nv{RsqBaseMmluMiKfive} & \nv{RsqBaseMmluEntropyKfive} & \nv{RsqBaseMmluHubKfive} & \nv{RsqBaseMmluPcaKfive} & \nv{RsqBaseMmluRandomKfive} & {\scriptsize\nvneg{PairDiffBaseMmluTraceVsFaclocKfive}\,{\scriptsize\nvcineg{PairDiffBaseMmluTraceVsFaclocKfive}}} & {\scriptsize\nvd{PairDiffBaseMmluFaclocVsRandomKfive}\,{\scriptsize\nvci{PairDiffBaseMmluFaclocVsRandomKfive}}} \\
\MTEB & \nv{RsqBaseMtebFaclocKfive} & \textbf{\nv{RsqBaseMtebTraceKfive}} & \nv{RsqBaseMtebMiKfive} & \nv{RsqBaseMtebEntropyKfive} & \nv{RsqBaseMtebHubKfive} & \nv{RsqBaseMtebPcaKfive} & \nv{RsqBaseMtebRandomKfive} & {\scriptsize\nvneg{PairDiffBaseMtebTraceVsFaclocKfive}\,{\scriptsize\nvcineg{PairDiffBaseMtebTraceVsFaclocKfive}}} & {\scriptsize\nvd{PairDiffBaseMtebFaclocVsRandomKfive}\,{\scriptsize\nvci{PairDiffBaseMtebFaclocVsRandomKfive}}} \\
\Merged & \textbf{\nv{RsqBaseMergedFaclocKfive}} & \nv{RsqBaseMergedTraceKfive} & \nv{RsqBaseMergedMiKfive} & \nv{RsqBaseMergedEntropyKfive} & \nv{RsqBaseMergedHubKfive} & \nv{RsqBaseMergedPcaKfive} & \nv{RsqBaseMergedRandomKfive} & {\scriptsize\nvneg{PairDiffBaseMergedTraceVsFaclocKfive}\,{\scriptsize\nvcineg{PairDiffBaseMergedTraceVsFaclocKfive}}} & {\scriptsize\nvd{PairDiffBaseMergedFaclocVsRandomKfive}\,{\scriptsize\nvci{PairDiffBaseMergedFaclocVsRandomKfive}}} \\
\BenchPress & \nv{RsqBaseBenchpressFaclocKfive} & \nv{RsqBaseBenchpressTraceKfive} & \nv{RsqBaseBenchpressMiKfive} & \nv{RsqBaseBenchpressEntropyKfive} & \nv{RsqBaseBenchpressHubKfive} & \nv{RsqBaseBenchpressPcaKfive} & \textbf{\nv{RsqBaseBenchpressRandomKfive}} & {\scriptsize\nvneg{PairDiffBaseBenchpressTraceVsFaclocKfive}\,{\scriptsize\nvcineg{PairDiffBaseBenchpressTraceVsFaclocKfive}}} & {\scriptsize\nvd{PairDiffBaseBenchpressFaclocVsRandomKfive}\,{\scriptsize\nvci{PairDiffBaseBenchpressFaclocVsRandomKfive}}} \\
\OLLtwo & \nv{RsqBaseOlvTwoSubFaclocKfive} & \textbf{\nv{RsqBaseOlvTwoSubTraceKfive}} & \nv{RsqBaseOlvTwoSubMiKfive} & \nv{RsqBaseOlvTwoSubEntropyKfive} & \nv{RsqBaseOlvTwoSubHubKfive} & \nv{RsqBaseOlvTwoSubPcaKfive} & \nv{RsqBaseOlvTwoSubRandomKfive} & {\scriptsize\nvneg{PairDiffBaseOlvTwoSubTraceVsFaclocKfive}\,{\scriptsize\nvcineg{PairDiffBaseOlvTwoSubTraceVsFaclocKfive}}} & {\scriptsize\nvd{PairDiffBaseOlvTwoSubFaclocVsRandomKfive}\,{\scriptsize\nvci{PairDiffBaseOlvTwoSubFaclocVsRandomKfive}}} \\
\OLLone & \nv{RsqBaseOlvOneItemsFaclocKfive} & \textbf{\nv{RsqBaseOlvOneItemsTraceKfive}} & \nv{RsqBaseOlvOneItemsMiKfive} & \nv{RsqBaseOlvOneItemsEntropyKfive} & \nv{RsqBaseOlvOneItemsHubKfive} & \nv{RsqBaseOlvOneItemsPcaKfive} & \nv{RsqBaseOlvOneItemsRandomKfive} & {\scriptsize\nvneg{PairDiffBaseOlvOneItemsTraceVsFaclocKfive}\,{\scriptsize\nvcineg{PairDiffBaseOlvOneItemsTraceVsFaclocKfive}}} & {\scriptsize\nvd{PairDiffBaseOlvOneItemsFaclocVsRandomKfive}\,{\scriptsize\nvci{PairDiffBaseOlvOneItemsFaclocVsRandomKfive}}} \\
\Epoch & \textbf{\nv{RsqBaseEpochFaclocKfive}} & \nv{RsqBaseEpochTraceKfive} & \nv{RsqBaseEpochMiKfive} & \nv{RsqBaseEpochEntropyKfive} & \nv{RsqBaseEpochHubKfive} & \nv{RsqBaseEpochPcaKfive} & \nv{RsqBaseEpochRandomKfive} & {\scriptsize\nvneg{PairDiffBaseEpochTraceVsFaclocKfive}\,{\scriptsize\nvcineg{PairDiffBaseEpochTraceVsFaclocKfive}}} & {\scriptsize\nvd{PairDiffBaseEpochFaclocVsRandomKfive}\,{\scriptsize\nvci{PairDiffBaseEpochFaclocVsRandomKfive}}} \\
\HELMLite & \nv{RsqBaseHelmLiteFaclocKfive} & \nv{RsqBaseHelmLiteTraceKfive} & \nv{RsqBaseHelmLiteMiKfive} & \textbf{\nv{RsqBaseHelmLiteEntropyKfive}} & \nv{RsqBaseHelmLiteHubKfive} & \nv{RsqBaseHelmLitePcaKfive} & \nv{RsqBaseHelmLiteRandomKfive} & {\scriptsize\nvneg{PairDiffBaseHelmLiteTraceVsFaclocKfive}\,{\scriptsize\nvcineg{PairDiffBaseHelmLiteTraceVsFaclocKfive}}} & {\scriptsize\nvd{PairDiffBaseHelmLiteFaclocVsRandomKfive}\,{\scriptsize\nvci{PairDiffBaseHelmLiteFaclocVsRandomKfive}}} \\
\HELMMMLU & \nv{RsqBaseHelmMmluFaclocKfive} & \textbf{\nv{RsqBaseHelmMmluTraceKfive}} & \nv{RsqBaseHelmMmluMiKfive} & \nv{RsqBaseHelmMmluEntropyKfive} & \nv{RsqBaseHelmMmluHubKfive} & \nv{RsqBaseHelmMmluPcaKfive} & \nv{RsqBaseHelmMmluRandomKfive} & {\scriptsize\nvneg{PairDiffBaseHelmMmluTraceVsFaclocKfive}\,{\scriptsize\nvcineg{PairDiffBaseHelmMmluTraceVsFaclocKfive}}} & {\scriptsize\nvd{PairDiffBaseHelmMmluFaclocVsRandomKfive}\,{\scriptsize\nvci{PairDiffBaseHelmMmluFaclocVsRandomKfive}}} \\
\HELMClassic & \textbf{\nv{RsqBaseHelmClassicFaclocKfive}} & \nv{RsqBaseHelmClassicTraceKfive} & \nv{RsqBaseHelmClassicMiKfive} & \nv{RsqBaseHelmClassicEntropyKfive} & \nv{RsqBaseHelmClassicHubKfive} & \nv{RsqBaseHelmClassicPcaKfive} & \nv{RsqBaseHelmClassicRandomKfive} & {\scriptsize\nvneg{PairDiffBaseHelmClassicTraceVsFaclocKfive}\,{\scriptsize\nvcineg{PairDiffBaseHelmClassicTraceVsFaclocKfive}}} & {\scriptsize\nvd{PairDiffBaseHelmClassicFaclocVsRandomKfive}\,{\scriptsize\nvci{PairDiffBaseHelmClassicFaclocVsRandomKfive}}} \\
\MTEBnew & \textbf{\nv{RsqBaseMtebTwothousandtwentysixFaclocKfive}} & \nv{RsqBaseMtebTwothousandtwentysixTraceKfive} & \nv{RsqBaseMtebTwothousandtwentysixMiKfive} & \nv{RsqBaseMtebTwothousandtwentysixEntropyKfive} & \nv{RsqBaseMtebTwothousandtwentysixHubKfive} & \nv{RsqBaseMtebTwothousandtwentysixPcaKfive} & \nv{RsqBaseMtebTwothousandtwentysixRandomKfive} & {\scriptsize\nvneg{PairDiffBaseMtebTwothousandtwentysixTraceVsFaclocKfive}\,{\scriptsize\nvcineg{PairDiffBaseMtebTwothousandtwentysixTraceVsFaclocKfive}}} & {\scriptsize\nvd{PairDiffBaseMtebTwothousandtwentysixFaclocVsRandomKfive}\,{\scriptsize\nvci{PairDiffBaseMtebTwothousandtwentysixFaclocVsRandomKfive}}} \\
\midrule
\MergedVone & \textbf{\nv{RsqBaseMergedVOneFaclocKfive}} & \nv{RsqBaseMergedVOneTraceKfive} & \nv{RsqBaseMergedVOneMiKfive} & \nv{RsqBaseMergedVOneEntropyKfive} & \nv{RsqBaseMergedVOneHubKfive} & \nv{RsqBaseMergedVOnePcaKfive} & \nv{RsqBaseMergedVOneRandomKfive} & {\scriptsize\nvneg{PairDiffBaseMergedVOneTraceVsFaclocKfive}\,{\scriptsize\nvcineg{PairDiffBaseMergedVOneTraceVsFaclocKfive}}} & {\scriptsize\nvd{PairDiffBaseMergedVOneFaclocVsRandomKfive}\,{\scriptsize\nvci{PairDiffBaseMergedVOneFaclocVsRandomKfive}}} \\
\bottomrule
\end{tabular}

\medskip
\textit{(b) Fewer training models: $k = 5$, holdout $p = 0.9$}\\[2pt]
\begin{tabular}{@{}lrrrrl@{}}
\toprule
Matrix & FL & EV & Entropy & Random & FL$\,-\,$Random \scriptsize [95\% CI] \\
\midrule
\MMLU & \nv{RsqMmluFaclocKfivePninety} & \textbf{\nv{RsqMmluTraceKfivePninety}} & \nv{RsqMmluEntropyKfivePninety} & \nv{RsqMmluRandomKfivePninety} & {\nvd{PairDiffMmluFaclocVsRandomKfivePninety}\,{\scriptsize\nvci{PairDiffMmluFaclocVsRandomKfivePninety}}} \\
\MTEB & \nv{RsqMtebFaclocKfivePninety} & \textbf{\nv{RsqMtebTraceKfivePninety}} & \nv{RsqMtebEntropyKfivePninety} & \nv{RsqMtebRandomKfivePninety} & {\nvd{PairDiffMtebFaclocVsRandomKfivePninety}\,{\scriptsize\nvci{PairDiffMtebFaclocVsRandomKfivePninety}}} \\
\Merged & \nv{RsqMergedFaclocKfivePninety} & \textbf{\nv{RsqMergedTraceKfivePninety}} & \nv{RsqMergedEntropyKfivePninety} & \nv{RsqMergedRandomKfivePninety} & {\nvd{PairDiffMergedFaclocVsRandomKfivePninety}\,{\scriptsize\nvci{PairDiffMergedFaclocVsRandomKfivePninety}}} \\
\BenchPress & \nv{RsqBenchpressFaclocKfivePninety} & \textbf{\nv{RsqBenchpressTraceKfivePninety}} & \nv{RsqBenchpressEntropyKfivePninety} & \nv{RsqBenchpressRandomKfivePninety} & {\nvd{PairDiffBenchpressFaclocVsRandomKfivePninety}\,{\scriptsize\nvci{PairDiffBenchpressFaclocVsRandomKfivePninety}}} \\
\Epoch & \textbf{\nv{RsqEpochFaclocKfivePninety}} & \nv{RsqEpochTraceKfivePninety} & \nv{RsqEpochEntropyKfivePninety} & \nv{RsqEpochRandomKfivePninety} & {\nvd{PairDiffEpochFaclocVsRandomKfivePninety}\,{\scriptsize\nvci{PairDiffEpochFaclocVsRandomKfivePninety}}} \\
\OLLtwo & \textbf{\nv{RsqOlvTwoSubFaclocKfivePninety}} & \textbf{\nv{RsqOlvTwoSubTraceKfivePninety}} & \nv{RsqOlvTwoSubEntropyKfivePninety} & \nv{RsqOlvTwoSubRandomKfivePninety} & {\nvd{PairDiffOlvTwoSubFaclocVsRandomKfivePninety}\,{\scriptsize\nvci{PairDiffOlvTwoSubFaclocVsRandomKfivePninety}}} \\
\MTEBnew & \nv{RsqMtebTwothousandtwentysixFaclocKfivePninety} & \textbf{\nv{RsqMtebTwothousandtwentysixTraceKfivePninety}} & \nv{RsqMtebTwothousandtwentysixEntropyKfivePninety} & \nv{RsqMtebTwothousandtwentysixRandomKfivePninety} & {\nvd{PairDiffMtebTwothousandtwentysixFaclocVsRandomKfivePninety}\,{\scriptsize\nvci{PairDiffMtebTwothousandtwentysixFaclocVsRandomKfivePninety}}} \\
\bottomrule
\end{tabular}}
 (FB) not yet generated}}
\end{table}

\paragraph{The extended matrices repeat the pattern.}
At $k = 5$ the best method is a coverage objective on every matrix except \BenchPress{} (random) and \HELMLite, where entropy is ahead of explained variance by \nvdiff{RsqBaseHelmLiteEntropyKfive}{RsqBaseHelmLiteTraceKfive}. On the three dense extended matrices (\OLLtwo, \OLLone{} and \HELMMMLU), facility location and explained variance lie close together (FL\,$-$\,EV in the table) and above random and hub. On \Epoch{} and \MTEBnew{} facility location has the higher mean, though not significantly. The intervals against random include zero on \MTEB, \Merged, \BenchPress{} and the two small HELM matrices.
Over every $2 \le k \le \min(15, N-3)$, facility location is never significantly ahead of explained variance and is significantly behind in \nv{ScoreAllKFaclocVsTraceSigWorse} of \nv{ScoreAllKNCells} settings, by at most \nv{ScoreAllKFaclocVsTraceMaxDeficit}, all on \MMLU, \OLLtwo{} and \OLLone.

\paragraph{Few training models hurt only where correlations are hard to estimate.}
Trained on a random tenth of the models, facility location keeps $R^2 = \nv{RsqMmluFaclocKfivePninety}$ on \MMLU{} and \nv{RsqOlvTwoSubFaclocKfivePninety} on \OLLtwo{} (against \nv{RsqBaseOlvTwoSubFaclocKfive} at $p = 0.1$), while on \Merged{} it drops to \nv{RsqMergedFaclocKfivePninety}, level with random (\nv{RsqMergedRandomKfivePninety}), and on \BenchPress{} too every selector falls to the level of random.
The Nadeau--Bengio correction alone makes the intervals of panel~(b) about 2.8 times wider than those of panel~(a), because it grows with the ratio of validation to training models.

\section{New models: prospective splits and model-family holdout}
\label{app:newmodels}

This appendix backs the results on new models in \cref{sec:robust}: \emph{prospective} splits, in which old models choose the benchmarks and models released later are predicted, and the \emph{model-family holdout}, in which each family of related models is held out in turn (designs in \cref{app:protocol-designs}). It adds further prospective settings, including \Epoch, a correction for a shift of the new models as a group, and the family holdout on all eleven matrices.

\subsection{Prospective splits}
\label{app:newmodels-prosp}

\paragraph{Settings.}
Besides the three settings of \cref{sec:robust-new}, \cref{tab:app-prosp} reports the \OLLtwo{} cutoffs of October 2024 and March 2025, the \OLLtwo{} variant that keeps the MATH columns (\nv{NBenchProspOlvTwoSubMathCutJanTwentyfive} instead of \nv{NBenchProspOlvTwoSubCutJanTwentyfive} benchmarks), the \nv{NTestProspMtebLate} new \MTEBnew{} models released after the snapshot (late), and \Epoch. On \Epoch{} models are split by release date at four cutoffs between July 2024 and January 2026.

\begin{table}[t]
\centering
\small
\caption{Prospective evaluation at $k = 5$: $R^2$ \eqref{eq:r2} on new models when old models estimate $\hat R$ and select. Old\,$\to$\,new gives the numbers of models. Control: facility location's mean $R^2$ over 50 random splits of the same models with the same sizes. FL\,$-$\,EV: difference on the new models with a 95\% percentile-bootstrap interval over new models (\cref{app:protocol-stats}); it omits the variability of the selection, and every setting is a single split. A new model counts for a method only if it observes one of the method's picks; ``---'' marks methods whose picks fewer than 10 new models observe.}
\label{tab:app-prosp}
{\footnotesize\setlength{\tabcolsep}{2.2pt}
\begin{tabular}{@{}lrrrrrrrrrl@{}}
\toprule
Setting & Old\,$\to$\,new & FL & EV & MI & Entropy & Hub & PCA & Random & Control & FL$\,-\,$EV \scriptsize [95\% CI] \\
\midrule
\multicolumn{11}{@{}l}{\HELMLite} \\
\quad v1.0\,$\to$\,later & \nv{NTrainProspHelm}\,$\to$\,\nv{NTestProspHelm} & \textbf{\nv{RsqProspHelmFaclocKfive}} & \nv{RsqProspHelmTraceKfive} & \nv{RsqProspHelmMiKfive} & \nv{RsqProspHelmEntropyKfive} & \nv{RsqProspHelmHubKfive} & \nv{RsqProspHelmPcaKfive} & \nv{RsqProspHelmRandomKfive} & \nv{RsqCtrlHelmFaclocKfive} & {\scriptsize\nvneg{BootDiffProspHelmTraceVsFaclocKfive}\,{\scriptsize\nvcineg{BootDiffProspHelmTraceVsFaclocKfive}}} \\
\addlinespace
\multicolumn{11}{@{}l}{\MTEBnew} \\
\quad all new & \nv{NTrainProspMteb}\,$\to$\,\nv{NTestProspMteb} & \nv{RsqProspMtebFaclocKfive} & \nv{RsqProspMtebTraceKfive} & \nv{RsqProspMtebMiKfive} & \nv{RsqProspMtebEntropyKfive} & \textbf{\nv{RsqProspMtebHubKfive}} & \nv{RsqProspMtebPcaKfive} & \nv{RsqProspMtebRandomKfive} & \nv{RsqCtrlMtebFaclocKfive} & {\scriptsize\nvneg{BootDiffProspMtebTraceVsFaclocKfive}\,{\scriptsize\nvcineg{BootDiffProspMtebTraceVsFaclocKfive}}} \\
\quad late releases & \nv{NTrainProspMtebLate}\,$\to$\,\nv{NTestProspMtebLate} & \nv{RsqProspMtebLateFaclocKfive} & \textbf{\nv{RsqProspMtebLateTraceKfive}} & \nv{RsqProspMtebLateMiKfive} & \nv{RsqProspMtebLateEntropyKfive} & \nv{RsqProspMtebLateHubKfive} & \nv{RsqProspMtebLatePcaKfive} & \nv{RsqProspMtebLateRandomKfive} & \nv{RsqCtrlMtebLateFaclocKfive} & {\scriptsize\nvneg{BootDiffProspMtebLateTraceVsFaclocKfive}\,{\scriptsize\nvcineg{BootDiffProspMtebLateTraceVsFaclocKfive}}} \\
\addlinespace
\multicolumn{11}{@{}l}{\OLLtwo, cutoff} \\
\quad 2024-10 & \nv{NTrainProspOlvTwoSubCutOctTwentyfour}\,$\to$\,\nv{NTestProspOlvTwoSubCutOctTwentyfour} & \textbf{\nv{RsqProspOlvTwoSubCutOctTwentyfourFaclocKfive}} & \nv{RsqProspOlvTwoSubCutOctTwentyfourTraceKfive} & \nv{RsqProspOlvTwoSubCutOctTwentyfourMiKfive} & \nv{RsqProspOlvTwoSubCutOctTwentyfourEntropyKfive} & \nv{RsqProspOlvTwoSubCutOctTwentyfourHubKfive} & \nv{RsqProspOlvTwoSubCutOctTwentyfourPcaKfive} & \nv{RsqProspOlvTwoSubCutOctTwentyfourRandomKfive} & \nv{RsqCtrlOlvTwoSubCutOctTwentyfourFaclocKfive} & {\scriptsize\nvneg{BootDiffProspOlvTwoSubCutOctTwentyfourTraceVsFaclocKfive}\,{\scriptsize\nvcineg{BootDiffProspOlvTwoSubCutOctTwentyfourTraceVsFaclocKfive}}} \\
\quad 2025-01 & \nv{NTrainProspOlvTwoSubCutJanTwentyfive}\,$\to$\,\nv{NTestProspOlvTwoSubCutJanTwentyfive} & \textbf{\nv{RsqProspOlvTwoSubCutJanTwentyfiveFaclocKfive}} & \nv{RsqProspOlvTwoSubCutJanTwentyfiveTraceKfive} & \nv{RsqProspOlvTwoSubCutJanTwentyfiveMiKfive} & \nv{RsqProspOlvTwoSubCutJanTwentyfiveEntropyKfive} & \nv{RsqProspOlvTwoSubCutJanTwentyfiveHubKfive} & \nv{RsqProspOlvTwoSubCutJanTwentyfivePcaKfive} & \nv{RsqProspOlvTwoSubCutJanTwentyfiveRandomKfive} & \nv{RsqCtrlOlvTwoSubCutJanTwentyfiveFaclocKfive} & {\scriptsize\nvneg{BootDiffProspOlvTwoSubCutJanTwentyfiveTraceVsFaclocKfive}\,{\scriptsize\nvcineg{BootDiffProspOlvTwoSubCutJanTwentyfiveTraceVsFaclocKfive}}} \\
\quad 2025-03 & \nv{NTrainProspOlvTwoSubCutMarTwentyfive}\,$\to$\,\nv{NTestProspOlvTwoSubCutMarTwentyfive} & \textbf{\nv{RsqProspOlvTwoSubCutMarTwentyfiveFaclocKfive}} & \nv{RsqProspOlvTwoSubCutMarTwentyfiveTraceKfive} & \nv{RsqProspOlvTwoSubCutMarTwentyfiveMiKfive} & \nv{RsqProspOlvTwoSubCutMarTwentyfiveEntropyKfive} & \nv{RsqProspOlvTwoSubCutMarTwentyfiveHubKfive} & \nv{RsqProspOlvTwoSubCutMarTwentyfivePcaKfive} & \nv{RsqProspOlvTwoSubCutMarTwentyfiveRandomKfive} & \nv{RsqCtrlOlvTwoSubCutMarTwentyfiveFaclocKfive} & {\scriptsize\nvneg{BootDiffProspOlvTwoSubCutMarTwentyfiveTraceVsFaclocKfive}\,{\scriptsize\nvcineg{BootDiffProspOlvTwoSubCutMarTwentyfiveTraceVsFaclocKfive}}} \\
\quad 2025-01, MATH & \nv{NTrainProspOlvTwoSubMathCutJanTwentyfive}\,$\to$\,\nv{NTestProspOlvTwoSubMathCutJanTwentyfive} & \textbf{\nv{RsqProspOlvTwoSubMathCutJanTwentyfiveFaclocKfive}} & \nv{RsqProspOlvTwoSubMathCutJanTwentyfiveTraceKfive} & \nv{RsqProspOlvTwoSubMathCutJanTwentyfiveMiKfive} & \nv{RsqProspOlvTwoSubMathCutJanTwentyfiveEntropyKfive} & \nv{RsqProspOlvTwoSubMathCutJanTwentyfiveHubKfive} & \nv{RsqProspOlvTwoSubMathCutJanTwentyfivePcaKfive} & \nv{RsqProspOlvTwoSubMathCutJanTwentyfiveRandomKfive} & \nv{RsqCtrlOlvTwoSubMathCutJanTwentyfiveFaclocKfive} & {\scriptsize\nvneg{BootDiffProspOlvTwoSubMathCutJanTwentyfiveTraceVsFaclocKfive}\,{\scriptsize\nvcineg{BootDiffProspOlvTwoSubMathCutJanTwentyfiveTraceVsFaclocKfive}}} \\
\addlinespace
\multicolumn{11}{@{}l}{\Epoch, cutoff} \\
\quad 2024-07 & \nv{NTrainProspEpochCutJulTwentyfour}\,$\to$\,\nv{NTestProspEpochCutJulTwentyfour} & \nv{RsqProspEpochCutJulTwentyfourFaclocKfive} & \nv{RsqProspEpochCutJulTwentyfourTraceKfive} & \nv{RsqProspEpochCutJulTwentyfourMiKfive} & \textbf{\nv{RsqProspEpochCutJulTwentyfourEntropyKfive}} & \nv{RsqProspEpochCutJulTwentyfourHubKfive} & \nv{RsqProspEpochCutJulTwentyfourPcaKfive} & \nv{RsqProspEpochCutJulTwentyfourRandomKfive} & \nv{RsqCtrlEpochCutJulTwentyfourFaclocKfive} & {\scriptsize\nvneg{BootDiffProspEpochCutJulTwentyfourTraceVsFaclocKfive}\,{\scriptsize\nvcineg{BootDiffProspEpochCutJulTwentyfourTraceVsFaclocKfive}}} \\
\quad 2025-01 & \nv{NTrainProspEpochCutJanTwentyfive}\,$\to$\,\nv{NTestProspEpochCutJanTwentyfive} & \nv{RsqProspEpochCutJanTwentyfiveFaclocKfive} & \nv{RsqProspEpochCutJanTwentyfiveTraceKfive} & \nv{RsqProspEpochCutJanTwentyfiveMiKfive} & \textbf{\nv{RsqProspEpochCutJanTwentyfiveEntropyKfive}} & \nv{RsqProspEpochCutJanTwentyfiveHubKfive} & \nv{RsqProspEpochCutJanTwentyfivePcaKfive} & \nv{RsqProspEpochCutJanTwentyfiveRandomKfive} & \nv{RsqCtrlEpochCutJanTwentyfiveFaclocKfive} & {\scriptsize\nvneg{BootDiffProspEpochCutJanTwentyfiveTraceVsFaclocKfive}\,{\scriptsize\nvcineg{BootDiffProspEpochCutJanTwentyfiveTraceVsFaclocKfive}}} \\
\quad 2025-07 & \nv{NTrainProspEpochCutJulTwentyfive}\,$\to$\,\nv{NTestProspEpochCutJulTwentyfive} & \nv{RsqProspEpochCutJulTwentyfiveFaclocKfive} & \textbf{\nv{RsqProspEpochCutJulTwentyfiveTraceKfive}} & \nv{RsqProspEpochCutJulTwentyfiveMiKfive} & \nv{RsqProspEpochCutJulTwentyfiveEntropyKfive} & \nv{RsqProspEpochCutJulTwentyfiveHubKfive} & \nv{RsqProspEpochCutJulTwentyfivePcaKfive} & \nv{RsqProspEpochCutJulTwentyfiveRandomKfive} & \nv{RsqCtrlEpochCutJulTwentyfiveFaclocKfive} & {\scriptsize\nvneg{BootDiffProspEpochCutJulTwentyfiveTraceVsFaclocKfive}\,{\scriptsize\nvcineg{BootDiffProspEpochCutJulTwentyfiveTraceVsFaclocKfive}}} \\
\quad 2026-01 & \nv{NTrainProspEpochCutJanTwentysix}\,$\to$\,\nv{NTestProspEpochCutJanTwentysix} & \nv{RsqProspEpochCutJanTwentysixFaclocKfive} & \textbf{\nv{RsqProspEpochCutJanTwentysixTraceKfive}} & \nv{RsqProspEpochCutJanTwentysixMiKfive} & \nv{RsqProspEpochCutJanTwentysixEntropyKfive} & \nv{RsqProspEpochCutJanTwentysixHubKfive} & --- & \nv{RsqProspEpochCutJanTwentysixRandomKfive} & \nv{RsqCtrlEpochCutJanTwentysixFaclocKfive} & {\scriptsize\nvneg{BootDiffProspEpochCutJanTwentysixTraceVsFaclocKfive}\,{\scriptsize\nvcineg{BootDiffProspEpochCutJanTwentysixTraceVsFaclocKfive}}} \\
\quad 2025-01, model & \nv{NTrainProspEpochByModelCutJanTwentyfive}\,$\to$\,\nv{NTestProspEpochByModelCutJanTwentyfive} & \nv{RsqProspEpochByModelCutJanTwentyfiveFaclocKfive} & \nv{RsqProspEpochByModelCutJanTwentyfiveTraceKfive} & \nv{RsqProspEpochByModelCutJanTwentyfiveMiKfive} & \nv{RsqProspEpochByModelCutJanTwentyfiveEntropyKfive} & \textbf{\nv{RsqProspEpochByModelCutJanTwentyfiveHubKfive}} & \nv{RsqProspEpochByModelCutJanTwentyfivePcaKfive} & \nv{RsqProspEpochByModelCutJanTwentyfiveRandomKfive} & \nv{RsqCtrlEpochByModelCutJanTwentyfiveFaclocKfive} & {\scriptsize\nvneg{BootDiffProspEpochByModelCutJanTwentyfiveTraceVsFaclocKfive}\,{\scriptsize\nvcineg{BootDiffProspEpochByModelCutJanTwentyfiveTraceVsFaclocKfive}}} \\
\bottomrule
\end{tabular}}
 (FB) not yet generated}}
\end{table}

\paragraph{More cutoffs, other $k$ and missing scores.}
Facility location has the highest $R^2$ at all three \OLLtwo{} cutoffs, and its lead over explained variance is significant only at January 2025 (\cref{tab:app-prosp}). With the MATH columns, facility location is significantly ahead at that cutoff and explained variance at March 2025 (EV~$-$~FL: \nvd[4]{BootDiffProspOlvTwoSubMathCutMarTwentyfiveTraceVsFaclocKfive} \nvci[4]{BootDiffProspOlvTwoSubMathCutMarTwentyfiveTraceVsFaclocKfive}; not in \cref{tab:app-prosp}). With three benchmarks the order reverses: explained variance is significantly ahead on \OLLtwo{} at January 2025 (EV~$-$~FL: \nvd{BootDiffProspOlvTwoSubCutJanTwentyfiveTraceVsFaclocKthree} \nvci{BootDiffProspOlvTwoSubCutJanTwentyfiveTraceVsFaclocKthree}) and on \MTEBnew{} (\nvd{BootDiffProspMtebTraceVsFaclocKthree} \nvci{BootDiffProspMtebTraceVsFaclocKthree}), and on \HELMLite{} the two make the same picks at $k = 1$ and~$3$. The new \MTEBnew{} models are mostly partial submissions: they observe \nvpct{FillTestProspMteb} of their cells against \nvpct{FillTestCtrlMteb} in the control splits, so their gap to the control mixes missing scores with drift.

\paragraph{Correcting for a shift of the new models.}
New models can differ from old ones as a group. The $R^2$ of \eqref{eq:r2} measures errors against the training mean, so it also rewards a selection that merely tracks such a shift. Measuring against the new models' own mean removes this effect. Under this shift-adjusted $R^2$, facility location is first on \HELMLite{} (\nv{RsqCenProspHelmFaclocKfive} against \nv{RsqCenProspHelmTraceKfive} for explained variance), on \MTEBnew{} (\nv{RsqCenProspMtebFaclocKfive} against \nv{RsqCenProspMtebTraceKfive} for explained variance and \nv{RsqCenProspMtebHubKfive} for hub) and at every \OLLtwo{} cutoff without MATH; only on the late \MTEBnew{} models is facility location not first (explained variance \nv{RsqCenProspMtebLateTraceKfive}, PCA \nv{RsqCenProspMtebLatePcaKfive}, facility location \nv{RsqCenProspMtebLateFaclocKfive}). Hub's lead on \MTEBnew{} thus disappears once the shift is removed. The adjustment also qualifies the \HELMLite{} result. Facility location's unadjusted $R^2$ is at the level of its control, but its shift-adjusted $R^2$ differs from the control's by \nvd{PenaltyCenProspHelmFaclocKfive}, and every other method loses a similar amount: the new \HELMLite{} models are harder to predict than random ones, for every selector. These are point estimates without intervals.

\paragraph{Epoch.}
\Epoch{} gives no stable ranking of the methods. New models there score far above old ones, so the unadjusted $R^2$ credits the shift: facility location's prospective $R^2$ exceeds its control by \nvd{PenaltyProspEpochCutJulTwentyfourFaclocKfive} at the July 2024 cutoff, while its shift-adjusted $R^2$ at the January 2025 cutoff is \nv{RsqCenProspEpochCutJanTwentyfiveFaclocKfive}, and no method's is above zero there. The methods' picks also reach different numbers of new models: at the July 2024 cutoff the five benchmarks of facility location are observed for \nv[0]{NModelsProspEpochCutJulTwentyfourFaclocKfive} new models and those of explained variance for only \nv[0]{NModelsProspEpochCutJulTwentyfourTraceKfive}. The ranking of the methods changes with the cutoff: facility location is far ahead at July 2024, and explained variance significantly ahead at the two latest cutoffs (\cref{tab:app-prosp}).

\subsection{Model-family holdout}
\label{app:newmodels-family}

\begin{table}[t]
\centering
\small
\caption{Model-family holdout at $k = 5$. (a)~Each family with enough models is held out in turn; $\hat R$ is estimated and the benchmarks are selected without it, and the held-out family is imputed. Entries are unweighted means over families. Gap: facility location's family-holdout $R^2$ minus its cross-validation $R^2$ (\cref{tab:app-results}); a negative gap is the cost of an unseen family. The gap compares an unweighted mean over families with the pooled $R^2$ of the matched splits. Jaccard: overlap between facility location's selection without the family and its selection on all models, averaged over families. (b)~Facility location's $R^2$ on each held-out family of the four main matrices. There are no intervals, because the families are few and not exchangeable.}
\label{tab:app-family}
{\footnotesize\setlength{\tabcolsep}{3.2pt}
\textit{(a) All matrices, $k = 5$}\\[2pt]
\begin{tabular}{@{}lrrrrrrrrrr@{}}
\toprule
Matrix & Families & FL & EV & MI & Entropy & Hub & PCA & Random & Gap & Jaccard \\
\midrule
\MMLU & \nv{NFamMmlu} & \nv{RsqFamMmluFaclocKfive} & \textbf{\nv{RsqFamMmluTraceKfive}} & \nv{RsqFamMmluMiKfive} & \nv{RsqFamMmluEntropyKfive} & \nv{RsqFamMmluHubKfive} & \nv{RsqFamMmluPcaKfive} & \nv{RsqFamMmluRandomKfive} & \nvdiff{RsqFamMmluFaclocKfive}{RsqBaseMmluFaclocKfive} & \nv[2]{JaccFamMmluFaclocKfive} \\
\MTEB & \nv{NFamMteb} & \textbf{\nv{RsqFamMtebFaclocKfive}} & \nv{RsqFamMtebTraceKfive} & \nv{RsqFamMtebMiKfive} & \nv{RsqFamMtebEntropyKfive} & \nv{RsqFamMtebHubKfive} & \nv{RsqFamMtebPcaKfive} & \nv{RsqFamMtebRandomKfive} & \nvdiff{RsqFamMtebFaclocKfive}{RsqBaseMtebFaclocKfive} & \nv[2]{JaccFamMtebFaclocKfive} \\
\Merged & \nv{NFamMerged} & \textbf{\nv{RsqFamMergedFaclocKfive}} & \nv{RsqFamMergedTraceKfive} & \nv{RsqFamMergedMiKfive} & \nv{RsqFamMergedEntropyKfive} & \nv{RsqFamMergedHubKfive} & \nv{RsqFamMergedPcaKfive} & \nv{RsqFamMergedRandomKfive} & \nvdiff{RsqFamMergedFaclocKfive}{RsqBaseMergedFaclocKfive} & \nv[2]{JaccFamMergedFaclocKfive} \\
\BenchPress & \nv{NFamBenchpress} & \nv{RsqFamBenchpressFaclocKfive} & \nv[4]{RsqFamBenchpressTraceKfive} & \nv{RsqFamBenchpressMiKfive} & \nv{RsqFamBenchpressEntropyKfive} & \nv{RsqFamBenchpressHubKfive} & \nv{RsqFamBenchpressPcaKfive} & \textbf{\nv{RsqFamBenchpressRandomKfive}} & \nvdiff[4]{RsqFamBenchpressFaclocKfive}{RsqBaseBenchpressFaclocKfive} & \nv[2]{JaccFamBenchpressFaclocKfive} \\
\OLLtwo & \nv{NFamOlvTwoSub} & \textbf{\nv{RsqFamOlvTwoSubFaclocKfive}} & \nv{RsqFamOlvTwoSubTraceKfive} & \nv{RsqFamOlvTwoSubMiKfive} & \nv{RsqFamOlvTwoSubEntropyKfive} & \nv{RsqFamOlvTwoSubHubKfive} & \nv{RsqFamOlvTwoSubPcaKfive} & \nv{RsqFamOlvTwoSubRandomKfive} & \nvdiff{RsqFamOlvTwoSubFaclocKfive}{RsqBaseOlvTwoSubFaclocKfive} & \nv[2]{JaccFamOlvTwoSubFaclocKfive} \\
\OLLone & \nv{NFamOlvOneItems} & \nv{RsqFamOlvOneItemsFaclocKfive} & \nv{RsqFamOlvOneItemsTraceKfive} & \nv{RsqFamOlvOneItemsMiKfive} & \textbf{\nv{RsqFamOlvOneItemsEntropyKfive}} & \nv{RsqFamOlvOneItemsHubKfive} & \nv{RsqFamOlvOneItemsPcaKfive} & \nv{RsqFamOlvOneItemsRandomKfive} & \nvdiff{RsqFamOlvOneItemsFaclocKfive}{RsqBaseOlvOneItemsFaclocKfive} & \nv[2]{JaccFamOlvOneItemsFaclocKfive} \\
\Epoch & \nv{NFamEpoch} & \textbf{\nv{RsqFamEpochFaclocKfive}} & \nv{RsqFamEpochTraceKfive} & \nv{RsqFamEpochMiKfive} & \nv{RsqFamEpochEntropyKfive} & \nv{RsqFamEpochHubKfive} & \nv{RsqFamEpochPcaKfive} & \nv{RsqFamEpochRandomKfive} & \nvdiff{RsqFamEpochFaclocKfive}{RsqBaseEpochFaclocKfive} & \nv[2]{JaccFamEpochFaclocKfive} \\
\HELMLite & \nv{NFamHelmLite} & \nv{RsqFamHelmLiteFaclocKfive} & \nv{RsqFamHelmLiteTraceKfive} & \textbf{\nv{RsqFamHelmLiteMiKfive}} & \nv{RsqFamHelmLiteEntropyKfive} & \nv{RsqFamHelmLiteHubKfive} & \nv{RsqFamHelmLitePcaKfive} & \nv{RsqFamHelmLiteRandomKfive} & \nvdiff{RsqFamHelmLiteFaclocKfive}{RsqBaseHelmLiteFaclocKfive} & \nv[2]{JaccFamHelmLiteFaclocKfive} \\
\HELMMMLU & \nv{NFamHelmMmlu} & \nv{RsqFamHelmMmluFaclocKfive} & \textbf{\nv{RsqFamHelmMmluTraceKfive}} & \nv{RsqFamHelmMmluMiKfive} & \nv{RsqFamHelmMmluEntropyKfive} & \nv{RsqFamHelmMmluHubKfive} & \nv{RsqFamHelmMmluPcaKfive} & \nv{RsqFamHelmMmluRandomKfive} & \nvdiff{RsqFamHelmMmluFaclocKfive}{RsqBaseHelmMmluFaclocKfive} & \nv[2]{JaccFamHelmMmluFaclocKfive} \\
\HELMClassic & \nv{NFamHelmClassic} & \textbf{\nv{RsqFamHelmClassicFaclocKfive}} & \nv{RsqFamHelmClassicTraceKfive} & \nv{RsqFamHelmClassicMiKfive} & \nv{RsqFamHelmClassicEntropyKfive} & \nv{RsqFamHelmClassicHubKfive} & \nv{RsqFamHelmClassicPcaKfive} & \nv{RsqFamHelmClassicRandomKfive} & \nvdiff{RsqFamHelmClassicFaclocKfive}{RsqBaseHelmClassicFaclocKfive} & \nv[2]{JaccFamHelmClassicFaclocKfive} \\
\MTEBnew & \nv{NFamMtebTwothousandtwentysix} & \nv{RsqFamMtebTwothousandtwentysixFaclocKfive} & \textbf{\nv{RsqFamMtebTwothousandtwentysixTraceKfive}} & \nv{RsqFamMtebTwothousandtwentysixMiKfive} & \nv{RsqFamMtebTwothousandtwentysixEntropyKfive} & \nv{RsqFamMtebTwothousandtwentysixHubKfive} & \nv{RsqFamMtebTwothousandtwentysixPcaKfive} & \nv{RsqFamMtebTwothousandtwentysixRandomKfive} & \nvdiff{RsqFamMtebTwothousandtwentysixFaclocKfive}{RsqBaseMtebTwothousandtwentysixFaclocKfive} & \nv[2]{JaccFamMtebTwothousandtwentysixFaclocKfive} \\
\bottomrule
\end{tabular}

\medskip
\textit{(b) Facility location's $R^2$ on each held-out family, $k = 5$}\\[2pt]
\begin{tabularx}{\linewidth}{@{}l>{\raggedright\arraybackslash}X@{}}
\toprule
\MMLU & gemma~\nv{RsqFamMmluFaclocKfiveGemma}; gpt~\nv{RsqFamMmluFaclocKfiveGpt}; llama~\nv{RsqFamMmluFaclocKfiveLlama}; mistral~\nv{RsqFamMmluFaclocKfiveMistral}; phi~\nv{RsqFamMmluFaclocKfivePhi}; qwen~\nv{RsqFamMmluFaclocKfiveQwen}; solar~\nv{RsqFamMmluFaclocKfiveSolar}; yi~\nv{RsqFamMmluFaclocKfiveYi} \\
\addlinespace
\MTEB & alibaba-nlp~\nv{RsqFamMtebFaclocKfiveAlibabaNlp}; baai~\nv{RsqFamMtebFaclocKfiveBaai}; intfloat~\nv{RsqFamMtebFaclocKfiveIntfloat}; jinaai~\nv{RsqFamMtebFaclocKfiveJinaai}; mcgill-nlp~\nv{RsqFamMtebFaclocKfiveMcgillNlp}; mihaiii~\nv{RsqFamMtebFaclocKfiveMihaiii}; minishlab~\nv{RsqFamMtebFaclocKfiveMinishlab}; sentence-transformers~\nv{RsqFamMtebFaclocKfiveSentenceTransformers}; snowflake~\nv{RsqFamMtebFaclocKfiveSnowflake}; voyageai~\nv{RsqFamMtebFaclocKfiveVoyageai} \\
\addlinespace
\Merged & claude~\nv{RsqFamMergedFaclocKfiveClaude}; gemini~\nv{RsqFamMergedFaclocKfiveGemini}; gemma~\nv{RsqFamMergedFaclocKfiveGemma}; gpt~\nv{RsqFamMergedFaclocKfiveGpt}; llama~\nv{RsqFamMergedFaclocKfiveLlama}; mistral~\nv{RsqFamMergedFaclocKfiveMistral}; phi~\nv{RsqFamMergedFaclocKfivePhi}; qwen~\nv{RsqFamMergedFaclocKfiveQwen}; yi~\nv{RsqFamMergedFaclocKfiveYi} \\
\addlinespace
\BenchPress & claude~\nv{RsqFamBenchpressFaclocKfiveClaude}; deepseek~\nv{RsqFamBenchpressFaclocKfiveDeepseek}; gemini~\nv{RsqFamBenchpressFaclocKfiveGemini}; gpt~\nv{RsqFamBenchpressFaclocKfiveGpt}; llama~\nv{RsqFamBenchpressFaclocKfiveLlama}; qwen~\nv{RsqFamBenchpressFaclocKfiveQwen} \\
\bottomrule
\end{tabularx}}
 (FB) not yet generated}}
\end{table}

Holding out each model family in turn leaves facility location and explained variance the two best methods on \MMLU, \MTEB{} and \Merged{} (\cref{tab:app-family}a), and on most matrices a held-out family is predicted less well than validation models in cross-validation (a negative gap). The gaps come with different changes of the selection. On \MMLU{} the selection without a family is the one on all models (Jaccard \nv[2]{JaccFamMmluFaclocKfive}), and on \MTEB{} and \MTEBnew{} it barely changes (\nv[2]{JaccFamMtebFaclocKfive} and \nv[2]{JaccFamMtebTwothousandtwentysixFaclocKfive}), so their large gaps come from new models that differ from the training families, not from a different selection. On \Merged{} the selections change a lot (\nv[2]{JaccFamMergedFaclocKfive}) at no cost in $R^2$. On the other matrices the best method varies: facility location on \Epoch, \OLLtwo{} and \HELMClassic, explained variance on \HELMMMLU{} and \MTEBnew, entropy on \OLLone, random on \BenchPress{} (as in cross-validation) and MI on \HELMLite. Families also differ among themselves (panel~b): on \MTEB{} facility location reaches \nv{RsqFamMtebFaclocKfiveJinaai} for the held-out jinaai models and \nv{RsqFamMtebFaclocKfiveMinishlab} for minishlab. The \MMLU{} gap understates the cost, because family labels come from model names and fine-tunes with uninformative names stay in training. The holdout is a split design: it tests transfer to unseen families, not the missing-at-random assumption of the estimator.

\section{Costs: budgets, fixed overheads, tokens and items}
\label{app:costs}

This appendix supports the budget paragraph of \cref{sec:robust}: it shows the budget curves, adds a fixed overhead per benchmark and token costs, and compares whole benchmarks with random items.
All comparisons are paired over the 50 matched splits with Nadeau--Bengio 95\% intervals (\cref{app:protocol-stats}).

\subsection{Knapsack and cost-blind selection}
\label{app:costs-budgets}

\begin{figure}[!hb]
\centering
\includegraphics[width=\linewidth]{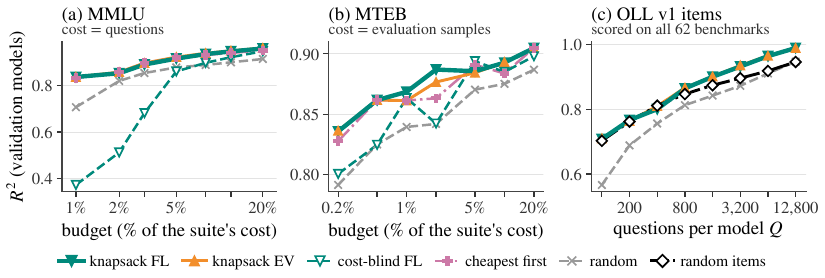}
\caption{Pricing benchmarks by cost matters when the budget is small.
(a,~b)~$R^2$ against the budget as a share of the suite's cost in questions (\MMLU) or evaluation samples (\MTEB), means over the 50 matched splits.
Cost-blind selection spends small budgets on large subjects; knapsack selection fixes this.
(c)~\OLLone: $R^2$ over all benchmarks against questions per model; whole benchmarks (knapsack FL and EV, random) against $Q$ uniformly random items.}
\label{fig:budgets}
\end{figure}

Pricing benchmarks by cost matters on \MMLU{} at small budgets, where cost-blind selection (the greedy order, filled to the budget) spends the budget on large subjects (\cref{fig:budgets}a).
We compare knapsack selection (\cref{prop:guarantees}b) with cost-blind selection, cheapest-first and random at seven budgets, given as shares of the suite's cost.
Knapsack explained variance is significantly ahead of knapsack facility location at \nv{CostFixedSigPosMmluKnapTraceVsKnapFaclocFzero} \MMLU{} budgets, by at most \nv{CostFixedPairDiffMmluKnapTraceVsKnapFaclocFzeroBfive}.
Cheapest-first is a strong baseline on \MMLU, where a few directions dominate, so many cheap subjects predict nearly as well as a few chosen ones: it is significantly ahead of knapsack facility location at \nv{CostFixedSigPosMmluCheapFirstVsKnapFaclocFzero} budgets, from 2\% to 5\% (by up to \nv{CostFixedPairDiffMmluCheapFirstVsKnapFaclocFzeroBthree}), and significantly behind it at the other \nv{CostFixedSigNegMmluCheapFirstVsKnapFaclocFzero}.
On \MTEB, knapsack and cost-blind facility location do not differ significantly, and knapsack facility location beats random significantly at only \nv{CostFixedSigPosMtebKnapFaclocVsRandomFzero} budgets.

\subsection{A fixed overhead per benchmark}
\label{app:costs-fixed}

Running a benchmark has a fixed cost (integrating it into the harness, loading data and model, checking the output) on top of its per-question cost.
We model it by adding a constant $\kappa_0$ to every benchmark's cost and set $\kappa_0$ to the median benchmark cost: \nv[0]{CostFixedMedianCostMmlu} questions on \MMLU{} and \nv[1]{CostFixedMedianCostMteb} evaluation samples on \MTEB.
Budgets are fractions of the suite's total cost including the overheads; otherwise the design is that of \cref{fig:budgets}.
An overhead makes many small benchmarks expensive, which is exactly what cheapest-first buys.

With the overhead (\cref{tab:app-costs}), no benchmark fits the 1\% budget of \MMLU{} on any split (\nv{CostFixedEmptyMmluFoneBone} of 50), cheapest-first falls behind at all affordable \MMLU{} budgets but one, and on \MTEB{} knapsack facility location leads it by \nvneg{CostFixedPairDiffMtebCheapFirstVsKnapFaclocFoneBzeropointtwo} \nvcineg{CostFixedPairDiffMtebCheapFirstVsKnapFaclocFoneBzeropointtwo} at 0.2\%.
The overhead also sharpens the case against random on \MTEB.
Knapsack explained variance is ahead of knapsack facility location at the three largest \MMLU{} budgets with the overhead, and at no \MTEB{} budget.

\begin{table}[t]
\centering
\caption{Knapsack facility location (FL) against five alternatives at every budget (share of the suite's cost), without and with a fixed overhead per benchmark (the median cost).
$\blacktriangle$/$\triangledown$: FL significantly ahead/behind (paired 95\% Nadeau--Bengio interval); $\cdot$: no significant difference; --: no benchmark fits.
Largest gap: FL minus the alternative at the budget where the difference is largest in absolute value.}
\label{tab:app-costs}
{\footnotesize\setlength{\tabcolsep}{3.4pt}
\begin{tabular}{@{}llcccccccr@{}}
\toprule
 & knapsack FL against & 1\% & 2\% & 3\% & 5\% & 8\% & 12\% & 20\% & largest gap \\
\midrule
\MMLU, no overhead & cheapest-first & $\blacktriangle$ & $\triangledown$ & $\triangledown$ & $\triangledown$ & $\blacktriangle$ & $\blacktriangle$ & $\blacktriangle$ & \nvneg{CostFixedPairDiffMmluCheapFirstVsKnapFaclocFzeroBtwenty} \\
 & random & $\blacktriangle$ & $\cdot$ & $\blacktriangle$ & $\blacktriangle$ & $\blacktriangle$ & $\blacktriangle$ & $\blacktriangle$ & \nvd{CostFixedPairDiffMmluKnapFaclocVsRandomFzeroBone} \\
 & cost-blind FL & $\blacktriangle$ & $\blacktriangle$ & $\blacktriangle$ & $\blacktriangle$ & $\blacktriangle$ & $\blacktriangle$ & $\blacktriangle$ & \nvd{CostFixedPairDiffMmluKnapFaclocVsCardFaclocFzeroBone} \\
 & knapsack MI & $\blacktriangle$ & $\triangledown$ & $\triangledown$ & $\blacktriangle$ & $\blacktriangle$ & $\blacktriangle$ & $\blacktriangle$ & \nvneg{CostFixedPairDiffMmluKnapMiVsKnapFaclocFzeroBone} \\
 & knapsack EV & $\cdot$ & $\triangledown$ & $\triangledown$ & $\triangledown$ & $\triangledown$ & $\triangledown$ & $\triangledown$ & \nvneg{CostFixedPairDiffMmluKnapTraceVsKnapFaclocFzeroBfive} \\
\addlinespace
\MMLU, overhead & cheapest-first & -- & $\blacktriangle$ & $\blacktriangle$ & $\blacktriangle$ & $\triangledown$ & $\blacktriangle$ & $\blacktriangle$ & \nvneg{CostFixedPairDiffMmluCheapFirstVsKnapFaclocFoneBtwo} \\
 & random & -- & $\blacktriangle$ & $\cdot$ & $\cdot$ & $\blacktriangle$ & $\blacktriangle$ & $\blacktriangle$ & \nvd{CostFixedPairDiffMmluKnapFaclocVsRandomFoneBtwo} \\
 & cost-blind FL & -- & $\blacktriangle$ & $\blacktriangle$ & $\blacktriangle$ & $\blacktriangle$ & $\blacktriangle$ & $\blacktriangle$ & \nvd{CostFixedPairDiffMmluKnapFaclocVsCardFaclocFoneBtwo} \\
 & knapsack MI & -- & $\blacktriangle$ & $\triangledown$ & $\cdot$ & $\blacktriangle$ & $\blacktriangle$ & $\blacktriangle$ & \nvneg{CostFixedPairDiffMmluKnapMiVsKnapFaclocFoneBtwenty} \\
 & knapsack EV & -- & $\cdot$ & $\blacktriangle$ & $\cdot$ & $\triangledown$ & $\triangledown$ & $\triangledown$ & \nvneg{CostFixedPairDiffMmluKnapTraceVsKnapFaclocFoneBeight} \\
\bottomrule
\end{tabular}
\medskip

\begin{tabular}{@{}llcccccccr@{}}
\toprule
 & knapsack FL against & 0.2\% & 0.5\% & 1\% & 2\% & 5\% & 10\% & 20\% & largest gap \\
\midrule
\MTEB, no overhead & cheapest-first & $\cdot$ & $\cdot$ & $\cdot$ & $\blacktriangle$ & $\cdot$ & $\blacktriangle$ & $\cdot$ & \nvneg{CostFixedPairDiffMtebCheapFirstVsKnapFaclocFzeroBtwo} \\
 & random & $\cdot$ & $\blacktriangle$ & $\cdot$ & $\blacktriangle$ & $\cdot$ & $\cdot$ & $\cdot$ & \nvd{CostFixedPairDiffMtebKnapFaclocVsRandomFzeroBtwo} \\
 & cost-blind FL & $\cdot$ & $\cdot$ & $\cdot$ & $\cdot$ & $\cdot$ & $\cdot$ & $\cdot$ & \nvd{CostFixedPairDiffMtebKnapFaclocVsCardFaclocFzeroBtwo} \\
 & knapsack MI & $\cdot$ & $\cdot$ & $\cdot$ & $\cdot$ & $\cdot$ & $\cdot$ & $\cdot$ & \nvneg{CostFixedPairDiffMtebKnapMiVsKnapFaclocFzeroBtwenty} \\
 & knapsack EV & $\cdot$ & $\cdot$ & $\cdot$ & $\cdot$ & $\cdot$ & $\cdot$ & $\cdot$ & \nvneg{CostFixedPairDiffMtebKnapTraceVsKnapFaclocFzeroBtwo} \\
\addlinespace
\MTEB, overhead & cheapest-first & $\blacktriangle$ & $\cdot$ & $\cdot$ & $\blacktriangle$ & $\blacktriangle$ & $\blacktriangle$ & $\blacktriangle$ & \nvneg{CostFixedPairDiffMtebCheapFirstVsKnapFaclocFoneBzeropointtwo} \\
 & random & $\cdot$ & $\cdot$ & $\blacktriangle$ & $\blacktriangle$ & $\blacktriangle$ & $\blacktriangle$ & $\blacktriangle$ & \nvd{CostFixedPairDiffMtebKnapFaclocVsRandomFoneBzeropointtwo} \\
 & cost-blind FL & $\cdot$ & $\cdot$ & $\cdot$ & $\cdot$ & $\cdot$ & $\cdot$ & $\cdot$ & \nvd{CostFixedPairDiffMtebKnapFaclocVsCardFaclocFoneBzeropointtwo} \\
 & knapsack MI & $\blacktriangle$ & $\cdot$ & $\cdot$ & $\cdot$ & $\cdot$ & $\cdot$ & $\blacktriangle$ & \nvneg{CostFixedPairDiffMtebKnapMiVsKnapFaclocFoneBzeropointtwo} \\
 & knapsack EV & $\cdot$ & $\cdot$ & $\cdot$ & $\cdot$ & $\cdot$ & $\cdot$ & $\cdot$ & \nvneg{CostFixedPairDiffMtebKnapTraceVsKnapFaclocFoneBtwo} \\
\bottomrule
\end{tabular}
}
 not yet generated}}
\end{table}

\subsection{Tokens instead of evaluation samples on \textsc{HELM}}
\label{app:costs-tokens}

\Cref{sec:robust} measures cost in questions (\MMLU) or evaluation samples (\MTEB), a proxy for compute.
For automatic evaluation the cost is closer to the number of tokens processed, which \textsc{HELM} reports per benchmark (\cref{app:data-sources}).
On \HELMMMLU{} the two agree well (Spearman correlation \nv[2]{TokSpearmanHelmMmlu} over benchmarks), but not on \HELMLite{} (\nv[2]{TokSpearmanHelmLite}) or \HELMClassic{} (\nv[2]{TokSpearmanHelmClassic}).

We hold every selection to a true token budget and let knapsack facility location price benchmarks either by tokens or by their number of evaluation samples (\textsc{HELM}'s instances).
The metric is $R^2$ over all benchmarks of the suite, with the bought benchmarks counted as known; it compares selections of different sizes on the same targets.
\Cref{tab:app-tokens} shows that counts of evaluation samples are a proxy for token cost that can mislead: pricing in tokens helps more often than it hurts, but not at every budget.
Priced in tokens, facility location beats random selection on \HELMMMLU{} at every budget from 2\%, on \HELMLite{} only at the large ones and on \HELMClassic{} at none, and it differs significantly from explained variance priced in tokens only once.

\begin{table}[t]
\centering
\caption{Facility location (FL) priced by tokens on the three \textsc{HELM} suites, against three alternatives at every token budget (share of the suite's tokens); $R^2$ over all benchmarks, with bought benchmarks counted as known.
Marks and the largest gap as in \cref{tab:app-costs}.}
\label{tab:app-tokens}
{\footnotesize\setlength{\tabcolsep}{3.4pt}
\begin{tabular}{@{}llcccccccr@{}}
\toprule
 & FL priced by tokens against & 1\% & 2\% & 5\% & 10\% & 20\% & 30\% & 50\% & largest gap \\
\midrule
\HELMMMLU & FL priced by instances & $\cdot$ & $\blacktriangle$ & $\cdot$ & $\blacktriangle$ & $\cdot$ & $\cdot$ & $\cdot$ & \nvd{TokSuitePairDiffHelmMmluFaclocTokensVsFaclocInstancesBtwo} \\
 & random & $\cdot$ & $\blacktriangle$ & $\blacktriangle$ & $\blacktriangle$ & $\blacktriangle$ & $\blacktriangle$ & $\blacktriangle$ & \nvd{TokSuitePairDiffHelmMmluFaclocTokensVsRandomBthirty} \\
 & EV priced by tokens & $\triangledown$ & $\cdot$ & $\cdot$ & $\cdot$ & $\cdot$ & $\cdot$ & $\cdot$ & \nvneg{TokSuitePairDiffHelmMmluTraceTokensVsFaclocTokensBone} \\
\addlinespace
\HELMLite & FL priced by instances & $\cdot$ & $\cdot$ & $\triangledown$ & $\blacktriangle$ & $\blacktriangle$ & $\blacktriangle$ & $\cdot$ & \nvd{TokSuitePairDiffHelmLiteFaclocTokensVsFaclocInstancesBtwenty} \\
 & random & $\cdot$ & $\cdot$ & $\cdot$ & $\cdot$ & $\blacktriangle$ & $\blacktriangle$ & $\blacktriangle$ & \nvd{TokSuitePairDiffHelmLiteFaclocTokensVsRandomBtwenty} \\
 & EV priced by tokens & $\cdot$ & $\cdot$ & $\cdot$ & $\cdot$ & $\cdot$ & $\cdot$ & $\cdot$ & \nvneg{TokSuitePairDiffHelmLiteTraceTokensVsFaclocTokensBtwenty} \\
\addlinespace
\HELMClassic & FL priced by instances & $\cdot$ & $\blacktriangle$ & $\cdot$ & $\triangledown$ & $\blacktriangle$ & $\cdot$ & $\cdot$ & \nvd{TokSuitePairDiffHelmClassicFaclocTokensVsFaclocInstancesBtwo} \\
 & random & $\cdot$ & $\cdot$ & $\cdot$ & $\cdot$ & $\cdot$ & $\cdot$ & $\cdot$ & \nvd{TokSuitePairDiffHelmClassicFaclocTokensVsRandomBone} \\
 & EV priced by tokens & $\cdot$ & $\cdot$ & $\cdot$ & $\cdot$ & $\cdot$ & $\cdot$ & $\cdot$ & \nvneg{TokSuitePairDiffHelmClassicTraceTokensVsFaclocTokensBtwo} \\
\bottomrule
\end{tabular}}
 not yet generated}}
\end{table}

\subsection{Whole benchmarks versus random items}
\label{app:costs-items}

\Cref{sec:robust-cost} compares whole benchmarks chosen by knapsack facility location with uniformly random items at a matched number~$Q$ of questions per model, on \OLLone{} (\cref{fig:budgets}c; design in \cref{app:protocol-designs}).
Without overhead, whole benchmarks are significantly ahead at every budget from \nv[0]{ItemBudgetFixedCrossQSigStayOlvOneItemsBenchFaclocVsItemsCenterCzero} questions on (at $Q=12{,}800$: \nvd{ItemBudgetPairDiffOlvOneItemsBenchFaclocVsItemsCenterQtwelvethousandeighthundred} \nvci{ItemBudgetPairDiffOlvOneItemsBenchFaclocVsItemsCenterQtwelvethousandeighthundred}).
We now charge $Q_0 \in \{10, 50, 100, 200\}$ extra questions for every benchmark that a model touches: once per selected benchmark, and once per benchmark that the random items happen to fall into.
With any of these overheads, whole benchmarks are significantly ahead from the smallest budget at which any benchmark fits on: \nv[0]{ItemBudgetFixedCrossQSigStayOlvOneItemsBenchFaclocVsItemsCenterCten}, \nv[0]{ItemBudgetFixedCrossQSigStayOlvOneItemsBenchFaclocVsItemsCenterCfifty}, \nv[0]{ItemBudgetFixedCrossQSigStayOlvOneItemsBenchFaclocVsItemsCenterConehundred} and \nv[0]{ItemBudgetFixedCrossQSigStayOlvOneItemsBenchFaclocVsItemsCenterCtwohundred} questions, the cost of the cheapest benchmark plus $Q_0$.
This result is partly a property of the uniform item design: random items touch many benchmarks, so the overhead eats most of their budget, for example \nvpct{ItemBudgetFixedOverheadShareOlvOneItemsItemsCtenQonehundredten} at $Q_0=10$ and 110 questions, and \nvpct{ItemBudgetFixedOverheadShareOlvOneItemsItemsCfiftyQeighthundred} at $Q_0=50$ and 800 questions.
An item design that knows the overhead, such as sampling a few benchmarks and many items within each, would fare better; we did not run one.

\section{Further checks}
\label{app:checks}

This appendix gives the design and the evidence behind the further checks of \cref{sec:robust} (the model-family holdout is in \cref{app:newmodels-family}) and behind four choices of \cref{sec:predict,sec:objectives}: the Gaussian conditional mean rather than a fitted imputer, the similarity inside facility location, the clipping of predictions and the scale on which scores are modeled.
Results are at $k=5$ on the 50 matched splits unless stated otherwise; intervals are paired Nadeau--Bengio 95\% intervals.

\subsection{Selection stability}
\label{app:checks-stability}

On sparse matrices the selections of different splits share few benchmarks.
Low overlap could mean that selections are unreliable, and we test this directly.
On every split we rerun each method on \nv{StabNSubMerged} random subsamples of \nvpct{StabSubFracMerged} of the training models and score each of these alternative selections on the split's validation models, with the same imputer as the full selection.
If low overlap were harmful, the alternatives would score below the full selection.

On \MMLU{} the alternatives barely differ: their mean pairwise Jaccard similarity is \nv[2]{StabJaccPairMmluFaclocKfive} for facility location, and its full selections of different splits are identical (Jaccard \nv[2]{StabCrossJaccMmluFaclocKfive}), the same five subjects in all 50 splits.
On \MTEB{} the alternatives share about three of their five benchmarks (Jaccard \nv[2]{StabJaccPairMtebFaclocKfive}) and score the same $R^2$ as the full selection: alternative minus full is \nvd{StabPairDiffMtebFaclocAltVsFullKfive} \nvci{StabPairDiffMtebFaclocAltVsFullKfive}.
On \Merged{} they share few benchmarks (\nv[2]{StabJaccPairMergedFaclocKfive}; full selections of different splits \nv[2]{StabCrossJaccMergedFaclocKfive}), yet score no worse: \nvd{StabPairDiffMergedFaclocAltVsFullKfive} \nvci{StabPairDiffMergedFaclocAltVsFullKfive}.
Explained variance behaves alike (\nv[2]{StabJaccPairMergedTraceKfive}; \nvd{StabPairDiffMergedTraceAltVsFullKfive} \nvci{StabPairDiffMergedTraceAltVsFullKfive}).
On \BenchPress{} the overlap is lowest (\nv[2]{StabJaccPairBenchpressFaclocKfive}) and the difference, \nvd{StabPairDiffBenchpressFaclocAltVsFullKfive} \nvci{StabPairDiffBenchpressFaclocAltVsFullKfive}, is too uncertain to say more.
Low overlap on sparse matrices thus reflects many proxies of similar quality, not a selection that falls apart.

\subsection{Contaminated benchmarks}
\label{app:checks-contam}

A benchmark whose questions leaked into training data scores high for the affected models, which distorts its estimated correlations; we test whether the selectors then pick it.
We simulate this on one target, MI's first pick (\nvtt{RevContamBenchmark} on \MMLU, \nvtt{RevContamMtebBenchmark} on \MTEB{} and \nvtt{RevContamMergedBenchmark} on \Merged, there among benchmarks observed for at least 30 models): for a random 10\% or 30\% of the models we add 0.1 or 0.2 of the benchmark's score range to its score (capped at the maximum), re-estimate the correlations and rerun every selector on all models, with \nv{RevContamConfigNDraws} random draws of the affected models for each of the four settings.
In the strongest setting, 30\% of the models and $+0.2$, facility location, explained variance and hub never place the target among their first five picks on any of the three matrices, in none of the draws; facility location moves it from place \nv{RevContamDrawsConfigsFracZeropointthreeDeltaZeropointtwoFaclocRankBefore} to a median place of \nv[0]{RevContamDrawsConfigsFracZeropointthreeDeltaZeropointtwoFaclocRankAfterMedian} on \MMLU.
Entropy places it among its first five in \nvpct{RevContamMtebDrawsConfigsFracZeropointthreeDeltaZeropointtwoEntropyTopFiveAfterFrac} of the \MTEB{} draws and moves it from place \nv{RevContamDrawsConfigsFracZeropointthreeDeltaZeropointtwoEntropyRankBefore} to a median place of \nv[0]{RevContamDrawsConfigsFracZeropointthreeDeltaZeropointtwoEntropyRankAfterMedian} on \MMLU: inflating some models' scores makes the target harder to predict from the other benchmarks, which is what entropy rewards.
MI demotes its former first pick on \MMLU{} and \MTEB, but on \Merged{} it keeps the target among its first five in \nvpct{RevContamMergedDrawsConfigsFracZeropointthreeDeltaZeropointtwoMiTopFiveAfterFrac} of the draws in this setting and in \nvpct{RevContamMergedDrawsConfigsFracZeropointoneDeltaZeropointoneMiTopFiveAfterFrac} of the draws in the mildest one (10\%, $+0.1$).
The check is synthetic: it inflates one benchmark by a fixed amount and does not model which models leak which questions.

\subsection{Correlation estimator}
\label{app:checks-estimator}

On sparse matrices the correlation matrix has to be estimated from incomplete data, and the estimator has several settings (\cref{app:protocol-em}).
We change one setting at a time and rerun selection and imputation on the matched splits; \cref{tab:app-estimator} shows the settings that matter on \BenchPress{} and \Merged.

\begin{table}[t]
\centering
\caption{Sensitivity to the correlation estimate: $R^2$ at $k=5$ when one setting of the estimator is changed and the rest of the pipeline is kept.
Default: the paper's estimator, EM on both matrices, with no shrinkage on \BenchPress{} (more training models than benchmarks) and shrinkage intensity $\alpha_0=(N-n_{\mathrm{tr}})/N$ toward the identity on \Merged.
Pairwise (forced): pairwise-complete correlations instead of EM.
$\alpha$ fixed: that shrinkage intensity on every split; $4\alpha_0$: four times the default intensity.
EM tol.: convergence tolerance of EM (default $5\cdot 10^{-4}$), the tighter one with four times as many iterations.
PSD floor: smallest eigenvalue allowed during EM.
Eligibility $m$: a benchmark is selectable if at least $m$ training models observe it (default 3). Bold: the best method in each row.}
\label{tab:app-estimator}
\small
{\footnotesize\setlength{\tabcolsep}{4pt}
\begin{tabular}{@{}lrrrr@{}}
\toprule
Setting & FL & EV & MI & Random \\
\midrule
\multicolumn{5}{@{}l}{\BenchPress} \\
\quad default & \nv{EstRsqBenchpressDefaultFaclocKfive} & \nv{EstRsqBenchpressDefaultTraceKfive} & \nv{EstRsqBenchpressDefaultMiKfive} & \textbf{\nv{EstRsqBenchpressDefaultRandomKfive}} \\
\quad pairwise (forced) & \nv{EstRsqBenchpressPairwiseFaclocKfive} & \textbf{\nv{EstRsqBenchpressPairwiseTraceKfive}} & \nv{EstRsqBenchpressPairwiseMiKfive} & \nv{EstRsqBenchpressPairwiseRandomKfive} \\
\quad $\alpha = 0.1$ fixed & \nv{EstRsqBenchpressAlphaPointOneFaclocKfive} & \nv{EstRsqBenchpressAlphaPointOneTraceKfive} & \nv{EstRsqBenchpressAlphaPointOneMiKfive} & \textbf{\nv{EstRsqBenchpressAlphaPointOneRandomKfive}} \\
\quad $\alpha = 0.3$ fixed & \nv{EstRsqBenchpressAlphaPointThreeFaclocKfive} & \nv{EstRsqBenchpressAlphaPointThreeTraceKfive} & \nv{EstRsqBenchpressAlphaPointThreeMiKfive} & \textbf{\nv{EstRsqBenchpressAlphaPointThreeRandomKfive}} \\
\quad EM tol.\ $5\cdot 10^{-3}$ & \textbf{\nv{EstRsqBenchpressTolLooseFaclocKfive}} & \nv{EstRsqBenchpressTolLooseTraceKfive} & \nv{EstRsqBenchpressTolLooseMiKfive} & \nv{EstRsqBenchpressTolLooseRandomKfive} \\
\quad EM tol.\ $5\cdot 10^{-5}$, $4\times$ iter. & \nv{EstRsqBenchpressTolTightFaclocKfive} & \nv{EstRsqBenchpressTolTightTraceKfive} & \nv{EstRsqBenchpressTolTightMiKfive} & \textbf{\nv{EstRsqBenchpressTolTightRandomKfive}} \\
\quad eligibility $m = 10$ & \nv{EstRsqBenchpressEligTenFaclocKfive} & \nv{EstRsqBenchpressEligTenTraceKfive} & \nv{EstRsqBenchpressEligTenMiKfive} & \textbf{\nv{EstRsqBenchpressEligTenRandomKfive}} \\
\addlinespace
\multicolumn{5}{@{}l}{\Merged} \\
\quad default & \textbf{\nv{EstRsqMergedDefaultFaclocKfive}} & \nv{EstRsqMergedDefaultTraceKfive} & \nv{EstRsqMergedDefaultMiKfive} & \nv{EstRsqMergedDefaultRandomKfive} \\
\quad pairwise (forced) & \textbf{\nv{EstRsqMergedPairwiseFaclocKfive}} & \nv{EstRsqMergedPairwiseTraceKfive} & \nv{EstRsqMergedPairwiseMiKfive} & \nv{EstRsqMergedPairwiseRandomKfive} \\
\quad $\alpha = 0.3$ fixed & \textbf{\nv{EstRsqMergedAlphaPointThreeFaclocKfive}} & \nv{EstRsqMergedAlphaPointThreeTraceKfive} & \nv{EstRsqMergedAlphaPointThreeMiKfive} & \nv{EstRsqMergedAlphaPointThreeRandomKfive} \\
\quad $\alpha = 4\alpha_0$ & \textbf{\nv{EstRsqMergedShrinkFourFaclocKfive}} & \nv{EstRsqMergedShrinkFourTraceKfive} & \nv{EstRsqMergedShrinkFourMiKfive} & \nv{EstRsqMergedShrinkFourRandomKfive} \\
\quad EM tol.\ $5\cdot 10^{-3}$ & \nv{EstRsqMergedTolLooseFaclocKfive} & \textbf{\nv{EstRsqMergedTolLooseTraceKfive}} & \nv{EstRsqMergedTolLooseMiKfive} & \nv{EstRsqMergedTolLooseRandomKfive} \\
\quad PSD floor $10^{-6}$ & \textbf{\nv{EstRsqMergedPsdMinusSixFaclocKfive}} & \nv{EstRsqMergedPsdMinusSixTraceKfive} & \nv{EstRsqMergedPsdMinusSixMiKfive} & \nv{EstRsqMergedPsdMinusSixRandomKfive} \\
\quad eligibility $m = 1$ & \textbf{\nv{EstRsqMergedEligOneFaclocKfive}} & \nv{EstRsqMergedEligOneTraceKfive} & \nv{EstRsqMergedEligOneMiKfive} & \nv{EstRsqMergedEligOneRandomKfive} \\
\quad eligibility $m = 2$ & \textbf{\nv{EstRsqMergedEligTwoFaclocKfive}} & \nv{EstRsqMergedEligTwoTraceKfive} & \nv{EstRsqMergedEligTwoMiKfive} & \nv{EstRsqMergedEligTwoRandomKfive} \\
\bottomrule
\end{tabular}}
}
\end{table}

On \BenchPress{} the estimate, not the selector, limits accuracy.
Forcing pairwise correlations raises every method, random included: facility location from \nv{EstRsqBenchpressDefaultFaclocKfive} to \nv{EstRsqBenchpressPairwiseFaclocKfive}, explained variance from \nv{EstRsqBenchpressDefaultTraceKfive} to \nv{EstRsqBenchpressPairwiseTraceKfive} and random from \nv{EstRsqBenchpressDefaultRandomKfive} to \nv{EstRsqBenchpressPairwiseRandomKfive}.
Fixed shrinkage and a looser EM tolerance help as well, and a tighter tolerance hurts.
Yet none of the \nv[0]{EstNCellsBenchpressFaclocKfive} settings that change the estimate moves facility location significantly (the largest change, pairwise, is \nvd{EstDiffBenchpressPairwiseFaclocKfive} \nvci{EstDiffBenchpressPairwiseFaclocKfive}), and even with pairwise correlations facility location does not beat random significantly (\nvd{EstPairBenchpressPairwiseFaclocVsRandomKfive} [\nvd{EstPairLoBenchpressPairwiseFaclocVsRandomKfive}, \nvd{EstPairHiBenchpressPairwiseFaclocVsRandomKfive}]).
We therefore keep the default instead of choosing an estimator after seeing the results.

On \Merged{} none of the \nv[0]{EstNCellsMergedFaclocKfive} settings changes facility location significantly.
On \MTEB{} every setting keeps it within \nv{EstMaxAbsDiffMtebFaclocKfive} of the default, and none of these changes is significant.
On \MMLU, where every training row is complete and the default is the sample correlation, only fixed shrinkage matters, and it lowers $R^2$ ($\alpha=0.3$: \nv{EstRsqMmluAlphaPointThreeFaclocKfive} against \nv{EstRsqMmluDefaultFaclocKfive}); a PSD floor of $10^{-2}$ changes $R^2$ by \nv[4]{EstMaxAbsDiffMmluPsdFloorFaclocKfive}.

\subsection{Held-out groups of benchmarks}
\label{app:checks-benchfam}

When a suite gains a new kind of benchmark, a selection made from the old benchmarks has to predict it.
We mimic this by holding out one group of related benchmarks at a time (groups in \cref{tab:app-benchfam}).
The selection may not pick from the held-out group, and its benchmarks are scored by the $R^2$ of their imputation from the selection.
A \emph{blind} selection never sees the group: the selector runs on correlations estimated without it.
An \emph{aware} selection also picks outside the group, but its objective counts the group's benchmarks as targets, using the training models' scores on them.

\begin{table}[t]
\centering
\caption{Held-out benchmark groups, $k=5$: $R^2$ on the benchmarks of the held-out group, pooled over the held-out groups within a split and averaged over the 50 splits.
Blind: selected without the group, from correlations estimated without it.
Aware: selected outside the group by an objective that targets all benchmarks.
Random: random benchmarks outside the group.
The last two columns are paired differences with 95\% intervals. Bold: the best $R^2$ in each row.
Groups: \MMLU{} and \HELMMMLU, the four \MMLU{} supercategories~\citep{Hendrycks21mmlu} (STEM, humanities, social sciences, other); \MTEB, the seven task types; \Merged, the source leaderboards, with all \MMLU{} copies in one group and both MATH copies in another (\nv{BenchFamNFamMerged} groups, of which \nv{BenchFamNFamMeanMerged} enter the pool; MT-Bench, observed for 3 models, is left out); \OLLtwo, the six parent benchmarks; \OLLone, the four \MMLU{} supercategories and each of the five other tasks.}
\label{tab:app-benchfam}
\small
{\footnotesize\setlength{\tabcolsep}{3.2pt}
\begin{tabular}{@{}lrrrrrll@{}}
\toprule
 & \multicolumn{2}{c}{FL} & EV & MI & & \multicolumn{2}{c}{paired differences, FL \scriptsize [95\% CI]} \\
\cmidrule(lr){2-3}\cmidrule(l){7-8}
Matrix & blind & aware & blind & blind & Random & blind$\,-\,$Random & aware$\,-\,$blind \\
\midrule
\MMLU & \nv{BenchFamRsqMmluAllFaclocBlindKfive} & \nv{BenchFamRsqMmluAllFaclocAwareKfive} & \textbf{\nv{BenchFamRsqMmluAllTraceBlindKfive}} & \nv{BenchFamRsqMmluAllMiBlindKfive} & \nv{BenchFamRsqMmluAllRandomKfive} & {\nvd{BenchFamPairDiffMmluAllFaclocBlindVsRandomKfive}\,{\scriptsize\nvci{BenchFamPairDiffMmluAllFaclocBlindVsRandomKfive}}} & {\nvd[4]{BenchFamPairDiffMmluAllFaclocAwareVsBlindKfive}\,{\scriptsize\nvci[4]{BenchFamPairDiffMmluAllFaclocAwareVsBlindKfive}}} \\
\MTEB & \nv{BenchFamRsqMtebAllFaclocBlindKfive} & \textbf{\nv{BenchFamRsqMtebAllFaclocAwareKfive}} & \nv{BenchFamRsqMtebAllTraceBlindKfive} & \nv{BenchFamRsqMtebAllMiBlindKfive} & \nv{BenchFamRsqMtebAllRandomKfive} & {\nvd{BenchFamPairDiffMtebAllFaclocBlindVsRandomKfive}\,{\scriptsize\nvci{BenchFamPairDiffMtebAllFaclocBlindVsRandomKfive}}} & {\nvd{BenchFamPairDiffMtebAllFaclocAwareVsBlindKfive}\,{\scriptsize\nvci{BenchFamPairDiffMtebAllFaclocAwareVsBlindKfive}}} \\
\Merged & \nv{BenchFamRsqMergedAllFaclocBlindKfive} & \textbf{\nv{BenchFamRsqMergedAllFaclocAwareKfive}} & \nv{BenchFamRsqMergedAllTraceBlindKfive} & \nv{BenchFamRsqMergedAllMiBlindKfive} & \nv{BenchFamRsqMergedAllRandomKfive} & {\nvd{BenchFamPairDiffMergedAllFaclocBlindVsRandomKfive}\,{\scriptsize\nvci{BenchFamPairDiffMergedAllFaclocBlindVsRandomKfive}}} & {\nvd{BenchFamPairDiffMergedAllFaclocAwareVsBlindKfive}\,{\scriptsize\nvci{BenchFamPairDiffMergedAllFaclocAwareVsBlindKfive}}} \\
\OLLtwo & \nv{BenchFamRsqOlvTwoSubAllFaclocBlindKfive} & \textbf{\nv{BenchFamRsqOlvTwoSubAllFaclocAwareKfive}} & \nv{BenchFamRsqOlvTwoSubAllTraceBlindKfive} & \nv{BenchFamRsqOlvTwoSubAllMiBlindKfive} & \nv{BenchFamRsqOlvTwoSubAllRandomKfive} & {\nvd{BenchFamPairDiffOlvTwoSubAllFaclocBlindVsRandomKfive}\,{\scriptsize\nvci{BenchFamPairDiffOlvTwoSubAllFaclocBlindVsRandomKfive}}} & {\nvd{BenchFamPairDiffOlvTwoSubAllFaclocAwareVsBlindKfive}\,{\scriptsize\nvci{BenchFamPairDiffOlvTwoSubAllFaclocAwareVsBlindKfive}}} \\
\OLLone & \nv{BenchFamRsqOlvOneItemsAllFaclocBlindKfive} & \textbf{\nv{BenchFamRsqOlvOneItemsAllFaclocAwareKfive}} & \nv{BenchFamRsqOlvOneItemsAllTraceBlindKfive} & \nv{BenchFamRsqOlvOneItemsAllMiBlindKfive} & \nv{BenchFamRsqOlvOneItemsAllRandomKfive} & {\nvd{BenchFamPairDiffOlvOneItemsAllFaclocBlindVsRandomKfive}\,{\scriptsize\nvci{BenchFamPairDiffOlvOneItemsAllFaclocBlindVsRandomKfive}}} & {\nvd{BenchFamPairDiffOlvOneItemsAllFaclocAwareVsBlindKfive}\,{\scriptsize\nvci{BenchFamPairDiffOlvOneItemsAllFaclocAwareVsBlindKfive}}} \\
\HELMMMLU & \nv{BenchFamRsqHelmMmluAllFaclocBlindKfive} & \nv{BenchFamRsqHelmMmluAllFaclocAwareKfive} & \textbf{\nv{BenchFamRsqHelmMmluAllTraceBlindKfive}} & \nv{BenchFamRsqHelmMmluAllMiBlindKfive} & \nv{BenchFamRsqHelmMmluAllRandomKfive} & {\nvd{BenchFamPairDiffHelmMmluAllFaclocBlindVsRandomKfive}\,{\scriptsize\nvci{BenchFamPairDiffHelmMmluAllFaclocBlindVsRandomKfive}}} & {\nvd{BenchFamPairDiffHelmMmluAllFaclocAwareVsBlindKfive}\,{\scriptsize\nvci{BenchFamPairDiffHelmMmluAllFaclocAwareVsBlindKfive}}} \\
\bottomrule
\end{tabular}}
}
\end{table}

A blind selection predicts the held-out group no better than random on \MTEB{} (\nvd{BenchFamPairDiffMtebAllFaclocBlindVsRandomKfive} \nvci{BenchFamPairDiffMtebAllFaclocBlindVsRandomKfive}) and \Merged{} (\nvd{BenchFamPairDiffMergedAllFaclocBlindVsRandomKfive} \nvci{BenchFamPairDiffMergedAllFaclocBlindVsRandomKfive}).
It is significantly ahead only on \MMLU{} (\nvd{BenchFamPairDiffMmluAllFaclocBlindVsRandomKfive}) and \OLLone{} (\nvd{BenchFamPairDiffOlvOneItemsAllFaclocBlindVsRandomKfive}), where one general factor links all groups.
Making the objective aware of the group helps significantly on \MTEB{} (\nvd{BenchFamPairDiffMtebAllFaclocAwareVsBlindKfive} \nvci{BenchFamPairDiffMtebAllFaclocAwareVsBlindKfive}), \OLLtwo{} (\nvd{BenchFamPairDiffOlvTwoSubAllFaclocAwareVsBlindKfive}), \OLLone{} and \MMLU, and its point estimate is positive on every matrix.
A coverage objective covers only the benchmarks it is asked to cover.
Once some evaluated models have scores on a new kind of benchmark, it should enter the objective as a target, even if none of its benchmarks is selected.

\subsection{Imputers}
\label{app:checks-imputers}

\Cref{sec:predict} keeps the Gaussian conditional mean because imputers fitted to the data did not predict significantly better from facility location's selections; here we also ask whether they favor a different selector.
We score every selection under two matrix-completion imputers, SoftImpute (nuclear-norm regularized completion) and HardImpute (rank-$r$ iterative SVD)~\citep{Mazumder10softimpute}, on the same cells as the Gaussian imputer (\cref{tab:app-imputers}).

\begin{table}[t]
\centering
\caption{Fitted imputers at fixed selections, $k=5$.
Each selection is scored by the Gaussian conditional mean of the paper and by SoftImpute and HardImpute, which are fitted to the training models together with the validation models' selected scores; all predictions are clipped to the valid range.
The regularization of SoftImpute and the rank of HardImpute (at most $k$) are tuned on an inner holdout of training models.
Last column: facility location's $R^2$ under the fitted imputer minus under the Gaussian imputer, with paired 95\% interval. Bold: the best selector in each row.}
\label{tab:app-imputers}
\small
{\footnotesize\setlength{\tabcolsep}{3.5pt}
\begin{tabular}{@{}llrrrrrl@{}}
\toprule
Matrix & Imputer & FL & EV & MI & Hub & Random & FL: imputer$\,-\,$Gaussian \scriptsize [95\% CI] \\
\midrule
\MMLU & Gaussian & \nv{RsqImpGaussMmluFaclocKfive} & \textbf{\nv{RsqImpGaussMmluTraceKfive}} & \nv{RsqImpGaussMmluMiKfive} & \nv{RsqImpGaussMmluHubKfive} & \nv{RsqImpGaussMmluRandomKfive} & \multicolumn{1}{c}{---} \\
 & SoftImpute & \nv{RsqImpSoftMmluFaclocKfive} & \nv{RsqImpSoftMmluTraceKfive} & \textbf{\nv{RsqImpSoftMmluMiKfive}} & \nv{RsqImpSoftMmluHubKfive} & \nv{RsqImpSoftMmluRandomKfive} & {\nvd{PairDiffImpMmluFaclocSoftVsGaussKfive}\,{\scriptsize\nvci{PairDiffImpMmluFaclocSoftVsGaussKfive}}} \\
 & HardImpute & \textbf{\nv{RsqImpHardMmluFaclocKfive}} & \textbf{\nv{RsqImpHardMmluTraceKfive}} & \nv{RsqImpHardMmluMiKfive} & \nv{RsqImpHardMmluHubKfive} & \nv{RsqImpHardMmluRandomKfive} & {\nvd{PairDiffImpMmluFaclocHardVsGaussKfive}\,{\scriptsize\nvci{PairDiffImpMmluFaclocHardVsGaussKfive}}} \\
\addlinespace
\MTEB & Gaussian & \nv{RsqImpGaussMtebFaclocKfive} & \textbf{\nv{RsqImpGaussMtebTraceKfive}} & \nv{RsqImpGaussMtebMiKfive} & \nv{RsqImpGaussMtebHubKfive} & \nv{RsqImpGaussMtebRandomKfive} & \multicolumn{1}{c}{---} \\
 & SoftImpute & \textbf{\nv{RsqImpSoftMtebFaclocKfive}} & \nv{RsqImpSoftMtebTraceKfive} & \nv{RsqImpSoftMtebMiKfive} & \nv{RsqImpSoftMtebHubKfive} & \nv{RsqImpSoftMtebRandomKfive} & {\nvd{PairDiffImpMtebFaclocSoftVsGaussKfive}\,{\scriptsize\nvci{PairDiffImpMtebFaclocSoftVsGaussKfive}}} \\
 & HardImpute & \nv{RsqImpHardMtebFaclocKfive} & \textbf{\nv{RsqImpHardMtebTraceKfive}} & \nv{RsqImpHardMtebMiKfive} & \nv{RsqImpHardMtebHubKfive} & \nv{RsqImpHardMtebRandomKfive} & {\nvd{PairDiffImpMtebFaclocHardVsGaussKfive}\,{\scriptsize\nvci{PairDiffImpMtebFaclocHardVsGaussKfive}}} \\
\addlinespace
\Merged & Gaussian & \textbf{\nv{RsqImpGaussMergedFaclocKfive}} & \nv{RsqImpGaussMergedTraceKfive} & \nv{RsqImpGaussMergedMiKfive} & \nv{RsqImpGaussMergedHubKfive} & \nv{RsqImpGaussMergedRandomKfive} & \multicolumn{1}{c}{---} \\
 & SoftImpute & \textbf{\nv{RsqImpSoftMergedFaclocKfive}} & \nv{RsqImpSoftMergedTraceKfive} & \nv{RsqImpSoftMergedMiKfive} & \nv{RsqImpSoftMergedHubKfive} & \nv{RsqImpSoftMergedRandomKfive} & {\nvd{PairDiffImpMergedFaclocSoftVsGaussKfive}\,{\scriptsize\nvci{PairDiffImpMergedFaclocSoftVsGaussKfive}}} \\
 & HardImpute & \textbf{\nv{RsqImpHardMergedFaclocKfive}} & \nv{RsqImpHardMergedTraceKfive} & \nv{RsqImpHardMergedMiKfive} & \nv{RsqImpHardMergedHubKfive} & \nv{RsqImpHardMergedRandomKfive} & {\nvd{PairDiffImpMergedFaclocHardVsGaussKfive}\,{\scriptsize\nvci{PairDiffImpMergedFaclocHardVsGaussKfive}}} \\
\addlinespace
\BenchPress & Gaussian & \nv{RsqImpGaussBenchpressFaclocKfive} & \nv{RsqImpGaussBenchpressTraceKfive} & \nv{RsqImpGaussBenchpressMiKfive} & \nv{RsqImpGaussBenchpressHubKfive} & \textbf{\nv{RsqImpGaussBenchpressRandomKfive}} & \multicolumn{1}{c}{---} \\
 & SoftImpute & \nv{RsqImpSoftBenchpressFaclocKfive} & \nv{RsqImpSoftBenchpressTraceKfive} & \nv{RsqImpSoftBenchpressMiKfive} & \textbf{\nv{RsqImpSoftBenchpressHubKfive}} & \nv{RsqImpSoftBenchpressRandomKfive} & {\nvd{PairDiffImpBenchpressFaclocSoftVsGaussKfive}\,{\scriptsize\nvci{PairDiffImpBenchpressFaclocSoftVsGaussKfive}}} \\
 & HardImpute & \nv{RsqImpHardBenchpressFaclocKfive} & \nv{RsqImpHardBenchpressTraceKfive} & \nv{RsqImpHardBenchpressMiKfive} & \textbf{\nv{RsqImpHardBenchpressHubKfive}} & \nv{RsqImpHardBenchpressRandomKfive} & {\nvd{PairDiffImpBenchpressFaclocHardVsGaussKfive}\,{\scriptsize\nvci{PairDiffImpBenchpressFaclocHardVsGaussKfive}}} \\
\bottomrule
\end{tabular}}
}
\end{table}

Neither fitted imputer beats the Gaussian conditional mean on \MMLU, \MTEB{} or \Merged.
Relative to the Gaussian imputer, SoftImpute changes facility location's $R^2$ by \nvd{PairDiffImpMmluFaclocSoftVsGaussKfive}, \nvd{PairDiffImpMtebFaclocSoftVsGaussKfive} and \nvd{PairDiffImpMergedFaclocSoftVsGaussKfive} on these three matrices, and HardImpute by \nvd{PairDiffImpMmluFaclocHardVsGaussKfive}, \nvd{PairDiffImpMtebFaclocHardVsGaussKfive} and \nvd{PairDiffImpMergedFaclocHardVsGaussKfive}, the last two not significantly.
On \BenchPress{} SoftImpute is ahead by \nvd{PairDiffImpBenchpressFaclocSoftVsGaussKfive} \nvci{PairDiffImpBenchpressFaclocSoftVsGaussKfive}, not significantly, and under it hub's selection does best (\nv{RsqImpSoftBenchpressHubKfive}).
Under both fitted imputers, facility location or explained variance stays the best selector on \MTEB{} and \Merged; under SoftImpute on \MMLU, MI's selection does best (\nv{RsqImpSoftMmluMiKfive} against \nv{RsqImpSoftMmluFaclocKfive} for facility location).
The pretrained zero-shot imputer TabImpute~\citep{Feitelberg26tabimpute}, run on the same splits and selections, falls far below the Gaussian conditional mean on \MMLU, \MTEB{} and \Merged{} (facility location: \nv{RsqTabMmluFaclocKfive}, \nv{RsqTabMtebFaclocKfive} and \nv{RsqTabMergedFaclocKfive} against \nv{RsqBaseMmluFaclocKfive}, \nv{RsqBaseMtebFaclocKfive} and \nv{RsqBaseMergedFaclocKfive}; on \MMLU{} its context was subsampled to 3{,}800 models).

\subsection{Similarity in facility location}
\label{app:checks-similarity}

Facility location credits benchmark $j$ with its best proxy's similarity, and the paper's similarity is the clipped correlation $r^+_{ij}=\max(R_{ij},0)$ of \citet{Yao26coresets}, fixed before any experiment.
Replacing it by $|R_{ij}|$ or by $R_{ij}^2$ changes no comparison significantly at $k\in\{1,3,5,10,15\}$ on any of the eleven matrices; over all $k=1,\dots,15$ the only significant difference is on \OLLone{} at $k=2$, where the clipped correlation is ahead of the squared one by \nvd{PairDiffBaseOlvOneItemsFaclocVsFaclocSqKtwo} \nvci{PairDiffBaseOlvOneItemsFaclocVsFaclocSqKtwo}.
The squared similarity changes facility location's weak first pick on \Merged{} (\cref{sec:mechanism}): its score for a single benchmark, $1+\sum_{i\neq j}R_{ij}^2$, is that of explained variance, so it makes the same first pick and reaches $R^2=\nv{RsqBaseMergedFaclocSqKone}$ instead of \nv{RsqBaseMergedFaclocKone} at $k=1$ (clipped minus squared: \nvd{PairDiffBaseMergedFaclocVsFaclocSqKone} \nvci{PairDiffBaseMergedFaclocVsFaclocSqKone}, not significant).

\subsection{Clipping and logit space}
\label{app:checks-clip}

\paragraph{Clipping to the valid range.}
Every prediction is clipped to its benchmark's valid score range (\cref{sec:predict}), and the clip is rarely active.
On \MMLU, \MTEB{} and \Merged{} the largest share of predictions outside the range is \nv{RevClipDatasetsMtebTraceOutside} of \nv{RevClipDatasetsMtebTraceTotal} (\MTEB, explained variance), and clipping changes $R^2$ by at most \nvdiff{RevClipDatasetsMtebEntropyRTwoClipListMean}{RevClipDatasetsMtebEntropyRTwoRawListMean} (\MTEB, entropy).
On \BenchPress{} the clip is active more often (up to \nv{RevClipDatasetsBenchpressMiOutside} of \nv{RevClipDatasetsBenchpressMiTotal} predictions) and matters more: it raises $R^2$ by \nvdiff{RevClipDatasetsBenchpressFaclocRTwoClipListMean}{RevClipDatasetsBenchpressFaclocRTwoRawListMean} for facility location and by \nvdiff{RevClipDatasetsBenchpressEntropyRTwoClipListMean}{RevClipDatasetsBenchpressEntropyRTwoRawListMean} for entropy.

\paragraph{Logit space.}
\label{app:checks-logit}
Fitting the Gaussian model to logit-transformed scores, which respects the bounds of accuracies, lowers the $R^2$ of facility location and explained variance on \MTEB, \Merged{} and \BenchPress{} (facility location on \Merged: \nv{RsqLogitMergedFaclocKfive} against \nv{RsqBaseMergedFaclocKfive}) and leaves them level on \MMLU; only the information objectives and random gain, and only on \MMLU{} and \BenchPress{} (entropy on \BenchPress: \nv{RsqLogitBenchpressEntropyKfive} against \nv{RsqBaseBenchpressEntropyKfive}), so we keep raw scores.

\section{Objective diagnostics}
\label{app:mechanism}

\Cref{tab:app-mech} collects the measurements behind the explanation of \cref{sec:mechanism} and the guarantees of \cref{prop:guarantees} on the four main matrices. Two effects need more than a table row: the late ties of facility location and the gap between the plug-in value of the metric and the realized~$R^2$.

\begin{table}[t]
\centering
\small
\caption{Objective diagnostics on the four main matrices, as means over the 50 matched splits. Rows~1--2: within-split Spearman correlation between an objective's value and the realized $R^2$, over all single benchmarks ($k = 1$) and over 30 random sets of five benchmarks per split. Row~3: $R^2_{j \mid \mathrm{rest}}$ of MI's first pick. Row~4: share of $P_{jj}$ ($P = \hat R^{-1}$) that comes from the 10\% smallest eigenvalues of $\hat R$, for the first picks of MI and EV. Row~5: mean $|\hat R_{ij}|$ among the five selected benchmarks. Row~6: the FL and EV sets at $k = 5$, as the share of shared picks and the Jaccard index. Row~7: greedy FL value relative to the brute-force optimum over all sets of size three (mean, worst split). Row~8: share of EV's candidate gains along its greedy path that grow after a pick, its submodularity ratio $\gamma$ at $k = 2$ and the implied guarantee $1 - e^{-\gamma}$ (\cref{app:proofs}). Row~9: realized minus plug-in $R^2$ at $k = 5$, in total and its observation-mask part (text). Row~10: FL's cover $F_{\mathrm{FL}}(\Aset_{10})/N$. Row~11: splits (of 50) with a tie-broken pick among FL's first ten.}
\label{tab:app-mech}
{\footnotesize\setlength{\tabcolsep}{4pt}
\begin{tabular}{@{}rllrrrr@{}}
\toprule
 & Quantity & & \MMLU & \MTEB & \Merged & \BenchPress \\
\midrule
1 & Spearman $\rho$, single benchmarks & FL & \nv[2]{ObjSpearmanMmluFaclocKone} & \nv[2]{ObjSpearmanMtebFaclocKone} & \nv[2]{ObjSpearmanMergedFaclocKone} & \nv[2]{ObjSpearmanBenchpressFaclocKone} \\
 &  & EV & \nv[2]{ObjSpearmanMmluTraceKone} & \nv[2]{ObjSpearmanMtebTraceKone} & \nv[2]{ObjSpearmanMergedTraceKone} & \nv[2]{ObjSpearmanBenchpressTraceKone} \\
 &  & MI & \nv[2]{ObjSpearmanMmluMiKone} & \nv[2]{ObjSpearmanMtebMiKone} & \nv[2]{ObjSpearmanMergedMiKone} & \nv[2]{ObjSpearmanBenchpressMiKone} \\
2 & Spearman $\rho$, random sets of five & FL & \nv[2]{ObjSpearmanMmluFaclocKfive} & \nv[2]{ObjSpearmanMtebFaclocKfive} & \nv[2]{ObjSpearmanMergedFaclocKfive} & \nv[2]{ObjSpearmanBenchpressFaclocKfive} \\
 &  & EV & \nv[2]{ObjSpearmanMmluTraceKfive} & \nv[2]{ObjSpearmanMtebTraceKfive} & \nv[2]{ObjSpearmanMergedTraceKfive} & \nv[2]{ObjSpearmanBenchpressTraceKfive} \\
 &  & entropy & \nv[2]{ObjSpearmanMmluEntropyKfive} & \nv[2]{ObjSpearmanMtebEntropyKfive} & \nv[2]{ObjSpearmanMergedEntropyKfive} & \nv[2]{ObjSpearmanBenchpressEntropyKfive} \\
 &  & MI & \nv[2]{ObjSpearmanMmluMiKfive} & \nv[2]{ObjSpearmanMtebMiKfive} & \nv[3]{ObjSpearmanMergedMiKfive} & \nv[2]{ObjSpearmanBenchpressMiKfive} \\
 &  & hub & \nv[2]{ObjSpearmanMmluHubKfive} & \nv[2]{ObjSpearmanMtebHubKfive} & \nv[2]{ObjSpearmanMergedHubKfive} & \nv[2]{ObjSpearmanBenchpressHubKfive} \\
\addlinespace
3 & MI's first pick: $R^2_{j\mid\mathrm{rest}}$ &  & \nv{ObjFirstMmluMiRestRsq} & \nv{ObjFirstMtebMiRestRsq} & \nv{ObjFirstMergedMiRestRsq} & \nv{ObjFirstBenchpressMiRestRsq} \\
4 & share of $P_{jj}$ from the smallest 10\% eigenvalues & MI first pick & \nvpct{ObjFirstMmluMiPLowShare} & \nvpct{ObjFirstMtebMiPLowShare} & \nvpct{ObjFirstMergedMiPLowShare} & \nvpct{ObjFirstBenchpressMiPLowShare} \\
 &  & EV first pick & \nvpct{ObjFirstMmluTracePLowShare} & \nvpct{ObjFirstMtebTracePLowShare} & \nvpct{ObjFirstMergedTracePLowShare} & \nvpct{ObjFirstBenchpressTracePLowShare} \\
\addlinespace
5 & mean $|\hat R_{ij}|$ within the five picks & FL & \nv[2]{ObjRedundMmluFaclocKfive} & \nv[2]{ObjRedundMtebFaclocKfive} & \nv[2]{ObjRedundMergedFaclocKfive} & \nv[2]{ObjRedundBenchpressFaclocKfive} \\
 &  & EV & \nv[2]{ObjRedundMmluTraceKfive} & \nv[2]{ObjRedundMtebTraceKfive} & \nv[2]{ObjRedundMergedTraceKfive} & \nv[2]{ObjRedundBenchpressTraceKfive} \\
 &  & MI & \nv[2]{ObjRedundMmluMiKfive} & \nv[2]{ObjRedundMtebMiKfive} & \nv[2]{ObjRedundMergedMiKfive} & \nv[2]{ObjRedundBenchpressMiKfive} \\
 &  & entropy & \nv[2]{ObjRedundMmluEntropyKfive} & \nv[2]{ObjRedundMtebEntropyKfive} & \nv[2]{ObjRedundMergedEntropyKfive} & \nv[2]{ObjRedundBenchpressEntropyKfive} \\
 &  & hub & \nv[2]{ObjRedundMmluHubKfive} & \nv[2]{ObjRedundMtebHubKfive} & \nv[2]{ObjRedundMergedHubKfive} & \nv[2]{ObjRedundBenchpressHubKfive} \\
 &  & PCA & \nv[2]{ObjRedundMmluPcaKfive} & \nv[2]{ObjRedundMtebPcaKfive} & \nv[2]{ObjRedundMergedPcaKfive} & \nv[2]{ObjRedundBenchpressPcaKfive} \\
 &  & random & \nv[2]{ObjRedundMmluRandomKfive} & \nv[2]{ObjRedundMtebRandomKfive} & \nv[2]{ObjRedundMergedRandomKfive} & \nv[2]{ObjRedundBenchpressRandomKfive} \\
\addlinespace
6 & FL $\cap$ EV at $k = 5$ & shared picks & \nvpct{ObjFlVsTraceSharedMmluKfive} & \nvpct{ObjFlVsTraceSharedMtebKfive} & \nvpct{ObjFlVsTraceSharedMergedKfive} & \nvpct{ObjFlVsTraceSharedBenchpressKfive} \\
 &  & Jaccard & \nv[2]{ObjFlVsTraceJaccMmluKfive} & \nv[2]{ObjFlVsTraceJaccMtebKfive} & \nv[2]{ObjFlVsTraceJaccMergedKfive} & \nv[2]{ObjFlVsTraceJaccBenchpressKfive} \\
7 & greedy / optimal $F_{\mathrm{FL}}$, $k = 3$ & mean & \nv{ObjBruteMmluFaclocKthreeFRatio} & \nv{ObjBruteMtebFaclocKthreeFRatio} & \nv{ObjBruteMergedFaclocKthreeFRatio} & \nv{ObjBruteBenchpressFaclocKthreeFRatio} \\
 &  & worst split & \nv{ObjBruteMmluFaclocKthreeFRatioMin} & \nv{ObjBruteMtebFaclocKthreeFRatioMin} & \nv{ObjBruteMergedFaclocKthreeFRatioMin} & \nv{ObjBruteBenchpressFaclocKthreeFRatioMin} \\
8 & EV: candidate gains that grow after a pick &  & \nvpct{ObjTheoryMmluTraceIncreaseFrac} & \nvpct{ObjTheoryMtebTraceIncreaseFrac} & \nvpct{ObjTheoryMergedTraceIncreaseFrac} & \nvpct{ObjTheoryBenchpressTraceIncreaseFrac} \\
 & EV: $\gamma$ at $k = 2$ &  & \nv[2]{ObjTheoryMmluTraceGamma} & \nv[2]{ObjTheoryMtebTraceGamma} & \nv[2]{ObjTheoryMergedTraceGamma} & \nv[2]{ObjTheoryBenchpressTraceGamma} \\
 & EV: bound $1 - e^{-\gamma}$ &  & \nv[2]{ObjTheoryMmluTraceGammaBound} & \nv[2]{ObjTheoryMtebTraceGammaBound} & \nv[2]{ObjTheoryMergedTraceGammaBound} & \nv[2]{ObjTheoryBenchpressTraceGammaBound} \\
\addlinespace
9 & realized $-$ plug-in $R^2$, $k = 5$ & total, FL & \nvd{ObjGapTotalTrainMmluFaclocKfive} & \nvd{ObjGapTotalTrainMtebFaclocKfive} & \nvd{ObjGapTotalTrainMergedFaclocKfive} & \nvd{ObjGapTotalTrainBenchpressFaclocKfive} \\
 &  & total, EV & \nvd{ObjGapTotalTrainMmluTraceKfive} & \nvd{ObjGapTotalTrainMtebTraceKfive} & \nvd{ObjGapTotalTrainMergedTraceKfive} & \nvd{ObjGapTotalTrainBenchpressTraceKfive} \\
 &  & mask part, FL & \nvd{ObjGapMaskTrainMmluFaclocKfive} & \nvd{ObjGapMaskTrainMtebFaclocKfive} & \nvd{ObjGapMaskTrainMergedFaclocKfive} & \nvd{ObjGapMaskTrainBenchpressFaclocKfive} \\
 &  & mask part, EV & \nvd[4]{ObjGapMaskTrainMmluTraceKfive} & \nvd{ObjGapMaskTrainMtebTraceKfive} & \nvd{ObjGapMaskTrainMergedTraceKfive} & \nvd{ObjGapMaskTrainBenchpressTraceKfive} \\
10 & FL cover $F_{\mathrm{FL}}(\mathcal{A}_{10})/N$ &  & \nv{FaclocCoverMmluKten} & \nv{FaclocCoverMtebKten} & \nv{FaclocCoverMergedKten} & \nv{FaclocCoverBenchpressKten} \\
11 & splits with a tie-broken FL pick (first 10) &  & \nv{TieFaclocSplitsMmluKten} & \nv{TieFaclocSplitsMtebKten} & \nv{TieFaclocSplitsMergedKten} & \nv{TieFaclocSplitsBenchpressKten} \\
\bottomrule
\end{tabular}}
 (FB) not yet generated}}
\end{table}

Rows~1--6 back \cref{sec:mechanism}, which discusses \MMLU{} and \Merged; the table adds \MTEB, where the correlations of row~2 lie between those of \MMLU{} and \Merged, and \BenchPress, where they are close to zero or negative. Three entries deserve comment. Hub's score ranks \MMLU{} sets in reverse (\nv[2]{ObjSpearmanMmluHubKfive}): the most redundant sets have the highest hub score and predict worst. On \MMLU{} the first pick of explained variance also draws most of its $P_{jj}$ from the smallest eigenvalues (row~4); the share separates MI from explained variance only on the three sparser matrices. And greedy facility location reaches at least \nvpct{ObjBruteMergedFaclocKthreeFRatioMin} of its optimal value on every split (row~7), far above its guarantee.

\paragraph{Late ties of facility location.}
After a few picks almost every benchmark has a close proxy. With ten picks the average cover $F_{\mathrm{FL}}(\Aset_{10})/N$ is \nv[2]{FaclocCoverMmluKten} on \MMLU{} and \nv[2]{FaclocCoverMergedKten} on \Merged. The remaining gains are small, and they can tie exactly. Suppose two candidates $i$ and $j$ improve only their own cover and each other's. Their gains are $(1 - c_i) + (r^+_{ij} - c_j)$ and $(1 - c_j) + (r^+_{ij} - c_i)$, which are equal for any $\hat R$: facility location cannot choose between two mutual proxies. The tie rule of \cref{app:protocol-rules} then picks the one with the larger explained-variance gain, which depends on $\hat R_{\Aset\Aset}^{-1}$. A tie-broken pick occurs among the first ten in \nv{TieFaclocSplitsMtebKten} of the 50 splits on \MTEB, \nv{TieFaclocSplitsMergedKten} on \Merged{} and \nv{TieFaclocSplitsBenchpressKten} on \BenchPress, and in \nv{TieFaclocSplitsMmluKten} on \MMLU. Facility location's gains need only pairwise correlations, but its late picks on sparse matrices are partly decided by the explained-variance rule.

\paragraph{Plug-in versus realized $R^2$.}
\Cref{prop:identity} turns the objective of explained variance into a forecast of the metric: the plug-in value $R^2_{\mathrm{pop}}(\Aset)$ computed from $\hat R$. We split its error, realized minus plug-in $R^2$, into two parts. The mask part is the expected $R^2$ of the actual predictor (with ridge, on the cells and models the metric scores) if $\hat R$ were the truth, minus the plug-in value; it measures the loss from validation models that observe only some of the selected benchmarks. The rest, realized minus expected $R^2$, comes from estimation error in $\hat R$, from sampling the validation models and from clipping. For explained variance at $k = 5$, realized minus plug-in $R^2$ is \nvd{ObjGapTotalTrainMmluTraceKfive} on \MMLU: the plug-in value is nearly exact. On \Merged{} it is \nvd{ObjGapTotalTrainMergedTraceKfive}, almost all of it mask (\nvd{ObjGapMaskTrainMergedTraceKfive}): sparse validation rows observe few of the selected benchmarks. On \BenchPress{} it is \nvd{ObjGapTotalTrainBenchpressTraceKfive}, of which only \nvd{ObjGapMaskTrainBenchpressTraceKfive} is mask: the estimate itself is wrong, which is why we call this matrix estimation-limited (\cref{sec:protocol,sec:results}). The same difference for facility location's sets is closer to zero on both sparse matrices (\nvd{ObjGapTotalTrainMergedFaclocKfive} on \Merged{} and \nvd{ObjGapTotalTrainBenchpressFaclocKfive} on \BenchPress). Read off the greedy path, the forecast of \cref{prop:identity} is accurate on dense matrices and optimistic on sparse ones.

\section{Changes since the submitted version}
\label{app:changes}

The submitted version recommended mutual information, evaluated selectors with ten cross-validation folds on three score matrices, and reported standard deviations over folds.
Numbers in the public reviews and author responses therefore differ from the numbers here.
\ifdefined\techreport The first arXiv version of this report is the submitted version.\fi

\paragraph{Recommendation and withdrawn claims.}
Facility location on clipped correlations, the objective of \citet{Yao26coresets}, is the recommended default; the submitted version did not include it.
Explained variance, which the submitted version only monitored, is now a main method, and we dropped a caveat about its stability that the data do not support.
Entropy and MI remain as the original objectives, now with a negative result and its explanation (\cref{sec:mechanism}).
We withdraw the claims that MI beats entropy at small $k$, the ``surrogate gap'' between them, and the recommendation of MI for small budgets.
We also withdraw statements from our author response that the new runs contradict: that hub ranking finds the same first benchmark as MI; that cost-blind MI is the worst method at the tightest budget; that knapsack MI beats cheapest-first on \MTEB; that MI ``performs well throughout'' and that objectives with guarantees degrade gracefully; and that temporal splits are infeasible.

\paragraph{What remains of the earlier claim.}
MI's lead over entropy at small~$k$ came from entropy's first step, where all benchmarks tie exactly and floating-point noise had picked the winner.
With the tie broken by the explained-variance gain (\cref{sec:objectives}), entropy makes the first pick of explained variance and beats MI on \MMLU{} at $k=1$ (entropy~$-$~MI: \nvneg{PairDiffMmluMiVsEntropyKone} \nvcineg{PairDiffMmluMiVsEntropyKone}; MI ahead in \nv{PairWinsMmluMiVsEntropyKone} of \nv{PairNMmluMiVsEntropyKone} splits).
Beyond $k=1$ neither is consistently ahead: over the \nv{ScoreNCells} scoreboard settings of \cref{tab:main}, MI is significantly ahead in \nv{ScoreMiVsEntropySigBetter} and entropy in \nv{ScoreMiVsEntropySigWorse}.

\paragraph{Protocol, data and theory.}
Ten folds with $\pm 1$ standard deviation bands became 50 matched splits shared by all methods and compared by paired Nadeau--Bengio intervals; the eligibility and target rules, the clipping of every imputer's predictions and the tie rule are new, and \MTEB{} costs count only the evaluation samples of the splits that the matrix scores (\cref{app:protocol}).
The Open LLM v2 table and \Merged{} were rebuilt to fix a downloader defect, \MMLU{} is kept as frozen with its caveats disclosed (\cref{app:data-fixes}), \BenchPress{} became a main matrix, and seven matrices were added, as were the experiments of \cref{app:newmodels,app:costs,app:checks}.
\Cref{prop:identity,prop:lower,prop:guarantees} were added: the first two are simple facts, and the third collects known guarantees.
MI's $1/e$ guarantee belongs to Random Greedy~\citep{BucFelNaoSch14}, which we now run, not to deterministic greedy or to the randomized check of our author response; fitted imputers replace the submitted matrix-completion appendix (\cref{app:checks-imputers}), and the spectral lemma and the residual-trace diagnostic were dropped.

\end{document}